\documentclass[lettersize,journal]{IEEEtran}

\usepackage{amsmath,amsfonts}
\usepackage{algorithmic}
\usepackage{algorithm}
\usepackage{array}

\usepackage{textcomp}
\usepackage{stfloats}
\usepackage{url}
\usepackage{verbatim}
\usepackage{cite}
\usepackage{graphbox}
\usepackage{makecell}
\usepackage{graphicx} 
\usepackage{multirow}
\usepackage{color}

\usepackage{bbm}

\usepackage[font=normalsize]{subfig}
\usepackage{booktabs} 
\usepackage{multirow} 
\usepackage{booktabs} 
\usepackage{pifont} 
\usepackage{tabularx} 

\usepackage{mathtools} 
\usepackage{amsthm}  
\usepackage{breakcites} 
\usepackage{tikz}
\usepackage{pgfplots}
\usepackage{verbatim}

\newcommand{\figref}[1]{Fig. \ref{#1}}
\newcommand{\tabref}[1]{Table \ref{#1}}

\newcommand{\secref}[1]{Sec. \ref{#1}}

\definecolor{srcolor}{rgb}{1,0,0}

\usepackage[switch]{lineno}
\usepackage{pgfplots}
\pgfplotsset{width=4cm,compat=1.9}
\pgfplotsset{every tick label/.append style={font=\tiny}}
\definecolor{ForestGreen}{RGB}{34,139,34}
\newcommand{\markgood}[1]{{\color{ForestGreen}#1}}

\newcommand{\cmark}{\ding{51}}
\newcommand{\xmark}{\ding{55}}

\usepackage{pifont}
\newcommand*\samethanks[1][\value{footnote}]{}
\renewcommand\footnotemark{}
\newcolumntype{M}[1]{>{\centering\arraybackslash}m{#1}}
\newcolumntype{N}{@{}m{0pt}@{}}

\usepackage[pagebackref=true,breaklinks=true,letterpaper=true,colorlinks,bookmarks=true]{hyperref}
\makeatletter
\def\hlinewd#1{%
\noalign{\ifnum0=`}\fi\hrule \@height #1 \futurelet
\reserved@a\@xhline}

\begin{document}

\title{AE-NeRF: Auto-Encoding Neural Radiance Fields for 3D-Aware Object Manipulation}

\author{
Mira Kim$^{*}$, Jaehoon Ko$^{*}$, Kyusun Cho, Junmyeong Choi, Daewon Choi,\\ and Seungryong Kim$^\dagger$
\thanks{$^*$Equal contribution}
\thanks{$^\dagger$Corresponding author}
\thanks{M. Kim, J. Ko, K. Cho, J. Choi, D.Choi and S. Kim are with the Department of Computer Science and Engineering, Korea University, Seoul 02841, Korea. (E-mail: \{miramira227, kjh9604, kyustorm7, chedge, daeone0920, seungryong\_kim\}@korea.ac.kr).}
}


\maketitle

\begin{abstract}
We propose a novel framework for 3D-aware object manipulation, called Auto-Encoding Neural Radiance Fields (AE-NeRF). Our model, which is formulated in an auto-encoder architecture, extracts disentangled 3D attributes such as 3D shape, appearance, and camera pose from an image, and a high-quality image is rendered from the attributes through disentangled generative Neural Radiance Fields (NeRF). To improve the disentanglement ability, we present two losses, global-local attribute consistency loss defined between input and output, and swapped-attribute classification loss. Since training such auto-encoding networks from scratch without ground-truth shape and appearance information is non-trivial, we present a stage-wise training scheme, which dramatically helps to boost the performance. We conduct experiments to demonstrate the effectiveness of the proposed model over the latest methods and provide extensive ablation studies. 
\end{abstract}


\section{Introduction}
Manipulating an object image in a 3D-aware fashion, e.g., deforming 3D shapes, adjusting textures, or moving camera viewpoints, is one of the most fundamental and essential tasks in Computer Vision (CV) and Computer Graphics (CG) fields with numerous applications such as 3D content creation~\cite{goel2020shape,kulkarni2019canonical} or augmented/virtual reality~\cite{Hu_2021_ICCV,infinite_nature_2020,Riegler2020FVS}. Due to its challenges, 2D image manipulation techniques~\cite{karras2019style,huang2018munit,park2020swapping} often fail to solve this problem due to lacking 3D awareness. 

Exploring how to extract disentangled 3D attributes such as 3D shape, appearance, and camera viewpoint, and how to render a high-quality image from these attributes is the key to the 3D-aware object manipulation. 
For the first task, conventional techniques for monocular 3D reconstruction~\cite{kanazawa2018end,kanazawa2018learning,bhattad2021view} rely on explicit 3D models such as mesh or voxels. They formally learn a model to extract disentangled 3D attributes, including 3D shape, texture, and camera viewpoint, from an image, and render an image from the attributes in an analysis-by-synthesis framework, where a consistency between the input and rendered image is encouraged~\cite{kato2018neural,NEURIPS2019_f5ac21cd,liu2019soft,loubet2019reparameterizing}. However, acquiring a high-quality explicit 3D model is notoriously challenging due to its limited representation capacity.

Recently, many literature have shown that an implicit volumetric representation allows for capturing and rendering high-resolution 3D structures~\cite{mescheder2019occupancy,Oechsle_2019_ICCV,Niemeyer_2019_ICCV,peng2020convolutional,niemeyer2020differentiable,sitzmann2019srns}. Among these works, Neural Radiance Field (NeRF)~\cite{mildenhall2020nerf} has recently proven success for novel view synthesis of a scene, which motivates us to use such representations for 3D-aware object manipulation task. However, since NeRF is an implicit model optimized per scene, it is non-trivial that directly controlling and editing an implicit continuous volumetric representation of a 3D object. To overcome this, some works presented image-conditional NeRF~\cite{yu2021pixelnerf,jain2021putting} that extracts an image feature through an image encoder and conditions the NeRF, which enables partially conducting 3D-aware manipulation, e.g., moving camera viewpoint~\cite{yu2021pixelnerf,chibane2021stereo,trevithick2020grf}. Nevertheless, they are not capable of editing 3D shape or appearance due to the lack of ability to extract full disentangled 3D attributes from an image.

On the other hand, some works such as GRAF~\cite{Schwarz2020graf} and GIRAFFE~\cite{niemeyer2021giraffe} presented a disentangled conditional NeRF, which aims to learn a generative model for radiance fields that renders a high-resolution image from the disentangled 3D attributes, such as 3D shape, appearance, and camera viewpoint. Since they are generative models starting from rafndomly-sampled latent vectors, they cannot be directly used for manipulation of given image. Some recent methods such as CodeNeRF~\cite{jang2021codenerf}, EditNeRF~\cite{liu2021editing}, and CLIP-NeRF~\cite{wang2021clip} presented techniques to invert such disentangled conditional NeRF through a test-time optimization to explicitly edit a given image. However, due to extremely high dimensionality of 3D shape, appearance, and camera viewpoint space, such optimization is also challenging, which generates sub-optimal results, as exemplified in 2D GAN inversion literature~\cite{zhu_2020_ECCV, xia_2021_survey}.

In this paper, we present a novel framework for 3D-aware object manipulation, dubbed Auto-Encoding Neural Radiance Fields (AE-NeRF), to tackle the aforementioned issues. Inspired by auto-encoder~\cite{hinton2006reducing}, we formulate, for the first time, an auto-encoder architecture for this task, consisting of encoder and decoder, where the former is designed to extract disentangled 3D attributes from an image and the latter is designed to render a high-quality image through disentangled conditional NeRF. In comparison to existing monocular 3D reconstruction methods~\cite{kanazawa2018end,kanazawa2018learning,bhattad2021view} that rely on explicit 3D representations, our architecture allows for reconstructing an image more accurately by leveraging implicit volumetric representations, thereby realizing an analysis-by-synthesis framework better. To train the networks by exclusively using 2D image supervision without any 3D supervision, we apply image rendering loss and perceptual reconstruction losses for better reconstruction and an adversarial loss for more realistic generation. To boost the disentanglement performance, we present two losses, namely global-local attribute consistency loss and swapped-attribute classification loss. We also present a stage-wise training scheme to stably learn our auto-encoder architecture. Comparison of our AE-NeRF with relevant existing works is summarized in~\tabref{tab:related_work_comparison}. 

The presented approach is evaluated on several standard benchmarks and examined in an ablation study. The experimental results for 3D-aware object manipulation show that this model outperforms the latest methods.


\begin{table*}[t]
\begin{center}
\caption{\textbf{Comparison of AE-NeRF with relevant existing methods,} including original NeRF~\cite{mildenhall2020nerf}, PixelNeRF~\cite{yu2021pixelnerf}, GIRAFFE~\cite{niemeyer2021giraffe}, CodeNeRF~\cite{jang2021codenerf},  EditNeRF~\cite{liu2021editing}, CG-NeRF~\cite{kyungmin2021cgnerf}, and CLIP-NeRF~\cite{wang2021clip}.}
    \label{tab:related_work_comparison}
\resizebox{\textwidth}{!}{
\begin{tabular}{l|ccccccc|c}
\toprule
&    NeRF&    PixelNeRF    &  GIRAFFE  &    EditNeRF  & CodeNeRF  & CG-NeRF  & CLIP-NeRF      & AE-NeRF \\
& \cite{mildenhall2020nerf}&    \cite{yu2021pixelnerf}    &  \cite{niemeyer2021giraffe}    & \cite{jang2021codenerf} &   \cite{liu2021editing}    & \cite{kyungmin2021cgnerf} & \cite{wang2021clip}     & (Ours)

\\ \midrule
Input image reconstruction?       & \markgood{\cmark} &  \markgood{\cmark}   & \xmark     & \markgood{\cmark}  & \markgood{\cmark}  & \xmark & \markgood{\cmark} & \markgood{\cmark}\\ 
Input image-conditioned?      & \xmark & \markgood{\cmark} & \xmark & \markgood{\cmark} & \markgood{\cmark}& \markgood{\cmark}& \markgood{\cmark}&\markgood{\cmark}  \\
Disentangled 3D attributes?  & \xmark & \xmark & \markgood{\cmark} & \markgood{\cmark} & \markgood{\cmark}& \markgood{\cmark} & \markgood{\cmark} & \markgood{\cmark}  \\
No optimization at test time? & \xmark &  \markgood{\cmark} & \markgood{\cmark}   &  \xmark  & \xmark &  \markgood{\cmark}    & \xmark & \markgood{\cmark}  \\
Camera estimation?       & \xmark & \xmark  & \xmark  & \xmark & \markgood{\cmark} & \markgood{\cmark}  & \xmark  & \markgood{\cmark} \\ \bottomrule
\end{tabular}}
\end{center}
\end{table*}

\section{Related Work}
\label{sec:related}
\subsection{Learning-based Monocular 3D Reconstruction}
Previous works on learning explicit 3D representation from an image can be categorized with respect to the representation types used to express an object, such as point clouds~\cite{lin2018learning,mandikal2019dense}, meshes~\cite{groueix2018papier,wang2018pixel2mesh}, or voxels~\cite{richter2018matryoshka,sitzmann2019deepvoxels}. 
As shown in~\figref{Fig:rel work} (a), they formally learn a model to extract disentangled 3D attributes, including 3D shape, texture, and camera viewpoint, from an image, and render an image from the attributes in an analysis-by-synthesis framework, where a consistency between the input and rendered image is encouraged~\cite{kanazawa2018learning,kanazawa2018end,goel2020shape,kulkarni2019canonical,NEURIPS2019_f5ac21cd,Liu_2019_ICCV,loubet2019reparameterizing}. 
Some others attempted to relax the constrains of using 3D supervisions, such as keypoints, camera pose~\cite{kulkarni2019canonical,goel2020shape}, or categorical mean shape~\cite{li2020self}. Meanwhile, others~\cite{NEURIPS2019_f5ac21cd, Liu_2019_ICCV, loubet2019reparameterizing} tried to improve approximation process of rasterization, that yields more compelling results on reconstruction tasks. However, since its innate memory inefficiency, such explicit representation results in limited rendering quality. 

\subsection{Implicit Neural 3D Representation}
Recent works have investigated the representation of continuous 3D shape by mapping xyz coordinates to occupancy fields~\cite{mescheder2019occupancy} or signed distance functions~\cite{jiang2020sdfdiff}. 
Recently, NeRF~\cite{mildenhall2020nerf} shows the success of the implicit representation of 3D models, which encodes a continuous volume representation of shape and view-dependent appearance with MLP. Their success inspired many follow-up works~\cite{martin2021nerf,park2021nerfies,pumarola2021d,zhi2021place,sitzmann2020implicit}. 
Although NeRF demonstrates outstanding results on novel view synthesis task, per-scene optimization induces high computational demands and requires multiple images with calibrated camera parameters for training a specific scene. 
Many approaches~\cite{trevithick2020grf,yu2021pixelnerf,chibane2021stereo} focus on acquiring a scene prior to extract features from input images to an image-conditioned rendering, which is often called conditional NeRF, as shown in~\figref{Fig:rel work} (b), but they are limited to manipulate given image, e.g., by disentangling its attributes. 

\subsection{Generative NeRF}
To reformulate the NeRF in a generative fashion, the adversarial learning framework~\cite{Schwarz2020graf,niemeyer2021giraffe,gu2021stylenerf,zhou2021cips} has been explored, where the mapping function is learned between an image and randomly-sampled attributes, e.g., 3D shape, appearance, and camera pose, as shown in~\figref{Fig:rel work} (c). GRAF~\cite{Schwarz2020graf} first took NeRF in an adversarial framework, and achieved high-quality generation performance. GIRAFFE~\cite{niemeyer2021giraffe} modeled a scene with multiple entities by feeding coarse feature volume to simple CNN-based neural renderer. More recently, some other works~\cite{gu2021stylenerf,zhou2021cips} adopt advanced neural rendering techniques, or ~\cite{deng2021gram,chan2021efficient} propose efficient point sampling strategy to achieve high-resolution image generation. Even though they achieved promising results, process of generating images from random latent codes cannot be directly used for image manipulation. 

\subsection{Disentangled Image Manipulation}
\begin{figure*}[t!]
  \centering
  \subfloat[][]{
  {\includegraphics[width=0.245\linewidth]{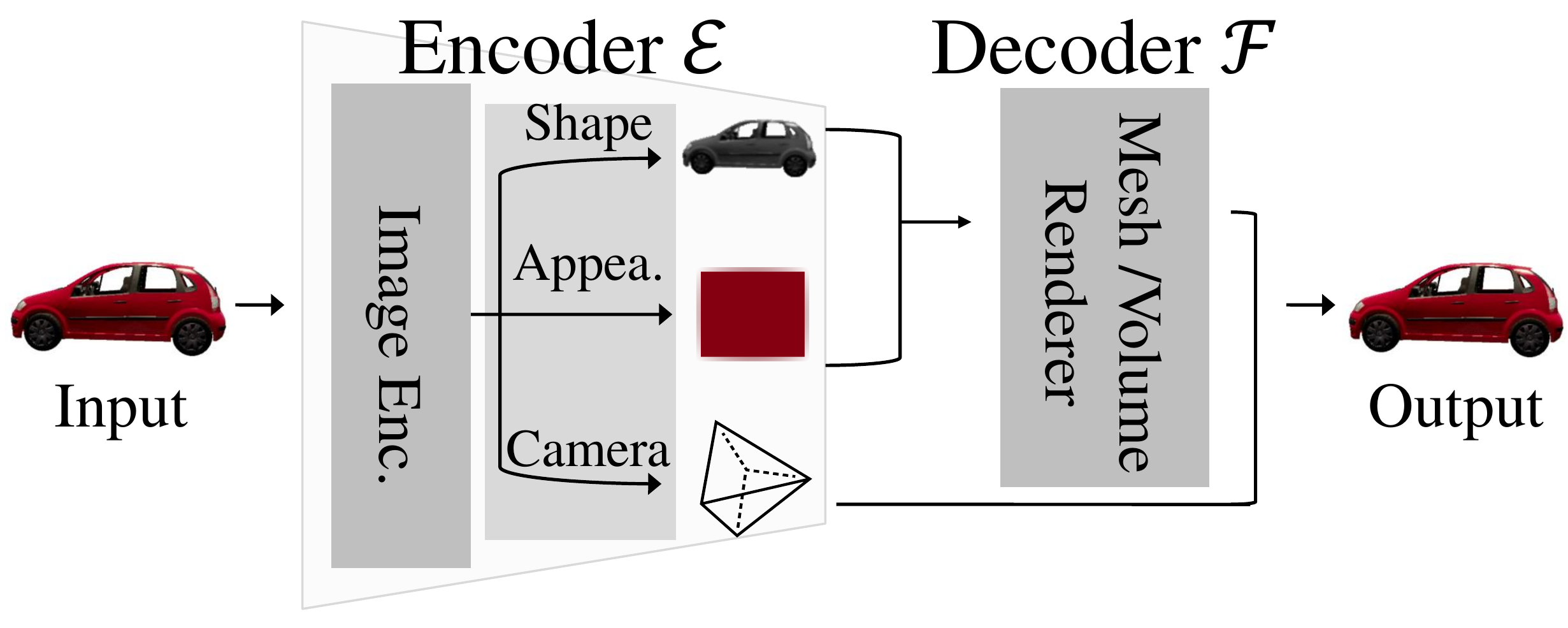}}\hfill}
  \subfloat[][]{
  {\includegraphics[width=0.245\linewidth]{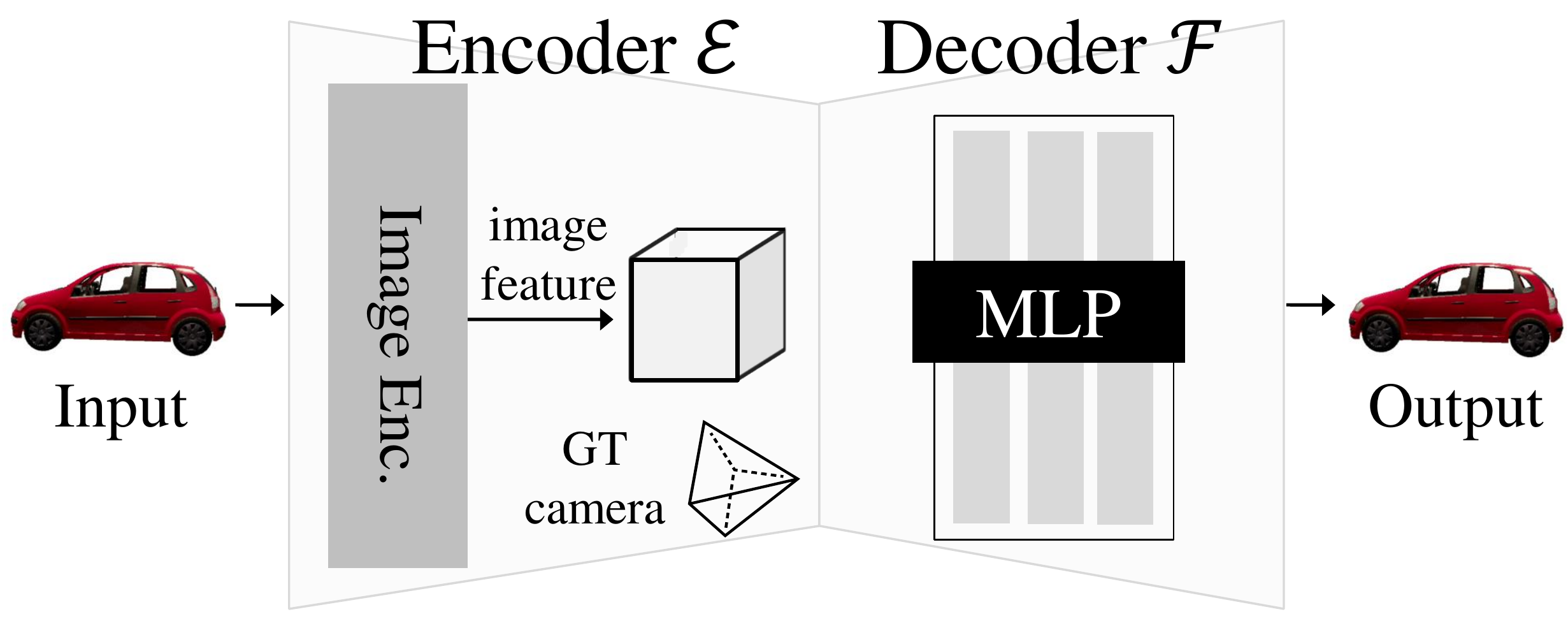}}\hfill}
  \subfloat[][]{
  {\includegraphics[width=0.245\linewidth]{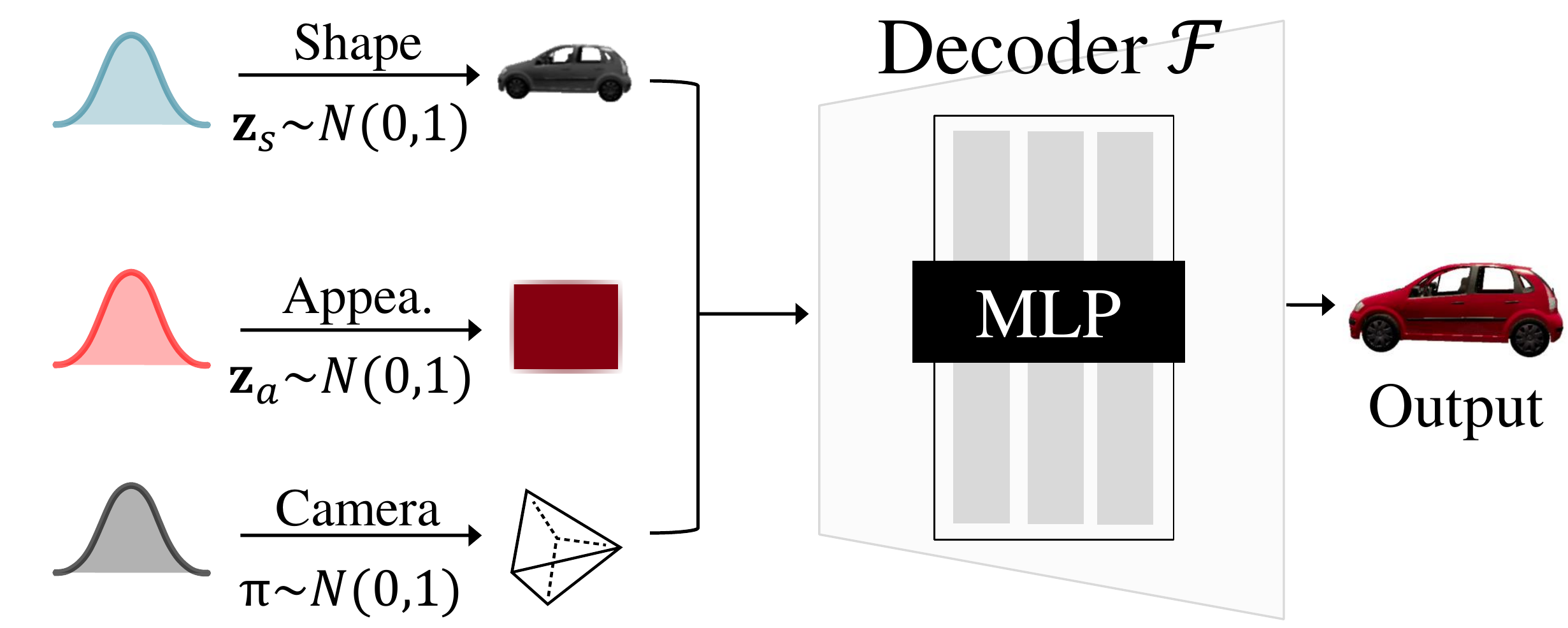}}\hfill}
  \subfloat[][]{
  {\includegraphics[width=0.245\linewidth]{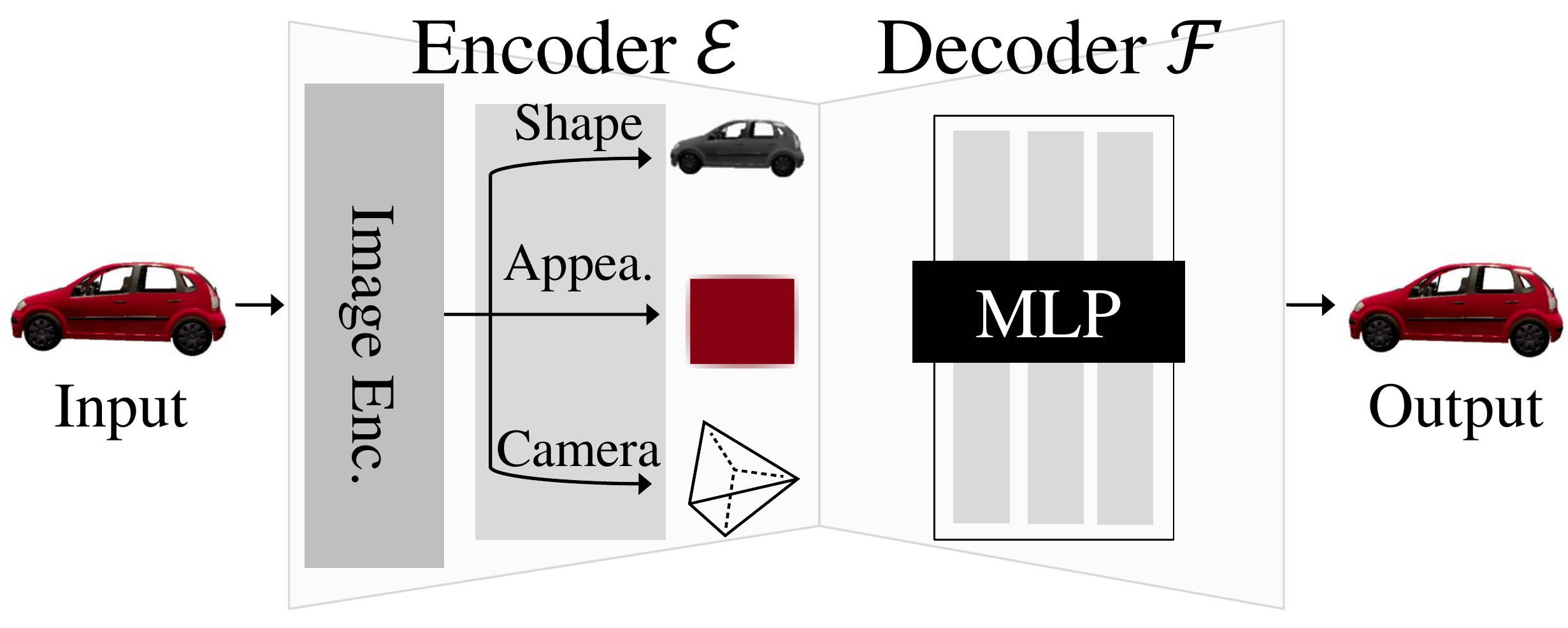}}\hfill}\\\vspace{-5pt}
    \caption{\textbf{Intuition of AE-NeRF:} (a) monocular 3D reconstruction~\cite{kanazawa2018learning,NEURIPS2019_f5ac21cd,bhattad2021view,Zhu2018VON} that extracts disentangled 3D attributes from input, (b) image-conditional NeRF~\cite{yu2021pixelnerf,jain2021putting,trevithick2020grf,chibane2021stereo} that trains the feature encoder and NeRF as decoder, (c) disentangled generative NeRF~\cite{Schwarz2020graf,niemeyer2021giraffe} that renders images from randomly sampled disentangled 3D attributes, and (d) auto-encoding NeRF (AE-NeRF) that extracts the disentangled 3D attributes from input and renders images from these attributes through disentangled generative NeRF, which can be effectively used for 3D-aware object manipulation.}
\label{Fig:rel work}
\end{figure*}

Disentangling an image as interpretable attributes can be used in many downstream applications, such as image-to-image translation~\cite{Bhattacharjee_2020_CVPR,kotovenko2019content,wu2019transgaga}, domain adaptation~\cite{li2019cross}, or image manipulation~\cite{karras2019style,karras2020analyzing,kazemi2019style,esser2019unsupervised}. For instance, Swapping Auto-encoder~\cite{park2020swapping} learns a disentangled embedding space and manipulates an image. However, these works mainly focused on modeling 2D attributes.  For 3D-aware object manipulation, a few recent works~\cite{jang2021codenerf,liu2021editing,wang2021clip} presented the use of NeRF. CodeNeRF~\cite{jang2021codenerf} estimated shape, texture, and camera viewpoint of an object from a single image by optimizing a pre-trained NeRF. CoNeRF~\cite{kacper2021conerf} trained NeRF with a small number of mask annotations via auto-decoding optimization so that the user can manipulate the semantic parts of the image at test time.
EditNeRF~\cite{liu2021editing} learned to disentangle color from shape when predicting the radiance of a point, enabling editing shape and color. As a concurrent work, CG-NeRF~\cite{kyungmin2021cgnerf} disentangles attributes given an input  with pose-consistent diversity loss for view-consistent multi-modal outputs. CLIP-NeRF~\cite{wang2021clip} exploits the GAN inversion~\cite{zhu2020domain} to effectively optimize disentangled latent codes. Our method differs from these methods in that our model directly predicts the disentangled 3D attributes through a learned encoder, as shown in~\figref{Fig:rel work} (d), 

\section{Methodology}
\subsection{Motivation and Overview}
Let us denote an image as $I$ that we would like to manipulate. Here, our objective is to manipulate $I$ in a 3D-aware fashion, e.g., novel view synthesis, shape and appearance swap, or camera pose swap with another image $J$. To achieve this, we focus on how to extract disentangled 3D attributes from an image and how to render a high-quality output from these attributes, which can be formulated with an auto-encoder architecture~\cite{hinton2006reducing}. We design a model $\mathcal{G}$, as shown in~\figref{fig:framework}, consisting of the encoder $\mathcal{E}$ and decoder $\mathcal{F}$, where the former is designed to extract disentangled 3D attributes, i.e., appearance code $\mathbf{z}_a$ and shape code $\mathbf{z}_s$, as well as camera pose $\pi$, from the input image $I$, and the latter is designed to render the output image $I'$ from these attributes through disentangled generative NeRF. At the manipulation phase, we extract disentangled 3D attributes through the learned encoder given an image, adjust the attributes conforming to user intention, and then render a result through the learned decoder. 

\subsection{Encoder: Disentangled 3D Attribute Extraction}
\label{Encoder}
To understand 3D perspective of an object, disentangling camera pose and object properties, e.g., shape and appearance, from an image is of prime importance~\cite{Schwarz2020graf}. However, achieving this from a single image is notoriously challenging without explicit supervisions of them or prior knowledge. In 2D image manipulation, there were many attempts to disentangle an image as content and style~\cite{DRIT,MSGAN,lee2020drit++,huang2018multimodal}, but they lack 3D awareness, thus hindering applicability. To overcome this, similar to existing 3D shape recovery methods~\cite{bhattad2021view,kanazawa2018learning}, we attempt to extract disentangled 3D geometry and appearance information. But, unlike them~\cite{bhattad2021view,kanazawa2018learning} that rely on \textit{explicit} 3D structure, e.g., mesh, point cloud or voxel, our encoder is designed to extract the latent codes for \textit{implicit} 3D representation, i.e., neural radiance fields, that can be used for NeRF in the decoder.

Specifically, our encoder $\mathcal{E}$ is composed of the image 
encoder and the separate attribute estimators; the former takes as input an image $I$ and extracts a semantically meaningful feature through CNNs~\cite{he2016deep}, and the latter outputs appearance code $\mathbf{z}_a$ and shape code $\mathbf{z}_s$, as well as camera pose $\pi$, as follows:
\begin{equation}
    \{\mathbf{z}_a,\mathbf{z}_s,\pi\}=\mathcal{E}(I).
\end{equation}

\begin{figure*}[t]
\centering
\includegraphics[width=1.0\linewidth]{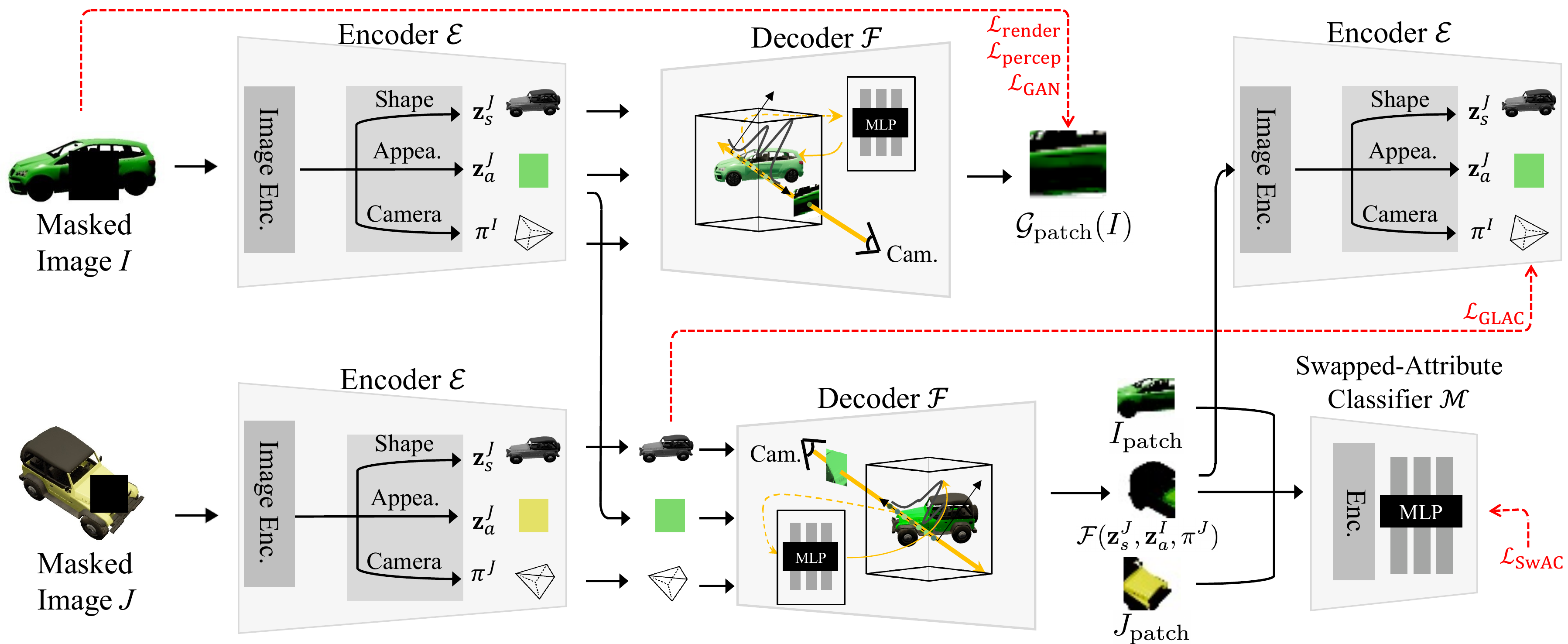}\\
\caption{\textbf{Overall network configuration and loss functions of AE-NeRF.} Our network consists of the encoder $\mathcal{E}$ for extracting disentangled 3D attributes, including appearance code, shape code, and camera pose, and the decoder $\mathcal{F}$ for rendering the output image through disentangled generative NeRF. To train the networks, 2D rendering loss, perceptual loss, and adversarial loss are used. 
To improve the disentanglement, we also apply a global-local attribute consistency loss between the attributes from inputs and outputs. Moreover, by swapping the attributes across different instances,
we boost the disentangling performance using swapped-attribute classifier.}
\label{fig:framework}
\end{figure*}

During training, to encourage the networks to focus more on the context information, we hide a part of input image (i.e., patch $I_\mathrm{patch}$) randomly. This simple technique dramatically improves the performance, as exemplified in~\figref{fig:framework}. 

\subsection{Decoder: Disentangled Conditional NeRF}\label{Decoder}
To design an auto-encoder architecture, we design the decoder $\mathcal{F}$ to form a mapping between disentangled 3D attributes $\{\mathbf{z}_a,\mathbf{z}_s,\pi\}$ and an image $I$. Existing image renderers for inverse graphics such as DIB-R~\cite{NEURIPS2019_f5ac21cd} demand explicit 3D representation such as mesh, thus having limited image generation quality. Unlike those~\cite{NEURIPS2019_f5ac21cd,kanazawa2018learning}, we leverage a neural rendering technique based on NeRF~\cite{mildenhall2020nerf,Schwarz2020graf,niemeyer2021giraffe} that encodes an image as a continuous volumetric radiance field of color and density. Thanks to its strong capability in capturing high-resolution geometry~\cite{mildenhall2020nerf,Schwarz2020graf,niemeyer2021giraffe}, it enables rendering photo-realistic images in the current camera view and even novel camera views, which allows for building an accurate cycle across an image to latent space and to the image within our auto-encoder architecture. 

Specifically, built upon the NeRF~\cite{mildenhall2020nerf}, our decoder is represented as a continuous volumetric function $\mathcal{F}$ which maps a 3D location $\mathbf{x}$ and a viewing direction\footnote{The viewing direction $\mathbf{d}$ is derived by estimated camera pose $\pi$.} $\mathbf{d}$ from the estimated camera pose $\pi$, together with appearance code $\mathbf{z}_a$ and shape code $\mathbf{z}_s$, to volume density $\sigma$ and RGB color $\mathbf{c}$, which can be formulated as
\begin{equation}
    \{\sigma,\mathbf{c}\} = \mathcal{F}(\gamma(\mathbf{x}),\gamma(\mathbf{d}),\mathbf{z}_a,\mathbf{z}_s),
\end{equation}
where $\gamma(\cdot)$ is a positional encoding~\cite{mildenhall2020nerf}. For architecture design of $\mathcal{F}$, we follow the GRAF~\cite{Schwarz2020graf}, while any other architectures~\cite{jang2021codenerf,liu2021editing,wang2021clip} can also be used. 

\subsection{Loss Functions}
\label{loss functions}

\subsubsection{Image Reconstruction Loss.}
Thanks to auto-encoding nature, we can utilize a 2D rendering loss between an input image $I$ and a rendered image $\mathcal{G}(I)$. However, due to high-computational burden of volume rendering in NeRF~\cite{mildenhall2020nerf}, generating an image at each iteration during training would be non-trivial. To overcome this, instead of an image, we render a patch $\mathcal{G}_\mathrm{patch}(I)$ similar to the GRAF~\cite{Schwarz2020graf}. In addition, to encourage the networks to more focus on context information, we artificially hide a patch $I_\mathrm{patch}$ in the image $I$ and make the networks recover the patch, which is proven to be effective in vision tasks~\cite{he2021masked,dosovitskiy2021image}. We formulate this with $l$-2 loss function between $I_\mathrm{patch}$ and $\mathcal{G}_\mathrm{patch}(I)$ as
\begin{equation}
    \mathcal{L}_\mathrm{render}(I) = \sum\nolimits_{I} {\|I_\mathrm{patch} - \mathcal{G}_\mathrm{patch}(I)\|_2}. 
\end{equation}

\subsubsection{Perceptual Reconstruction Loss.}
To improve the generation quality, we further utilize a perceptual loss~\cite{johnson2016perceptual,zhu2020indomain} between the extracted features from an input image patch $I_\mathrm{patch}$ and a rendered patch $\mathcal{G}_\mathrm{patch}(I)$ such that
\begin{equation}
    \mathcal{L}_\mathrm{percep}(I) = \sum\nolimits_{I} {\|\mathcal{R}(I_{\mathrm{patch}}) - \mathcal{R}(\mathcal{G}_{\mathrm{patch}}(I))\|_2},
\end{equation}
where $\mathcal{R}$ represents VGG feature extractor~\cite{simonyan2015deep}. Here, as~\cite{johnson2016perceptual} suggests, we compare the feature output of the first convolutional block of pretrained VGG-16~\cite{simonyan2015deep}, extracted from input image and the generated patch. 

\subsubsection{Adversarial Loss.}   
To generate the photo-realistic output and apply losses on the novel views projected from another camera pose, we leverage an adversarial learning with a non-saturating GAN objective~\cite{goodfellow2014generative} and $R_1$ gradient penalty~\cite{gulrajani2017improved}: 
\vspace{-5pt}
\begin{equation}
\begin{split}
       \mathcal{L}_\mathrm{GAN}(I, p) =&
    \mathbb{E}_{I,p} [-\mathrm{log}(\mathcal{D}(\mathcal{G}_{\mathrm{patch}}(I, p)))+ \\ & \lambda\|\nabla(\mathcal{D}(\mathcal{G}_{\mathrm{patch}}(I, p)))\|_1],
    \label{loss:gan} 
\end{split}
\end{equation}
where $\mathcal{D}(\cdot)$ is a discriminator and $p$ is a perturbation that is applied to latent attributes for better modelling an image manifold, which will be discussed in \secref{training_strategy}. We apply
the GAN loss to not only reconstructed patches, but also the generated patches from the combinations of swapped attributes. 

\subsubsection{Swapped-Attribute Classification Loss.}
\label{paragraph:Swapped-Attribute Classification Loss.}
To improve the disentanglement, we first swap the estimated attributes across different instances $I$ and $J$ to render attribute-swapped images, and then distinguish which attribute-swapping is applied, e.g., appearance swap or shape swap.
In specific, we design a swapped-attribute classifier $\mathcal{M}$ that takes $I_{\mathrm{patch}}$, $J_{\mathrm{patch}}$, and a rendered patch from swapped-attribute as input and outputs the probability whether the swapped attribute is either shape or appearance. Even though there exist many possible combinations of swapped-attributes, we only consider appearance swap and shape swap from $J$ to $I$, since estimating camera pose swap is notoriously challenging. We define a swapped-attribute classification loss such that
\vspace{-5pt}
\begin{multline}
         \mathcal{L}_\mathrm{SwAC}(I, J) =  \\ 
            -\sum\nolimits_{I, J}y\mathrm{log}(\mathcal{M}(I_{\mathrm{patch}}, J_{\mathrm{patch}}, \mathcal{F}(\mathbf{z}^{J}_{s}, \mathbf{z}^{I}_{a}, \pi^{J}))) \\
            +(1-y)\mathrm{log}(1-\mathcal{M}(I_{\mathrm{patch}}, J_{\mathrm{patch}}, \mathcal{F}(\mathbf{z}^{I}_{s}, \mathbf{z}^{J}_{a}, \pi^{J}))),
\end{multline}

where we arbitrarily set the label ${y}$ as 0 when the shape attribute is swapped and set the label as 1 in the other case. During training, $I_{\mathrm{patch}}$ and $J_{\mathrm{patch}}$ are randomly sampled to maximize generalization ability. Supervised with this binary classification loss, the encoder can strictly differentiate the shape and appearance attributes. 

\subsubsection{Global-Local Attribute Consistency Loss.} 
To further improve the disentanglement ability of the network, we synthesize a patch from swapped attributes and pass it through the encoder, enforcing the output of encoder to be the same as the swapped attributes. In detail, given the rendered patch from swapped-attributes, e.g., $\mathcal{F}(\mathbf{z}^{J}_{s}, \mathbf{z}^{I}_{a}, \pi^{J})$, we extract the attributes from the patch. We then constrain the output attributes to follow the original swapped attributes, which helps the encoder to boost the disentanglement performance and to relate the attributes from local patch and the corresponding attributes from global images. This can be interpreted as the local patch contains the attribute information from the two images that generate the swapped attribute pairs. Note that it is possible to extract the attribute from inputs with different resolutions by the same encoder as our encoder contains a global average pooling (GAP) module~\cite{he2016deep}. We define the global-local attribute consistency loss in case of appearance swapping for image J as:
\begin{equation}
    \mathcal{L}_\mathrm{GLAC}(I, J) = \sum\nolimits_{I, J} \mathcal{S}_{\mathrm{latent}}(\{\mathbf{z}^{J}_{s}, \mathbf{z}^{I}_{a}\}, \mathcal{E}'(\mathcal{F}(\mathbf{z}_s^{J}, \mathbf{z}_a^{I}, \pi^{I}))),
\end{equation}
where $\mathcal{E}'(I)$ outputs $\{\mathbf{z}_s^{I}, \mathbf{z}_a^{I}\}$ and $\mathcal{S}_{\mathrm{latent}}(\mathbf{z}^I, \mathbf{z}^J) = \|\mathbf{z}^{I} - \mathbf{z}^{J}\|_{2}$. In this loss, we ignore the camera pose consistency since it is extremely challenging to estimate a camera pose from a local patch unlike appearance and shape codes. We swap shape attribute for image J by replacing $\mathbf{z}^{J}_{s}$ to $\mathbf{z}^{I}_{s}$ and $\mathbf{z}^{J}_{a}$ to $\mathbf{z}^{I}_{s}$.

\begin{figure}[t]
  \centering
  \subfloat[][]{
  {\includegraphics[width=0.18\linewidth]{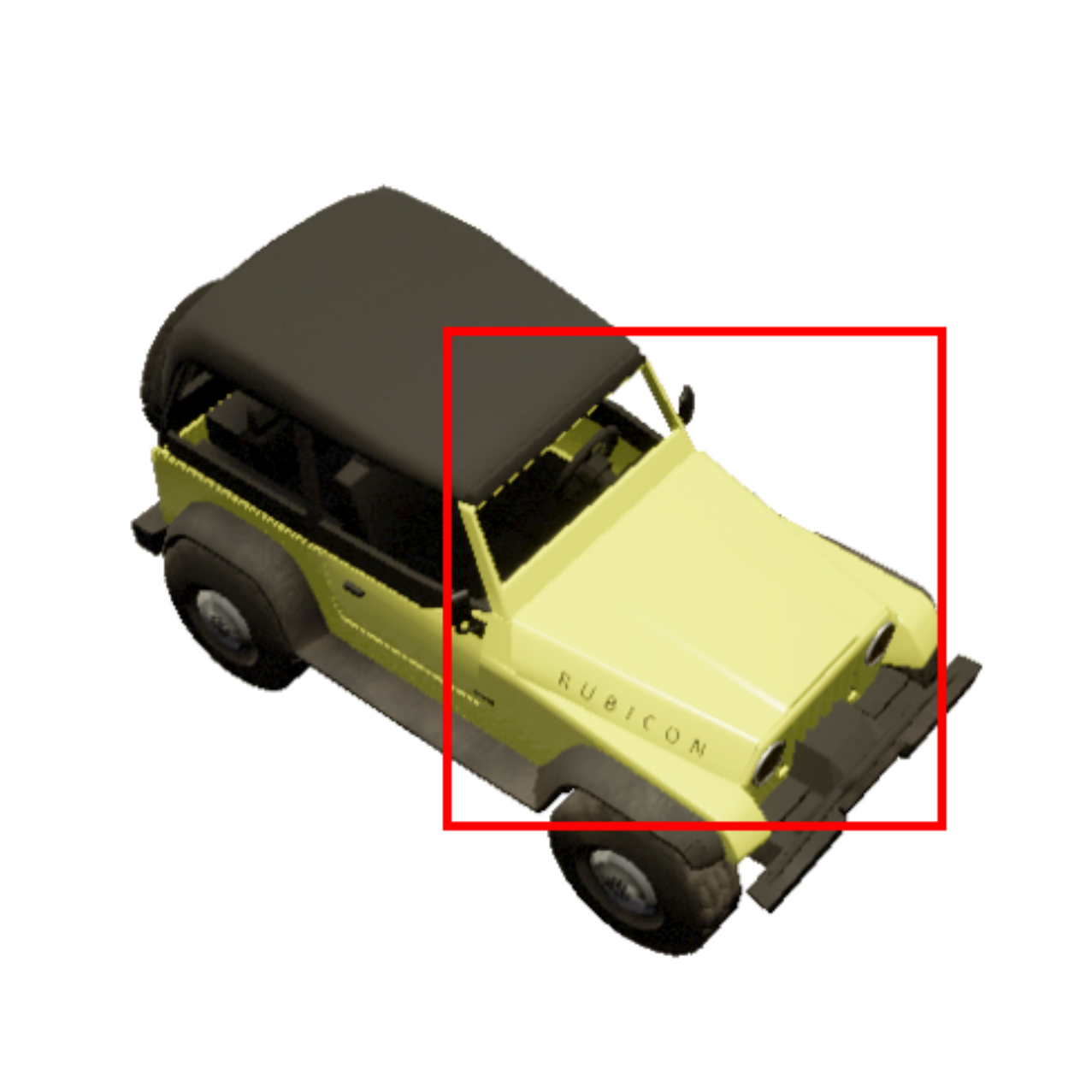}}\hfill}
  \subfloat[][]{
  {\includegraphics[width=0.18\linewidth]{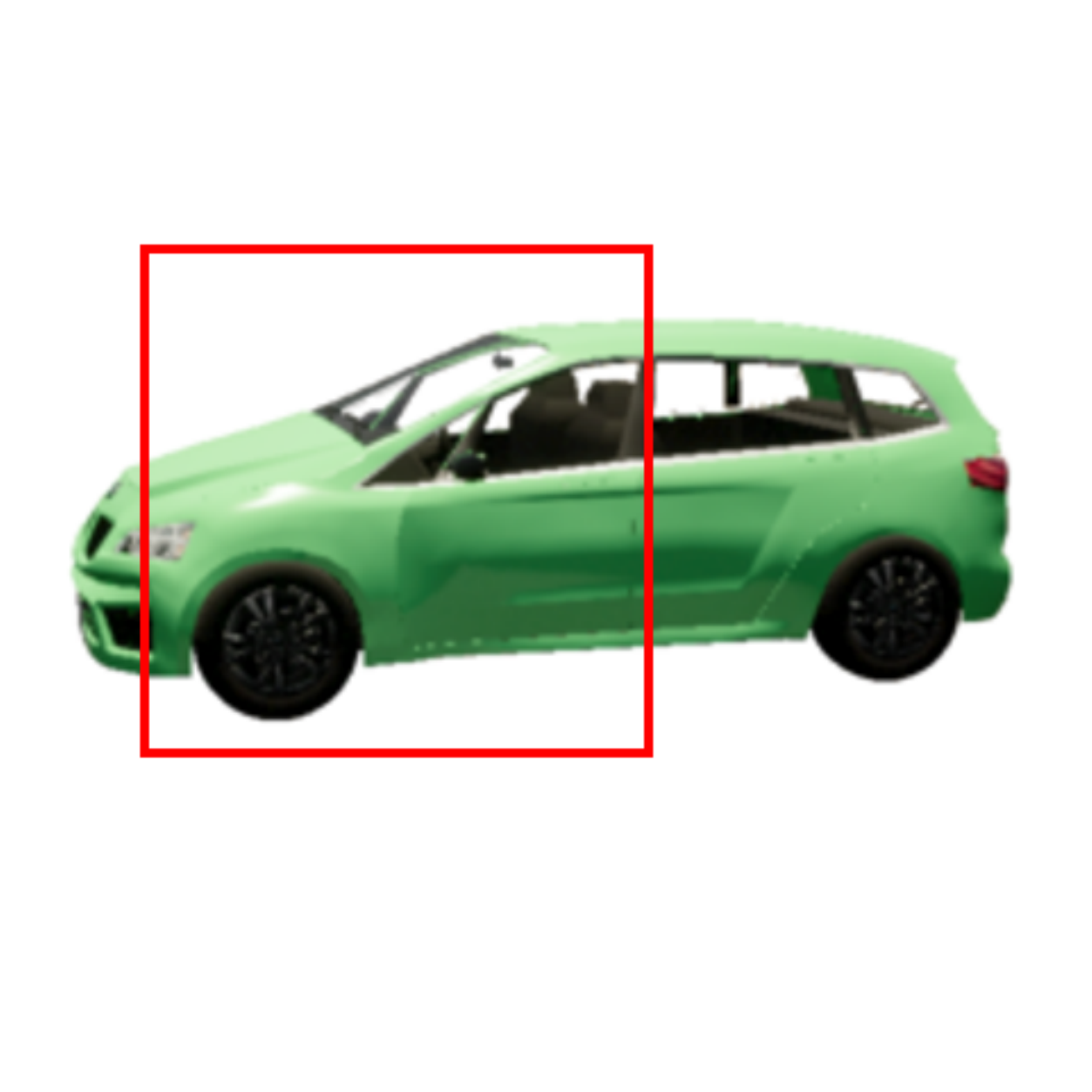}}\hfill}
  \subfloat[][]{
  {\includegraphics[width=0.18\linewidth]{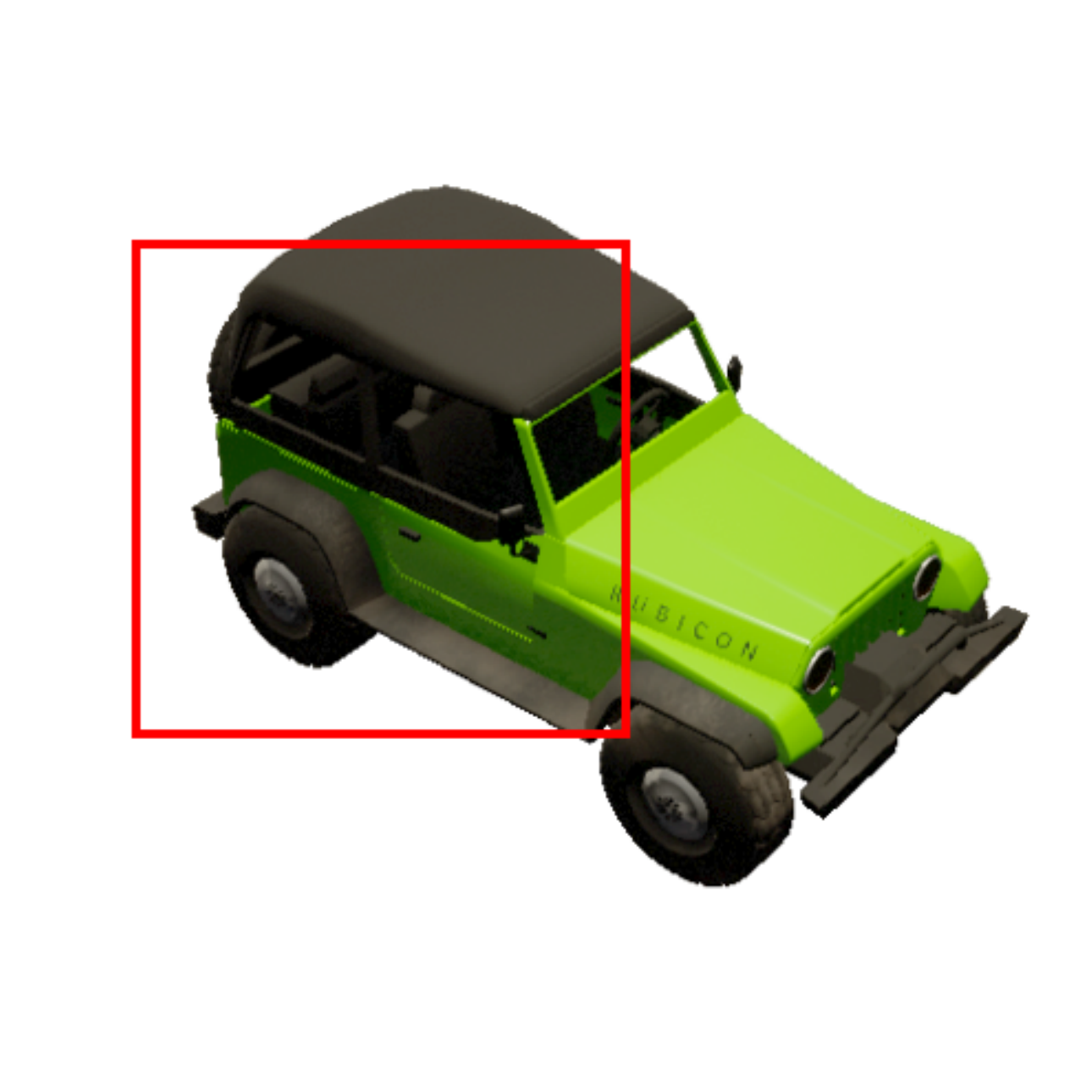}}\hfill}
  \subfloat[][]{
  {\includegraphics[width=0.18\linewidth]{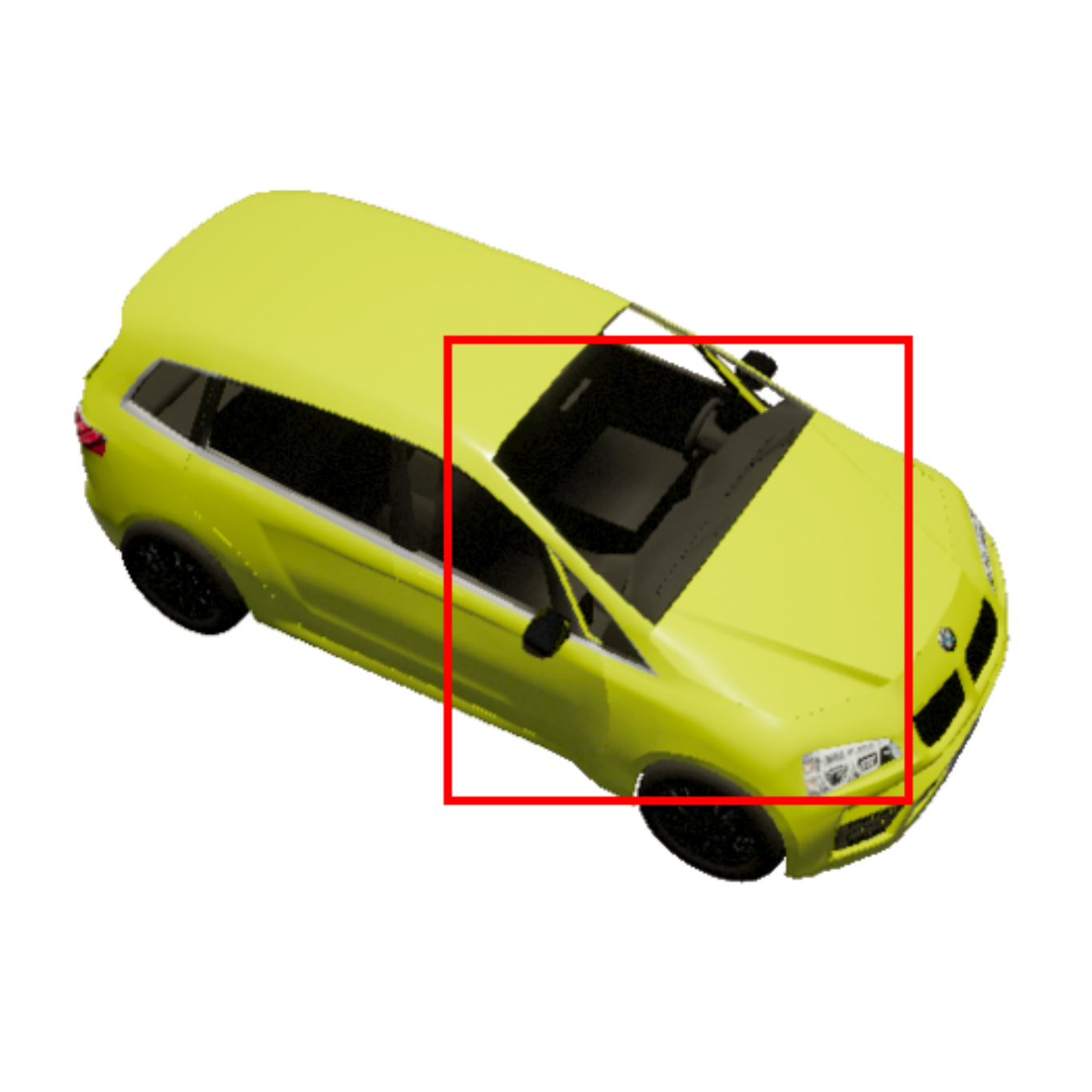}}\hfill}
  \subfloat[][]{
  {\includegraphics[width=0.18\linewidth]{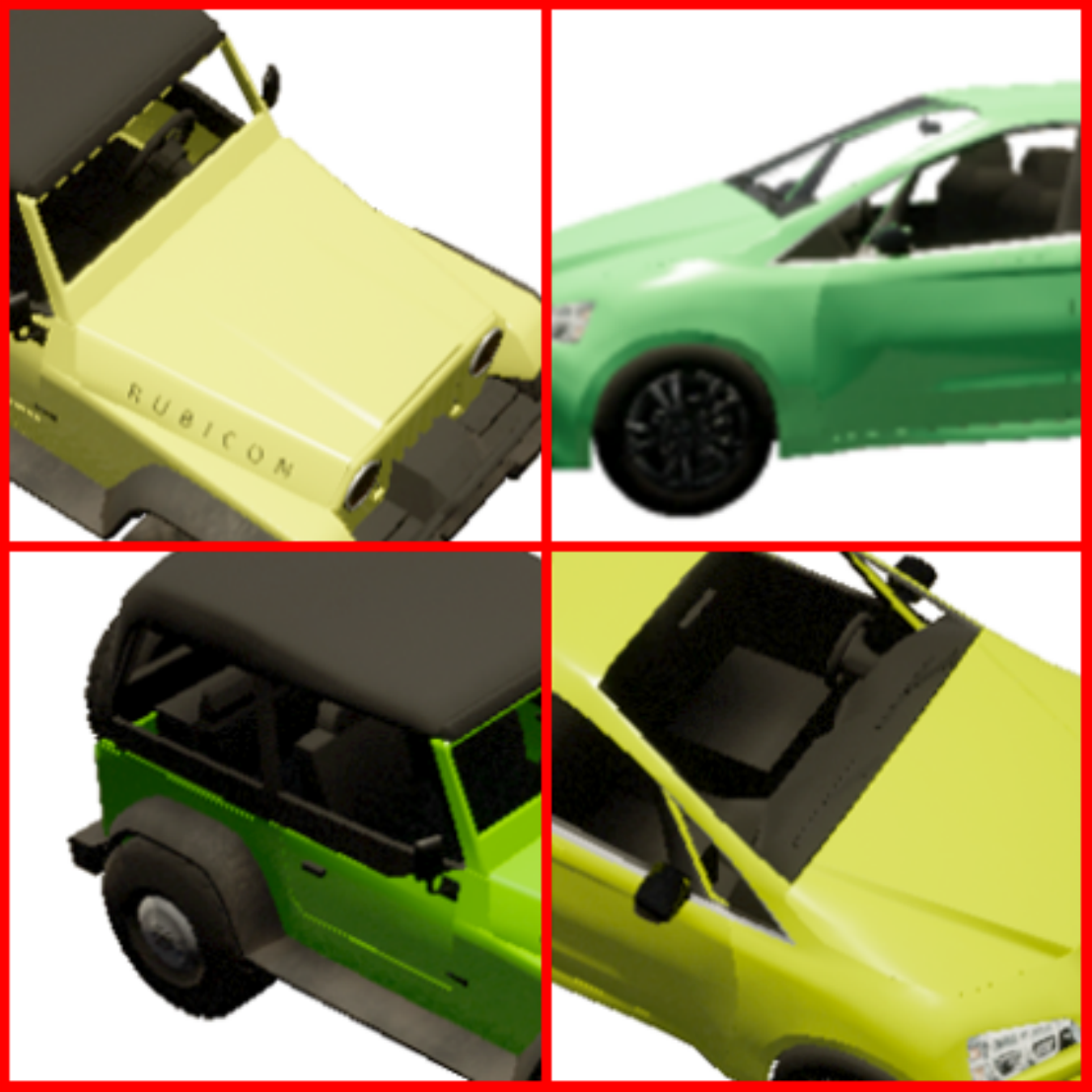}}\hfill}\\
\vspace{-5pt}
   \caption{\textbf{Intuitions of swapped-attribute classification loss:} 
   (a), (b) input images, generated images by (c) appearance swap and (d) shape swap between (a) and (b), and (e) enlarged extracted patches. We present a swapped-attribute classification loss that classifies a set of patches from (a), (b), (c) and (a), (b), (d) to make the networks be aware of which attribute swap, which in turn helps to boost the disentanglement.} 
  \label{Fig:disentangle-pose,shape,appearance}
\end{figure}

\subsection{Stage-Wise Training}
\label{training_strategy}
Training the proposed auto-encoder model end-to-end from a scratch is extremely challenging due to inherent ambiguity of disentangled 3D attributes.
To overcome this, we propose a novel stage-wise training scheme tailored for our auto-encoder architecture, described in the following.

\subsubsection{Stage 1: Pre-training Decoder $\mathcal{F}$.}
We first train the decoder $\mathcal{F}$. Before training an image-conditioned model, we first sample random latent vectors for appearance and shape codes, $\mathbf{z}_a$ and $\mathbf{z}_s$, from a Gaussian distribution such that $\mathbf{z}_a,\mathbf{z}_s \sim \mathcal{N}(\mathbf{0}, \mathbf{I})$ and sample camera pose parameters $\pi$ from uniform distribution on the upper hemisphere of neural radiance field, and then learn a volume renderer that renders a patch image $I_{\mathrm{patch}}$ from $\mathbf{z}_a$, $\mathbf{z}_s$ and $\pi$. Note that the patch center $\mathbf{u} = (u, v) \in \mathbb{R}^2$ and the scale $s \in \mathbb{R}+$ of the virtual $K \times K$ patch are randomly sampled from a uniform distribution. The total loss for stage 1 training is defined as
\begin{equation}
    \mathcal{L}_\mathrm{stage-1} =  \mathcal{L}_\mathrm{GAN}(I,p).
\end{equation}
Since this loss function is learned with random latent codes $\mathbf{z}_a$, $\mathbf{z}_s$, and camera $\pi$, the perturbation $p$ is inherently considered. 
After training, we render an image with learned $\mathcal{F}$ and use $\{\mathbf{z}_a, \mathbf{z}_s, \pi, I_\mathrm{pseudo}\}$ as \textit{pseudo} pairs for next stages.

\subsubsection{Stage 2: Pre-training Encoder $\mathcal{E}$.}
After stage 1, we now have the pre-trained decoder $\mathcal{F}$, and \textit{pseudo} pairs $\{\mathbf{z}_a, \mathbf{z}_s, \pi, I_\mathrm{pseudo}\}$. 
In stage 2, we train the encoder $\mathcal{E}$ that extracts shape, appearance codes and camera pose from an image given $\{\mathbf{z}_a, \mathbf{z}_s, \pi, I\}$. In addition, we use the pre-trained decoder with its parameters fixed to better train the encoder. The total loss for stage 2 is defined as follows:
\begin{equation}
    \begin{split}
        \mathcal{L}_\mathrm{stage-2} =  &\lambda_\mathrm{render}\mathcal{L}_\mathrm{render}(I_{\mathrm{pseudo}})  + \lambda_\mathrm{percep}\mathcal{L}_\mathrm{percep}(I_{\mathrm{pseudo}}) \\ &+  \lambda_\mathrm{pseudo}\mathcal{L}_\mathrm{pseudo} (\mathcal{E}(I_\mathrm{pseudo}),\mathbf{z}_a, \mathbf{z}_s, \pi),
    \end{split}
\end{equation}
where $\mathcal{L}_\mathrm{pseudo}$ is defined such that
\begin{equation}
    \begin{split}
        \mathcal{L}_\mathrm{pseudo}(\mathcal{E}(I_\mathrm{pseudo}),\mathbf{z}_a, \mathbf{z}_s, \pi) &= \|\mathbf{z}_a^{I_\mathrm{p}} - \mathbf{z}_a\|_{2} 
        + \|\mathbf{z}_s^{I_\mathrm{p}} - \mathbf{z}_s\|_{2} \\
        &+\mathcal{L}_{\mathrm{cam}}(\pi^{I_\mathrm{p}}, \pi),
    \end{split} 
\end{equation}
where $\{\mathbf{z}^{I_\mathrm{p}}_a,\mathbf{z}^{I_\mathrm{p}}_s,\pi^{I_\mathrm{p}}\}=\mathcal{E}(I_\mathrm{pseudo})$ and $\mathcal{L}_{\mathrm{cam}}(\pi^{I_\mathrm{p}},\pi) = \|\mathbf{s}^{I_\mathrm{p}} - \mathbf{s}\|_{2} + \|\mathbf{t}^{I_\mathrm{p}} - \mathbf{t}\|_{2} + \|(\mathbf{R}^{I_\mathrm{p}})^{-1}\mathbf{R} - \mathbf{I}_{3\times3}\|_{2}$, with $\mathbf{s}^{I_\mathrm{p}}$, $\mathbf{t}^{I_\mathrm{p}}$, and $\mathbf{R}^{I_\mathrm{p}}$ are the predicted camera parameters, respectively, with scale $\mathbf{s} \in \mathbb{R}^{1}$, translation $\mathbf{t} \in \mathbb{R}^{3}$, and rotation matrix $\mathbf{R} \in \mathbb{R}^{3 \times 3}$. We use an identity matrix $\mathbf{I}_{3\times3}$ for the rotation matrix loss. $\lambda_\mathrm{render}$, $\lambda_\mathrm{percep}$, and $\lambda_\mathrm{pseudo}$ are the weight parameters for corresponding losses.

\subsubsection{Stage 3: Fine-tuning Encoder $\mathcal{E}$.}
After stage 2, we now have the pre-trained encoder $\mathcal{E}$ and the decoder $\mathcal{F}$, which can be an auto-encoder themselves. Now we have the pre-trained encoder $\mathcal{E}$ and the decoder $\mathcal{F}$ after stage 2, which can be an auto-encoder themselves. However, the disentanglement performance still depends on the architectural designs of the decoder. In addition, the encoder $\mathcal{E}$ only takes a pseudo label, which is a generated image itself, and thus, it may be sub-optimal to handle a real image. To overcome these, in stage 3, we use all the proposed loss functions including global-local attribute consistency loss as $\mathcal{L}_\mathrm{GLAC}$ and swapped-attribute classification loss $\mathcal{L}_\mathrm{SwAC}$. 

The total loss for stage 3 training is defined as
\begin{equation}
    \begin{split}
        \mathcal{L}_\mathrm{stage-3} =& 
        \lambda_\mathrm{GAN}\mathcal{L}_\mathrm{GAN}(I) + \lambda_\mathrm{render}\mathcal{L}_\mathrm{render}(I) \\
        &+\lambda_\mathrm{percep}\mathcal{L}_\mathrm{percep}(I)  +\lambda_\mathrm{GLAC}\mathcal{L}_\mathrm{GLAC}(I,J) \\
        &+ \lambda_\mathrm{SwAC}\mathcal{L}_\mathrm{SwAC}(I,J).
    \end{split}
\end{equation}
where $\lambda_\mathrm{GLAC}$ and $\lambda_\mathrm{SwAC}$ are the weight parameters for corresponding losses.

\begin{figure*}[t!]
\vspace{10pt}
  \centering
  {\includegraphics[width=0.1\linewidth]{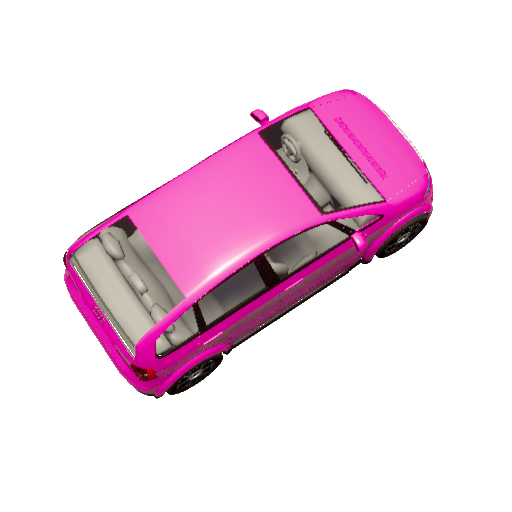}}\hfill \hspace{-2pt}
  {\includegraphics[width=0.1\linewidth]{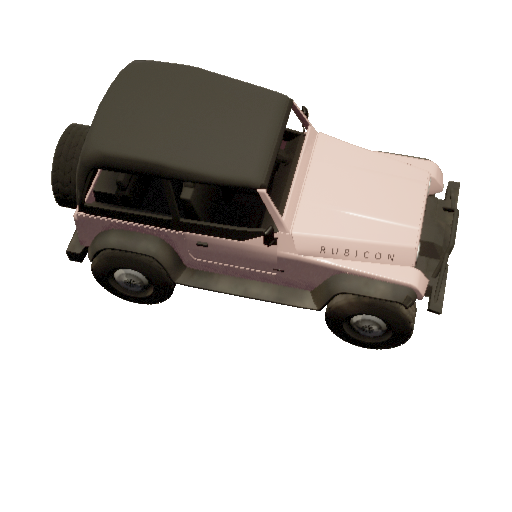}}\hfill \hspace{-2pt}
  {\includegraphics[width=0.1\linewidth]{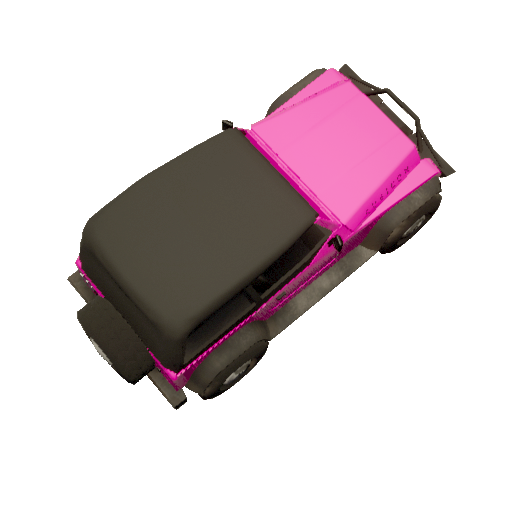}}\hfill \hspace{7pt}
  {\includegraphics[width=0.1\linewidth]{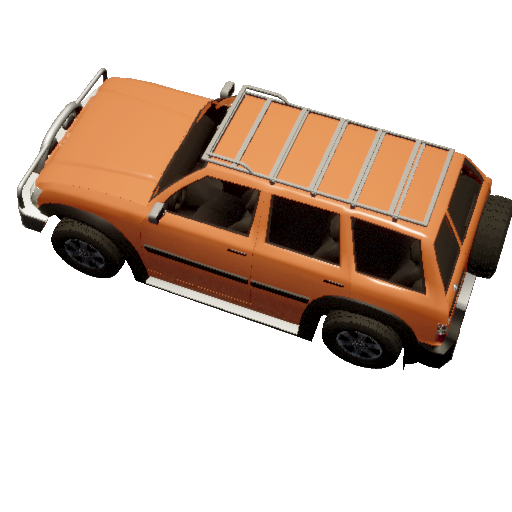}}\hfill \hspace{-2pt}
  {\includegraphics[width=0.1\linewidth]{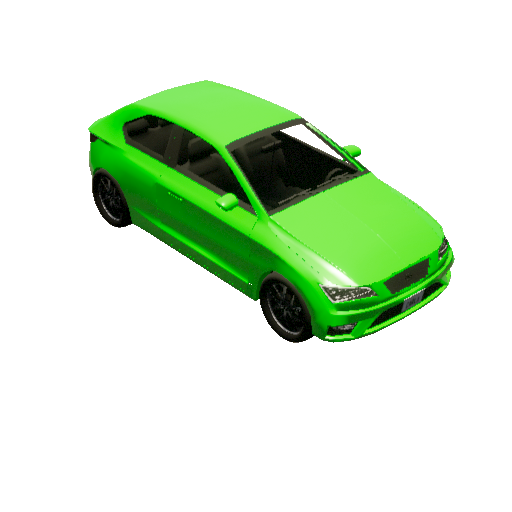}}\hfill \hspace{-2pt}
  {\includegraphics[width=0.1\linewidth]{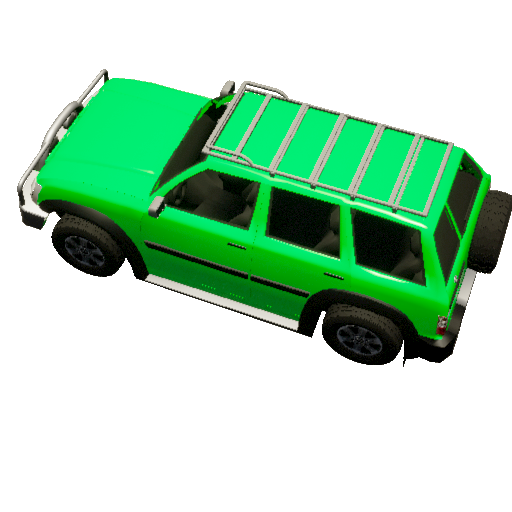}}\hfill \hspace{7pt}
  {\includegraphics[width=0.1\linewidth]{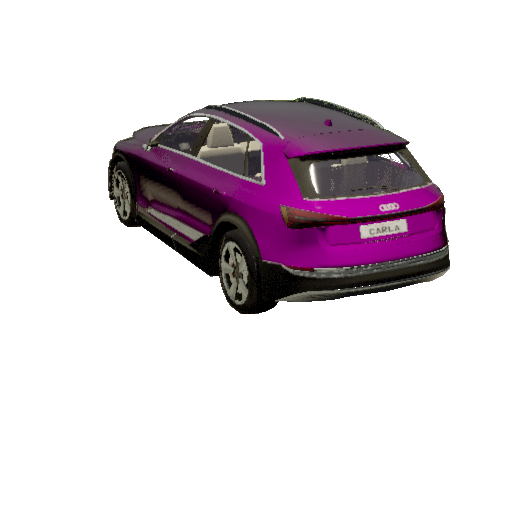}}\hfill \hspace{-2pt}
  {\includegraphics[width=0.1\linewidth]{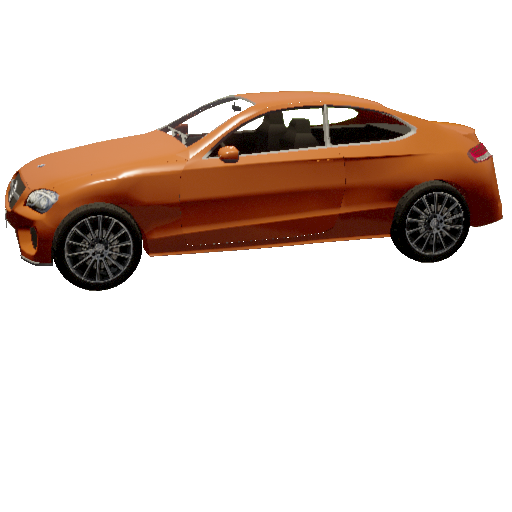}}\hfill \hspace{-2pt}
  {\includegraphics[width=0.1\linewidth]{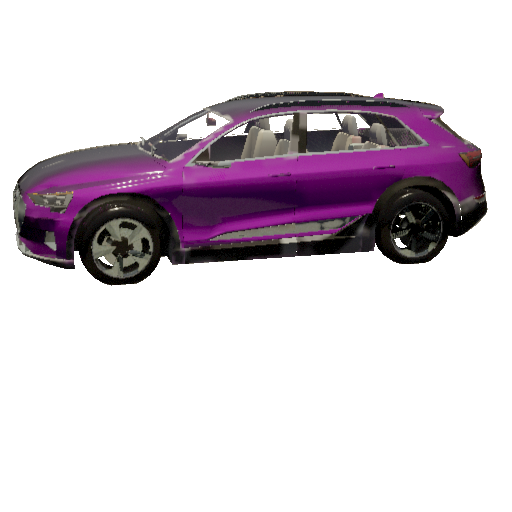}}\hfill\\
  {\includegraphics[width=0.1\linewidth]{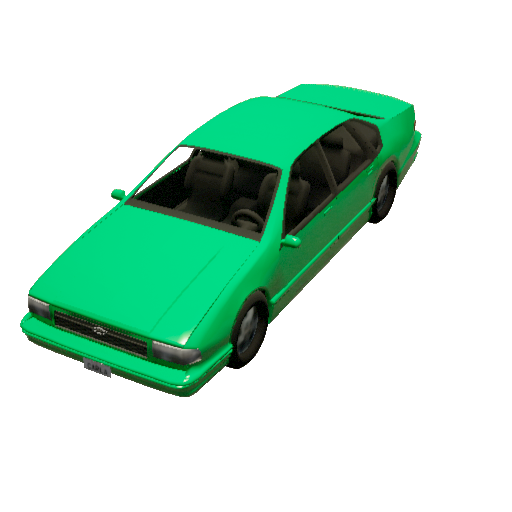}}\hfill \hspace{-2pt}
  {\includegraphics[width=0.1\linewidth]{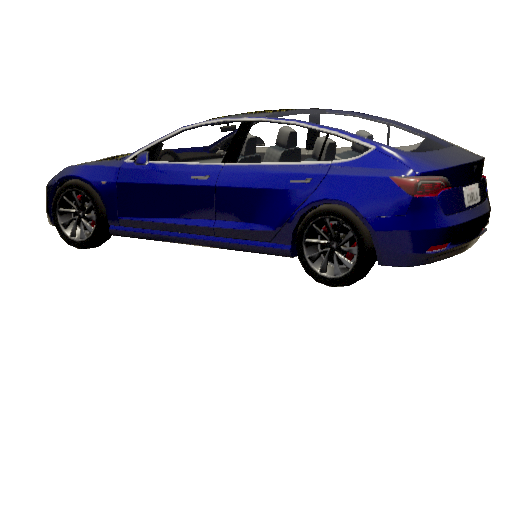}}\hfill \hspace{-2pt}
  {\includegraphics[width=0.1\linewidth]{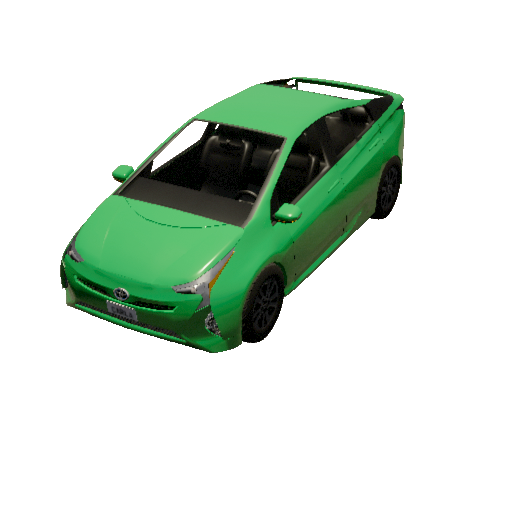}}\hfill \hspace{7pt}
  {\includegraphics[width=0.1\linewidth]{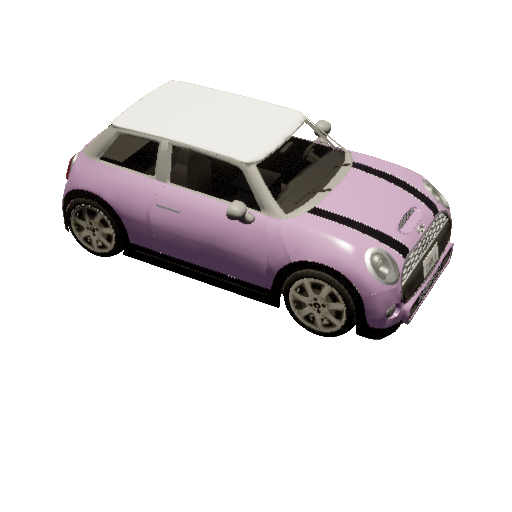}}\hfill \hspace{-2pt}
  {\includegraphics[width=0.1\linewidth]{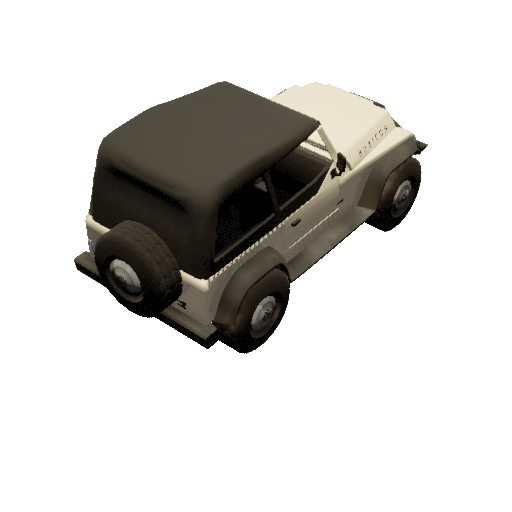}}\hfill \hspace{-2pt}
  {\includegraphics[width=0.1\linewidth]{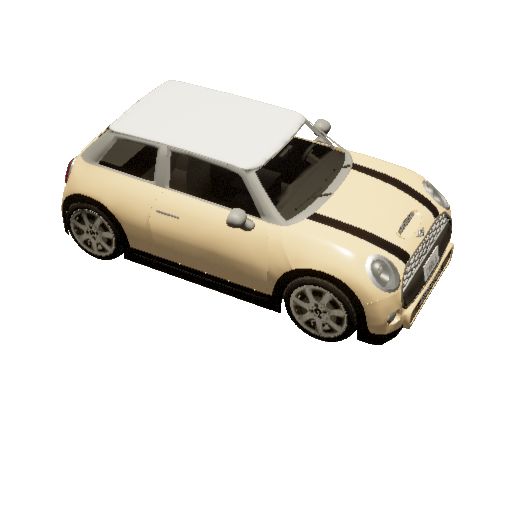}}\hfill \hspace{7pt}
  {\includegraphics[width=0.1\linewidth]{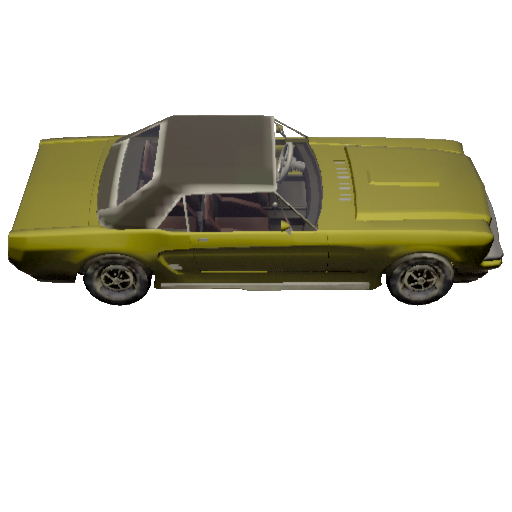}}\hfill \hspace{-2pt}
  {\includegraphics[width=0.1\linewidth]{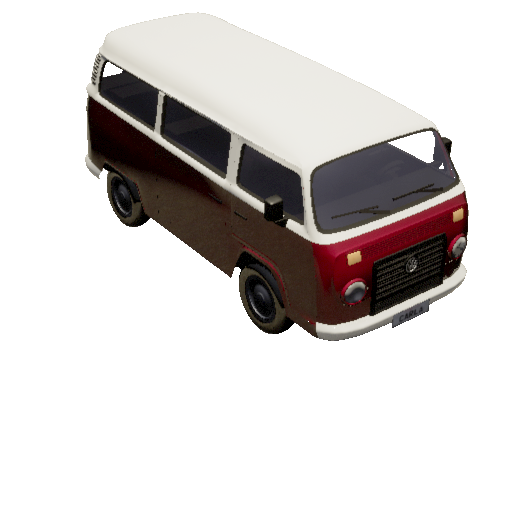}}\hfill \hspace{-2pt}
  {\includegraphics[width=0.1\linewidth]{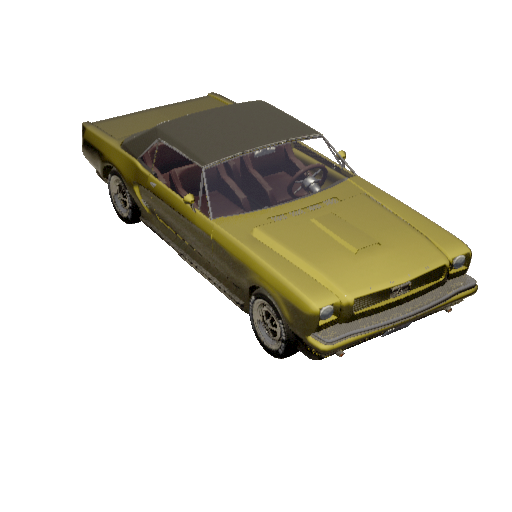}}\hfill\\
\caption{\textbf{Examples of proposed dataset for measuring disentanglement quality.} Given pairs of images (col. 1-2, 4-5, 7-8), we set the image with swapped attribute as a ground-truth image for attribute-swap; shape-swapped (col. 3), appearance-swapped (col. 6), and camera-swapped (col. 9).}
\label{proposed_dataset}
\end{figure*}

\section{Experiments}
\label{Experiments}
\subsection{Implementation Details} 
Our decoder architecture follows the default setting for GRAF~\cite{Schwarz2020graf}. To extract feature from a given input, we use a ResNet-18 model~\cite{he2016deep} pre-trained on ImageNet. We dropped the last fully connected layer and instead attached 2 fully connected layers followed by shape, appearance, and camera pose estimators. For the shared 2 layers we use batch normalization and ReLU activation and its final activation as LeakyReLU. We use a PatchGAN discriminator, following Pix2Pix~\cite{isola2017image}. We sample 1,024 rays within the bounding box in an image. We use CNNs with additional fully connected layers for a swapped-attribute binary classifier. The encoder predict 6 dimensional vector for camera pose regression, following the representation of 6D rotation~\cite{Zhou_2019_CVPR}. Camera translation vector is extracted from the rotation matrix, identical to the implementation of GRAF~\cite{Schwarz2020graf}.

\begin{figure*}[!t]
  \centering
  \renewcommand{\thesubfigure}{}
\begin{tabular}{M{0.06\linewidth} M{0.9\linewidth}}
& 
{\includegraphics[width=0.11\linewidth]{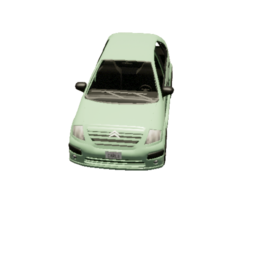}}\hfill
{\includegraphics[width=0.11\linewidth]{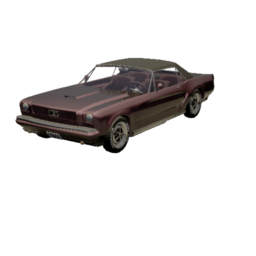}}\hfill
{\includegraphics[width=0.11\linewidth]{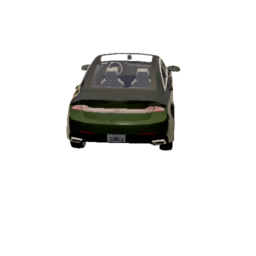}}\hfill 
{\includegraphics[width=0.11\linewidth]{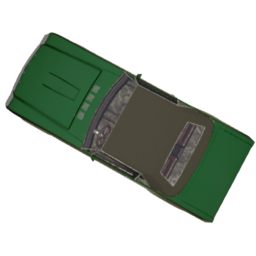}}\hfill 
{\includegraphics[width=0.11\linewidth]{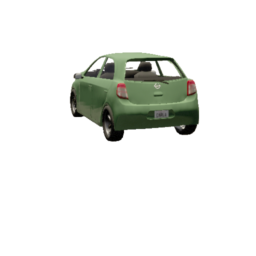}}\hfill 
{\includegraphics[width=0.11\linewidth]{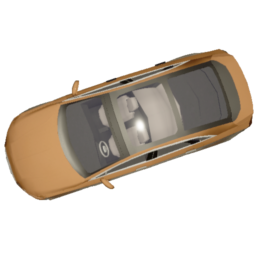}}\hfill 
{\includegraphics[width=0.11\linewidth]{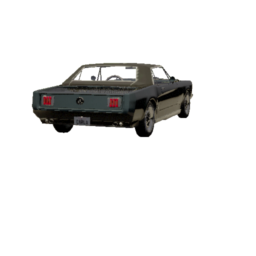}}\hfill 
{\includegraphics[width=0.11\linewidth]{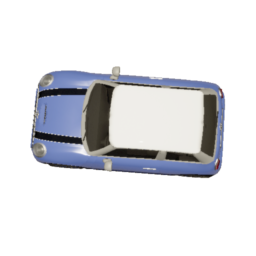}}\hfill \\

\rotatebox{90}{\makecell{SA}} &
{\includegraphics[width=0.11\linewidth]{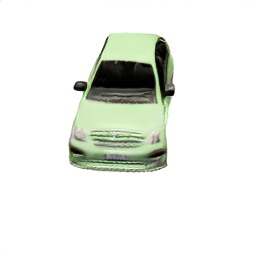}}\hfill
{\includegraphics[width=0.11\linewidth]{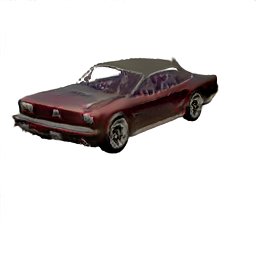}}\hfill
{\includegraphics[width=0.11\linewidth]{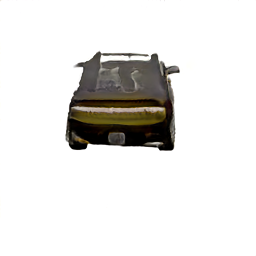}}\hfill 
{\includegraphics[width=0.11\linewidth]{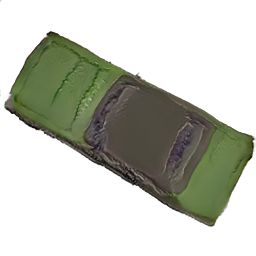}}\hfill
{\includegraphics[width=0.11\linewidth]{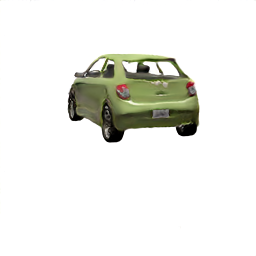}}\hfill
{\includegraphics[width=0.11\linewidth]{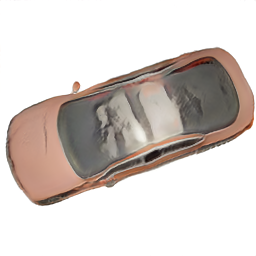}}\hfill 
{\includegraphics[width=0.11\linewidth]{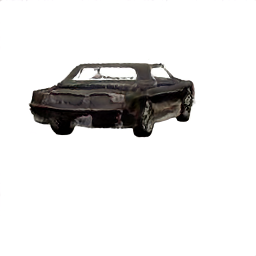}}\hfill
{\includegraphics[width=0.11\linewidth]{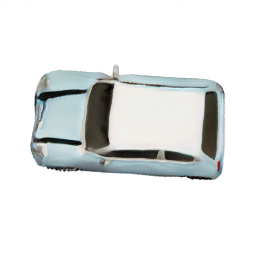}}\hfill\\
\rotatebox{90}{\makecell{Code \\ NeRF}} &
{\includegraphics[width=0.11\linewidth]{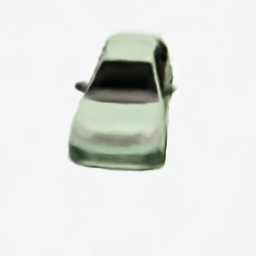}}\hfill 
{\includegraphics[width=0.11\linewidth]{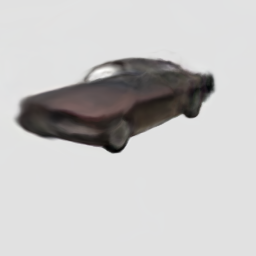}}\hfill
{\includegraphics[width=0.11\linewidth]{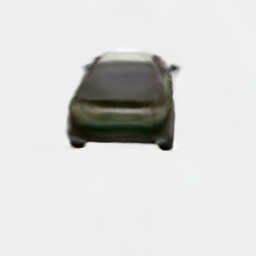}}\hfill
{\includegraphics[width=0.11\linewidth]{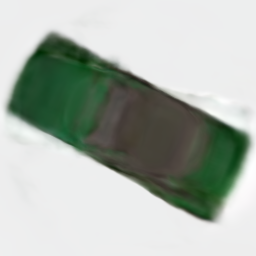}}\hfill
{\includegraphics[width=0.11\linewidth]{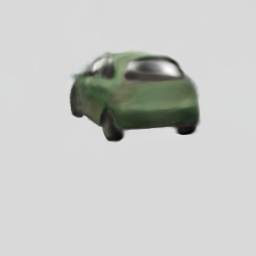}}\hfill
{\includegraphics[width=0.11\linewidth]{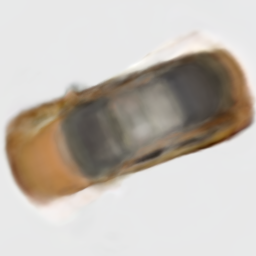}}\hfill 
{\includegraphics[width=0.11\linewidth]{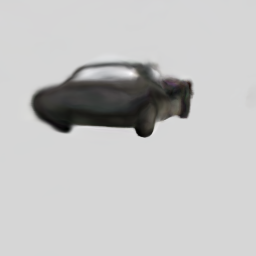}}\hfill
{\includegraphics[width=0.11\linewidth]{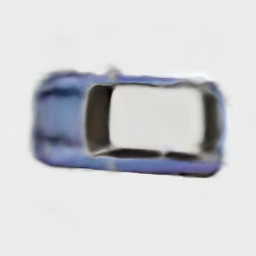}}\hfill
\\        
\rotatebox{90}{\makecell{Edit \\ NeRF}} &
{\includegraphics[width=0.11\linewidth]{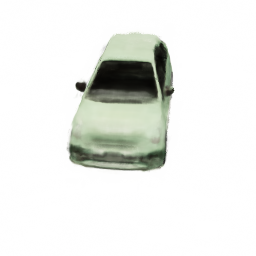}}\hfill
{\includegraphics[width=0.11\linewidth]{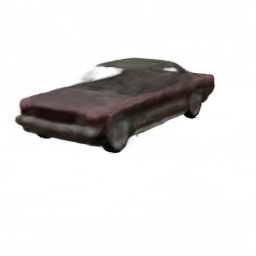}}\hfill
{\includegraphics[width=0.11\linewidth]{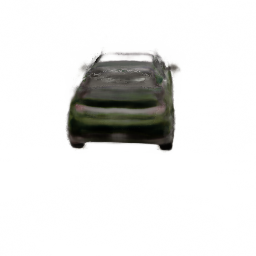}}\hfill 
{\includegraphics[width=0.11\linewidth]{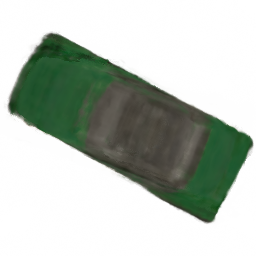}}\hfill
{\includegraphics[width=0.11\linewidth]{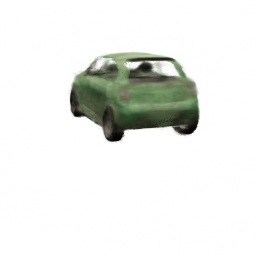}}\hfill
{\includegraphics[width=0.11\linewidth]{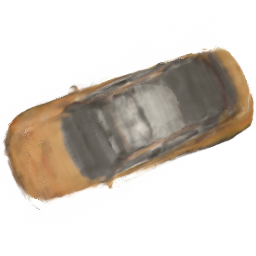}}\hfill 
{\includegraphics[width=0.11\linewidth]{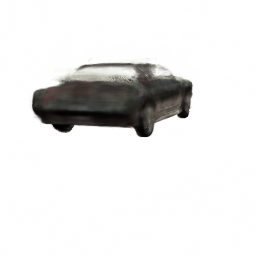}}\hfill
{\includegraphics[width=0.11\linewidth]{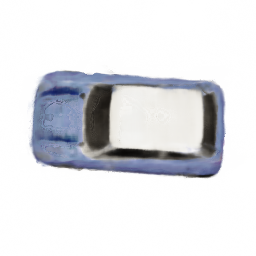}}\hfill\\

\rotatebox{90}{\makecell{AE- \\ NeRF}} &
{\includegraphics[width=0.11\linewidth]{recon/0001_gt_0001_gt.png}}\hfill
{\includegraphics[width=0.11\linewidth]{recon/0611_gt_0611_gt.png}}\hfill
{\includegraphics[width=0.11\linewidth]{recon/0068_gt_0068_gt.png}}\hfill 
{\includegraphics[width=0.11\linewidth]{recon/0830_gt_0830_gt.png}}\hfill
{\includegraphics[width=0.11\linewidth]{recon/0175_gt_0175_gt.png}}\hfill
{\includegraphics[width=0.11\linewidth]{recon/0780_gt_0780_gt.png}}\hfill 
{\includegraphics[width=0.11\linewidth]{recon/0221_gt_0221_gt.png}}\hfill
{\includegraphics[width=0.11\linewidth]{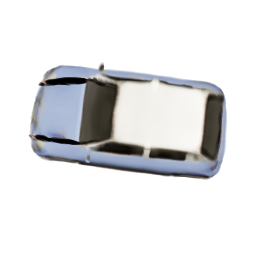}}\hfill \\
\end{tabular}
\caption{\textbf{Comparison of auto-encoding results} of AE-NeRF with SA~\cite{park2020swapping}, EditNeRF~\cite{liu2021editing}, and CodeNeRF~\cite{jang2021codenerf} on CARLA~\cite{dosovitskiy2017carla} benchmark.}\vspace{-5pt}
\label{recon_comparison_fig}
\end{figure*}

\begin{figure*}[!t]
\centering
\begin{tabular}{M{0.06\linewidth} M{0.9\linewidth}}
& 
{\includegraphics[width=0.11\linewidth]{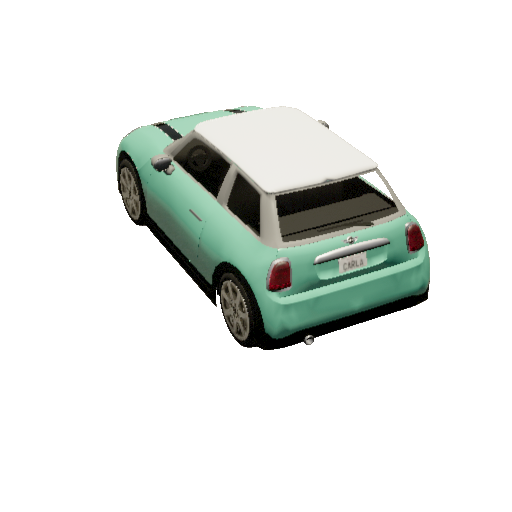}}\hfill
{\includegraphics[width=0.11\linewidth]{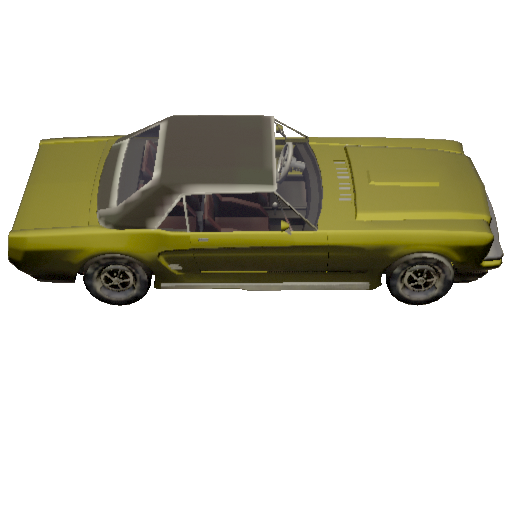}}\hfill 
{\includegraphics[width=0.11\linewidth]{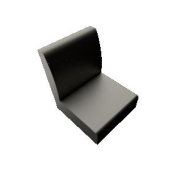}}\hfill 
{\includegraphics[width=0.11\linewidth]{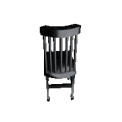}}\hfill 
{\includegraphics[width=0.11\linewidth]{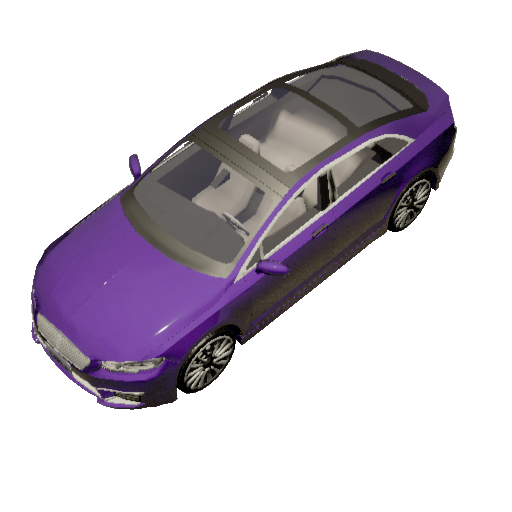}}\hfill 
{\includegraphics[width=0.11\linewidth]{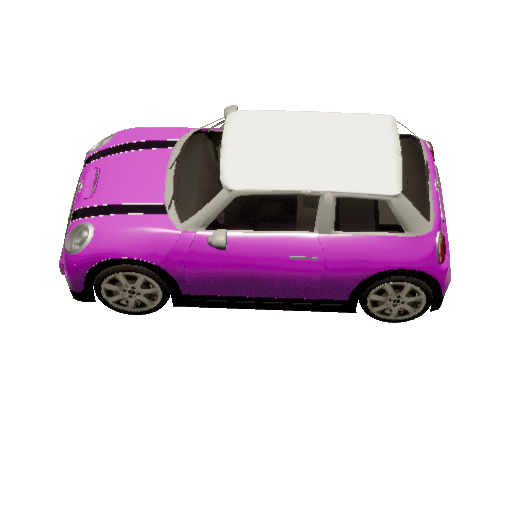}}\hfill 
{\includegraphics[width=0.11\linewidth]{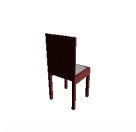}}\hfill 
{\includegraphics[width=0.11\linewidth]{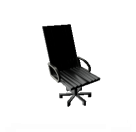}}\hfill 
\\

\rotatebox{90}{\makecell{Clip \\ NeRF}} &
{\includegraphics[width=0.11\linewidth]{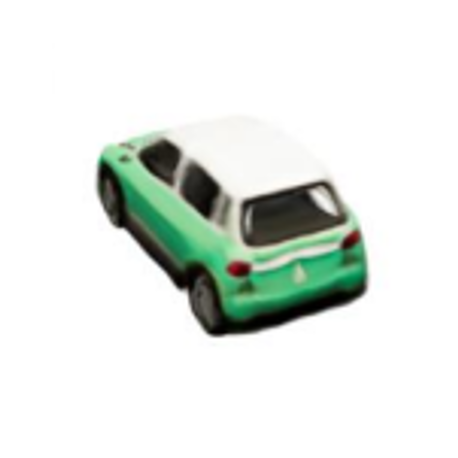}}\hfill
{\includegraphics[width=0.11\linewidth]{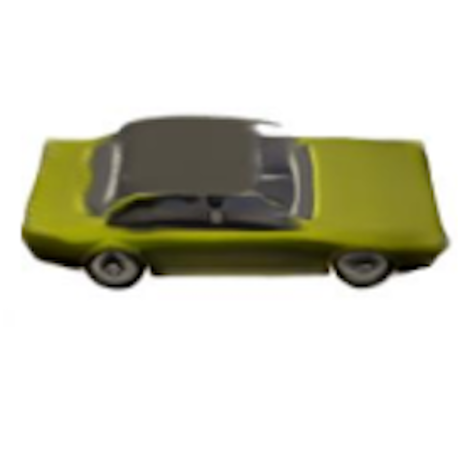}}\hfill
{\includegraphics[width=0.11\linewidth]{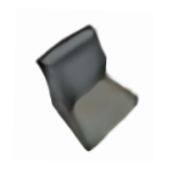}}\hfill
{\includegraphics[width=0.11\linewidth]{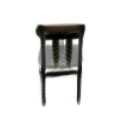}}\hfill
{\includegraphics[width=0.11\linewidth]{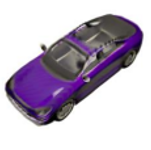}}\hfill
{\includegraphics[width=0.11\linewidth]{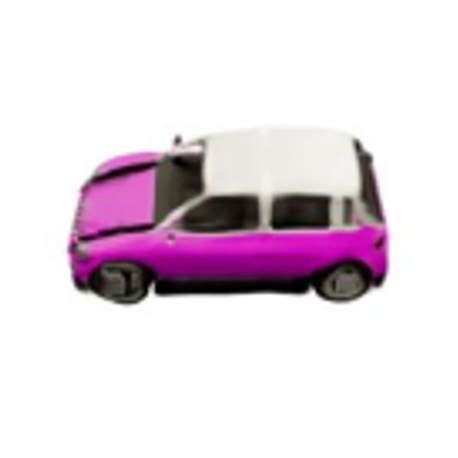}}\hfill
{\includegraphics[width=0.11\linewidth]{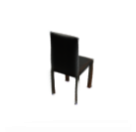}}\hfill
{\includegraphics[width=0.11\linewidth]{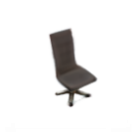}}\hfill\\

\rotatebox{90}{\makecell{AE- \\ NeRF}} &
{\includegraphics[width=0.11\linewidth]{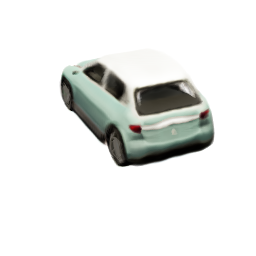}}\hfill
{\includegraphics[width=0.11\linewidth]{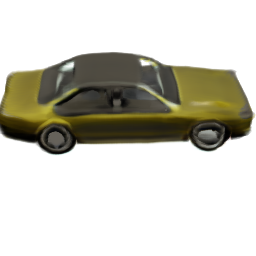}}\hfill
{\includegraphics[width=0.11\linewidth]{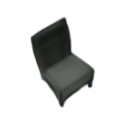}}\hfill 
\includegraphics[width=0.11\linewidth]{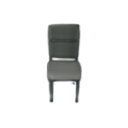}\hfill
{\includegraphics[width=0.11\linewidth]{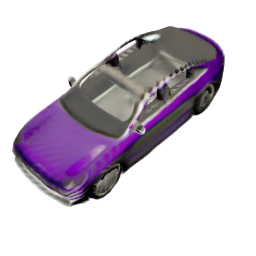}}\hfill
{\includegraphics[width=0.11\linewidth]{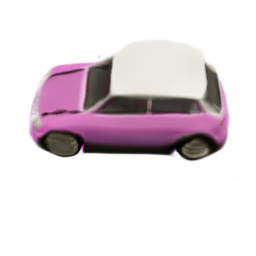}}\hfill 
{\includegraphics[width=0.11\linewidth]{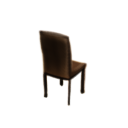}}\hfill
\includegraphics[width=0.11\linewidth]{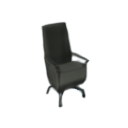}\hfill \\
\end{tabular}

\caption{\textbf{Auto-encoding results} of AE-NeRF compared with CLIP-NeRF~\cite{wang2021clip} on CARLA~\cite{dosovitskiy2017carla} and Photoshapes~\cite{park2018photoshapes} benchmark. }
\label{auto-encoding-chairs}
\end{figure*}
 
\begin{figure*}[!tp]
  \centering
  \renewcommand{\thesubfigure}{}
\begin{tabular}{M{0.06\textwidth} M{0.27\textwidth} M{0.27\textwidth} M{0.27\textwidth}}
& {\includegraphics[width=0.33\linewidth]{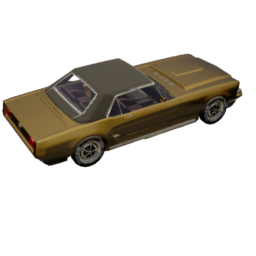}}\hfill &
{\includegraphics[width=0.33\linewidth]{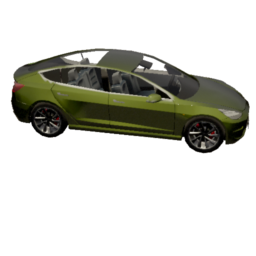}}\hfill &
{\includegraphics[width=0.33\linewidth]{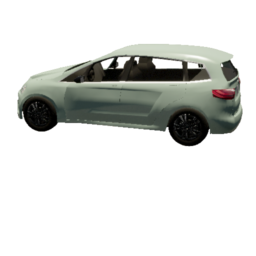}}\hfill \\

\rotatebox{90}{SA} &
{\includegraphics[width=0.33\linewidth]{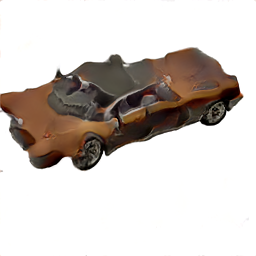}}\hfill
{\includegraphics[width=0.33\linewidth]{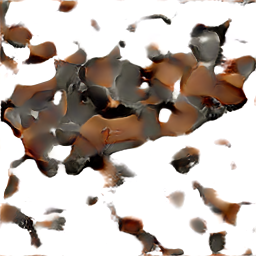}}\hfill
{\includegraphics[width=0.33\linewidth]{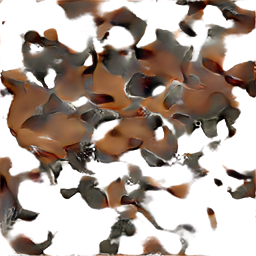}}\hfill &
{\includegraphics[width=0.33\linewidth]{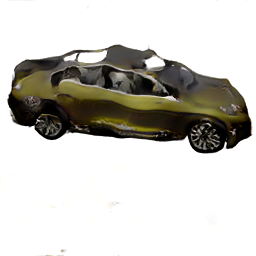}}\hfill
{\includegraphics[width=0.33\linewidth]{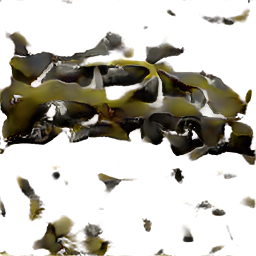}}\hfill
{\includegraphics[width=0.33\linewidth]{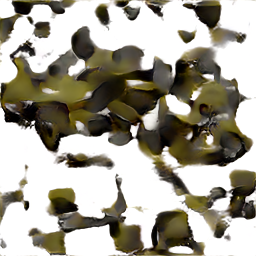}}\hfill &
{\includegraphics[width=0.33\linewidth]{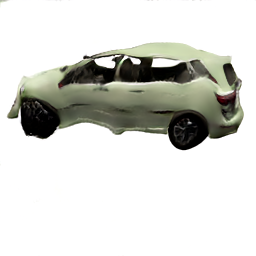}}\hfill
{\includegraphics[width=0.33\linewidth]{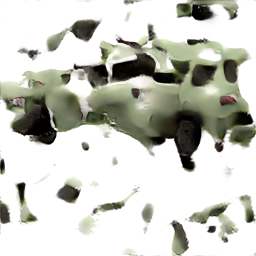}}\hfill
{\includegraphics[width=0.33\linewidth]{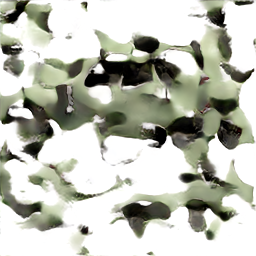}}\hfill  \\

\rotatebox{90}{\makecell{Code \\ NeRF}} &
{\includegraphics[width=0.33\linewidth]{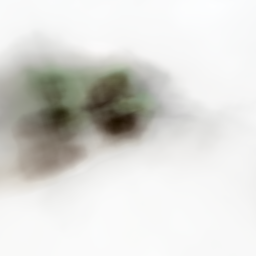}}\hfill
\includegraphics[width=0.33\linewidth]{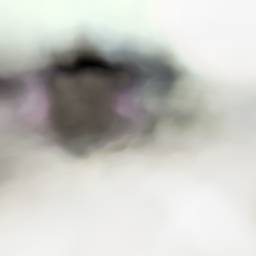}\hfill
{\includegraphics[width=0.33\linewidth]{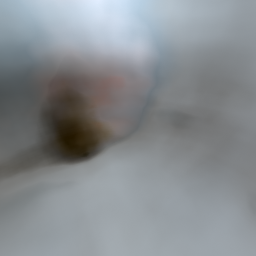}}\hfill &
\includegraphics[width=0.33\linewidth]{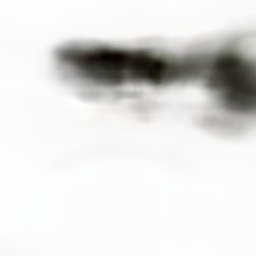}\hfill
{\includegraphics[width=0.33\linewidth]{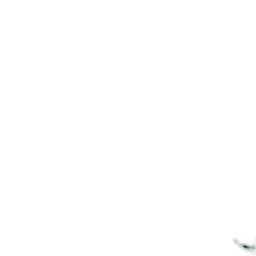}}\hfill
{\includegraphics[width=0.33\linewidth]{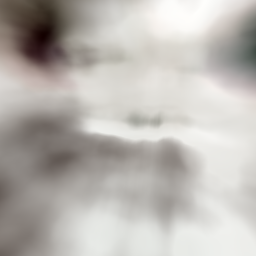}}\hfill &
{\includegraphics[width=0.33\linewidth]{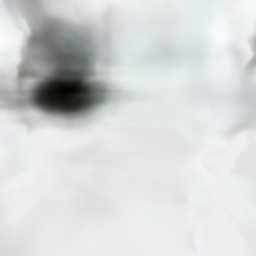}}\hfill
{\includegraphics[width=0.33\linewidth]{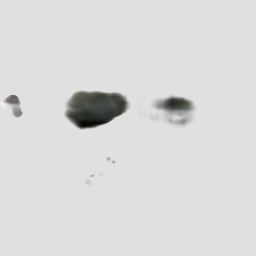}}\hfill
{\includegraphics[width=0.33\linewidth]{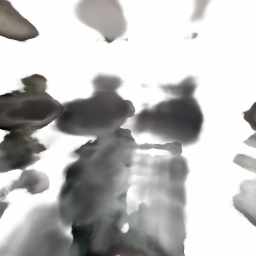}}\hfill \\
\rotatebox{90}{\makecell{Edit \\ NeRF}} &
{\includegraphics[width=0.33\linewidth]{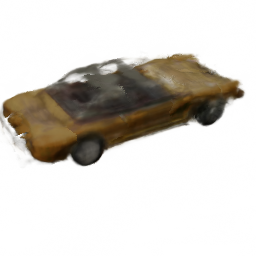}}\hfill
{\includegraphics[width=0.33\linewidth]{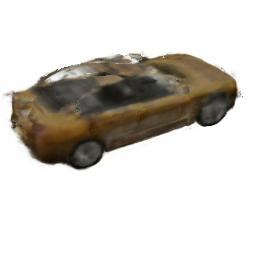}}\hfill
{\includegraphics[width=0.33\linewidth]{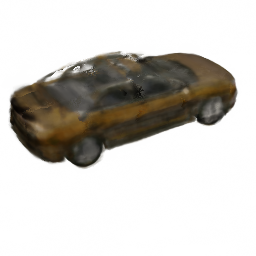}}\hfill &
{\includegraphics[width=0.33\linewidth]{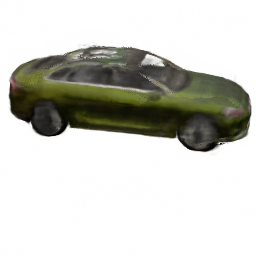}}\hfill
{\includegraphics[width=0.33\linewidth]{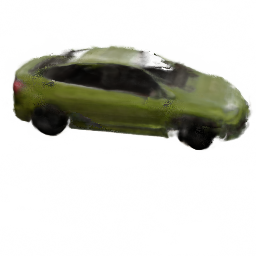}}\hfill
{\includegraphics[width=0.33\linewidth]{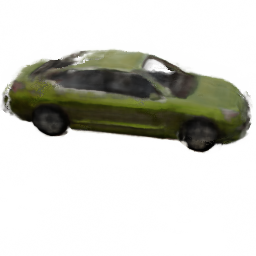}}\hfill &
{\includegraphics[width=0.33\linewidth]{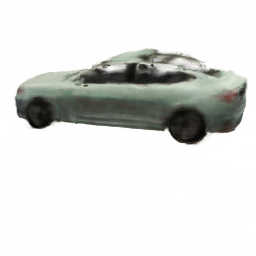}}\hfill
{\includegraphics[width=0.33\linewidth]{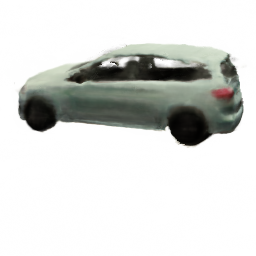}}\hfill
{\includegraphics[width=0.33\linewidth]{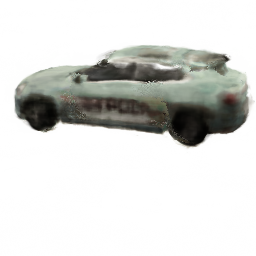}}\hfill \\

\rotatebox{90}{\makecell{AE- \\ NeRF}} &
{\includegraphics[width=0.33\linewidth]{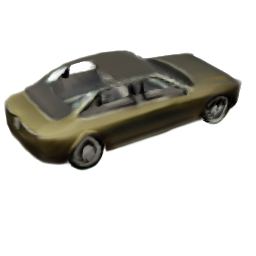}}\hfill
{\includegraphics[width=0.33\linewidth]{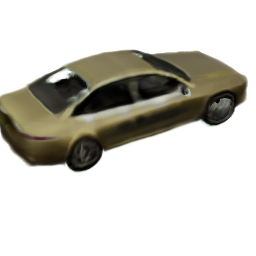}}\hfill
{\includegraphics[width=0.33\linewidth]{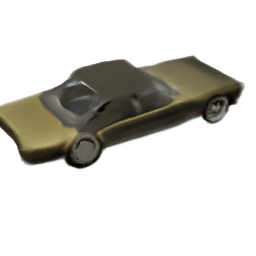}}\hfill &
{\includegraphics[width=0.33\linewidth]{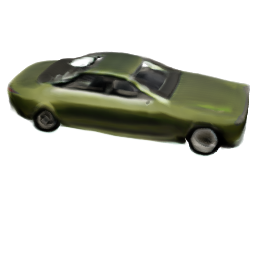}}\hfill
{\includegraphics[width=0.33\linewidth]{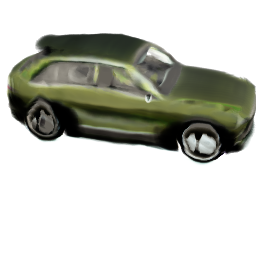}}\hfill
{\includegraphics[width=0.33\linewidth]{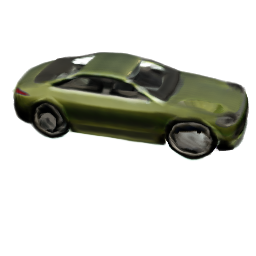}}\hfill &
{\includegraphics[width=0.33\linewidth]{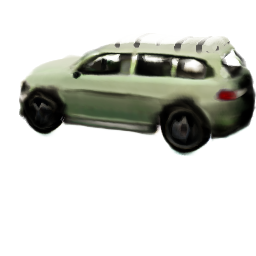}}\hfill
{\includegraphics[width=0.33\linewidth]{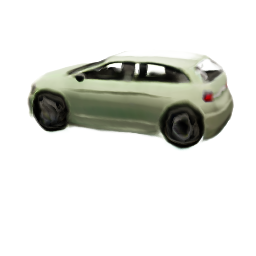}}\hfill
{\includegraphics[width=0.33\linewidth]{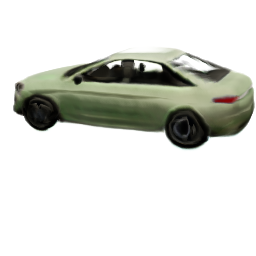}}\hfill \\
\end{tabular}

\vspace{10pt}
 
\begin{tabular}{M{0.06\textwidth} M{0.27\textwidth} M{0.27\textwidth} M{0.27\textwidth}}
\rotatebox{90}{SA} &
{\includegraphics[width=0.33\linewidth]{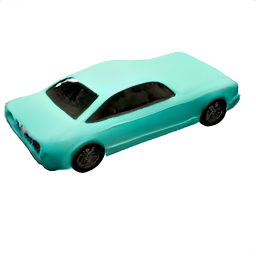}}\hfill
{\includegraphics[width=0.33\linewidth]{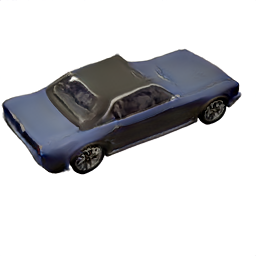}}\hfill
{\includegraphics[width=0.33\linewidth]{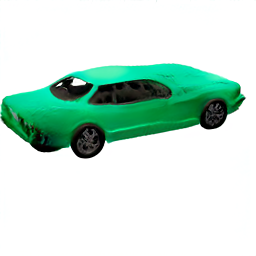}}\hfill &
{\includegraphics[width=0.33\linewidth]{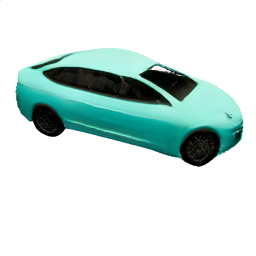}}\hfill
{\includegraphics[width=0.33\linewidth]{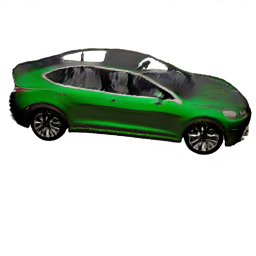}}\hfill
{\includegraphics[width=0.33\linewidth]{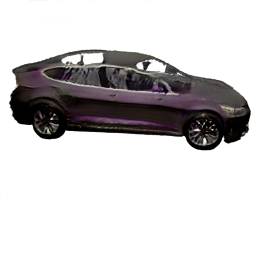}}\hfill &
{\includegraphics[width=0.33\linewidth]{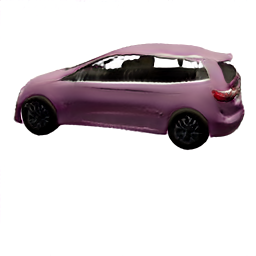}}\hfill
{\includegraphics[width=0.33\linewidth]{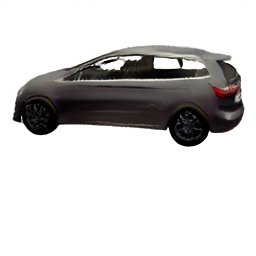}}\hfill
{\includegraphics[width=0.33\linewidth]{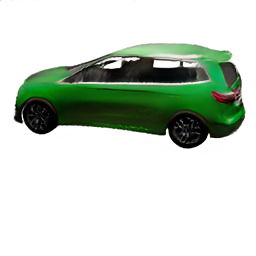}}\hfill \\

\rotatebox{90}{\makecell{Code \\ NeRF}} &
{\includegraphics[width=0.33\linewidth]{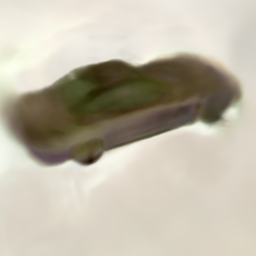}}\hfill
{\includegraphics[width=0.33\linewidth]{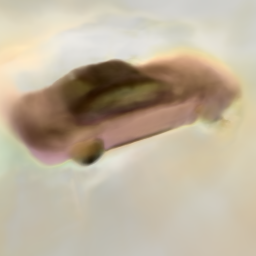}}\hfill
{\includegraphics[width=0.33\linewidth]{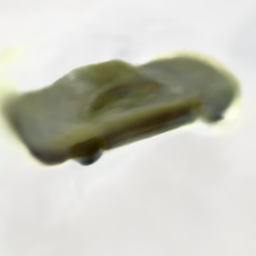}}\hfill &
{\includegraphics[width=0.33\linewidth]{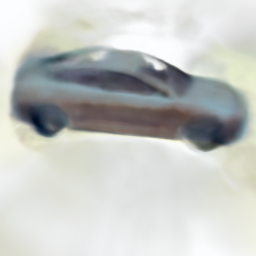}}\hfill
{\includegraphics[width=0.33\linewidth]{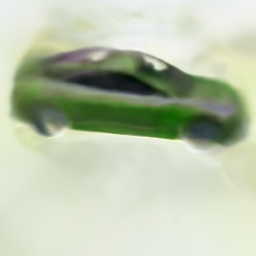}}\hfill
{\includegraphics[width=0.33\linewidth]{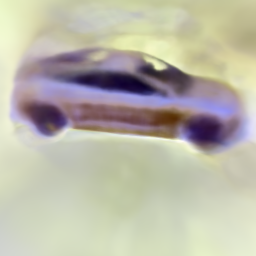}}\hfill &
{\includegraphics[width=0.33\linewidth]{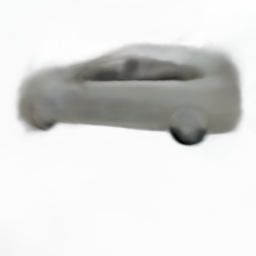}}\hfill
{\includegraphics[width=0.33\linewidth]{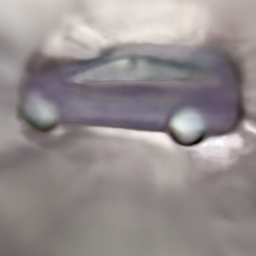}}\hfill
{\includegraphics[width=0.33\linewidth]{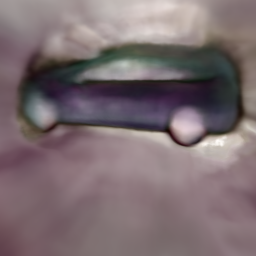}}\hfill \\
\rotatebox{90}{\makecell{Edit \\ NeRF}} &
{\includegraphics[width=0.33\linewidth]{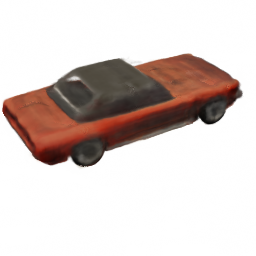}}\hfill
{\includegraphics[width=0.33\linewidth]{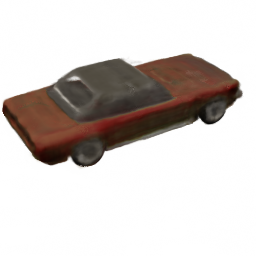}}\hfill
{\includegraphics[width=0.33\linewidth]{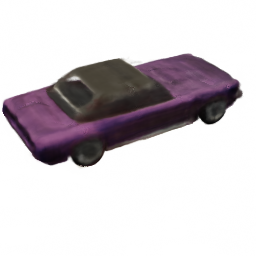}}\hfill &
{\includegraphics[width=0.33\linewidth]{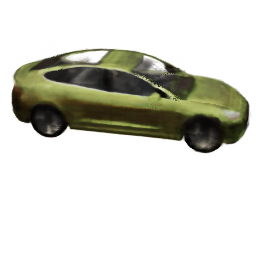}}\hfill
{\includegraphics[width=0.33\linewidth]{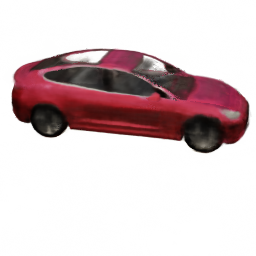}}\hfill
{\includegraphics[width=0.33\linewidth]{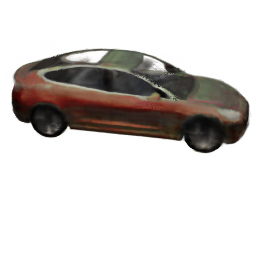}}\hfill &
{\includegraphics[width=0.33\linewidth]{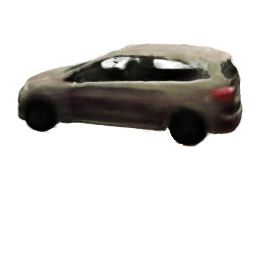}}\hfill
{\includegraphics[width=0.33\linewidth]{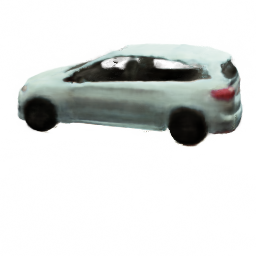}}\hfill
{\includegraphics[width=0.33\linewidth]{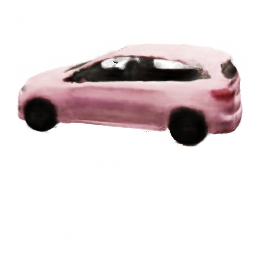}}\hfill \\

\rotatebox{90}{\makecell{AE- \\ NeRF}} &
{\includegraphics[width=0.33\linewidth]{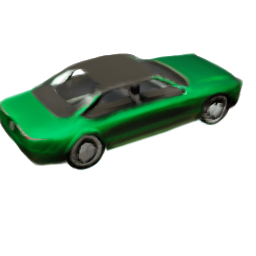}}\hfill
{\includegraphics[width=0.33\linewidth]{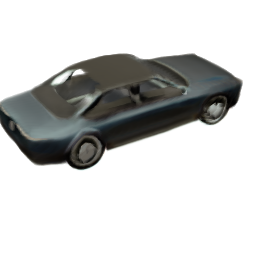}}\hfill
{\includegraphics[width=0.33\linewidth]{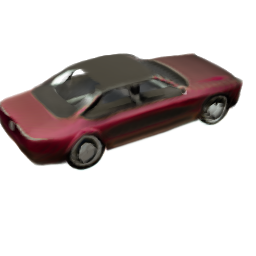}}\hfill &
{\includegraphics[width=0.33\linewidth]{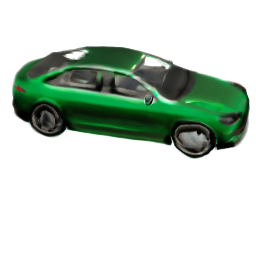}}\hfill
{\includegraphics[width=0.33\linewidth]{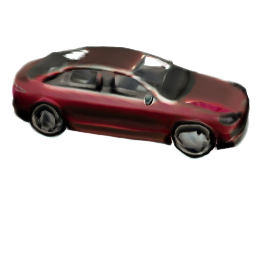}}\hfill
{\includegraphics[width=0.33\linewidth]{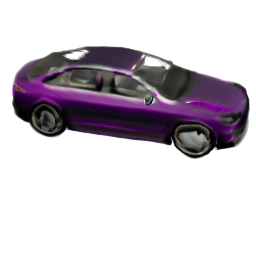}}\hfill &
{\includegraphics[width=0.33\linewidth]{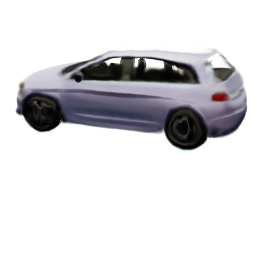}}\hfill
{\includegraphics[width=0.33\linewidth]{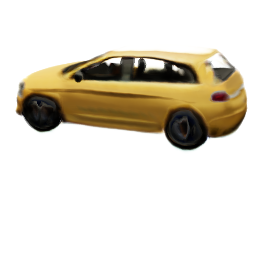}}\hfill
{\includegraphics[width=0.33\linewidth]{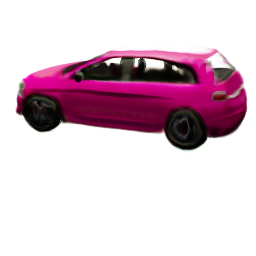}}\hfill \\
\end{tabular}

\vspace{10pt}

\begin{tabular}{M{0.06\textwidth} M{0.27\textwidth} M{0.27\textwidth} M{0.27\textwidth}}
\rotatebox{90}{\makecell{Code \\ NeRF}} &
{\includegraphics[width=0.33\linewidth]{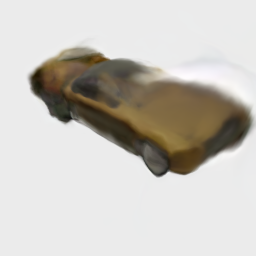}}\hfill
{\includegraphics[width=0.33\linewidth]{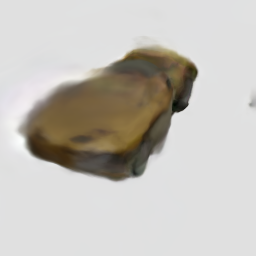}}\hfill
{\includegraphics[width=0.33\linewidth]{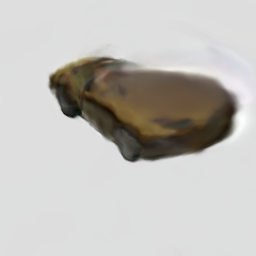}}\hfill &
{\includegraphics[width=0.33\linewidth]{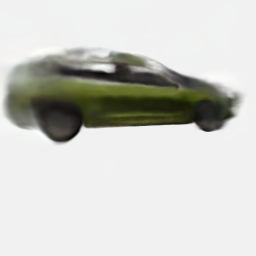}}\hfill
{\includegraphics[width=0.33\linewidth]{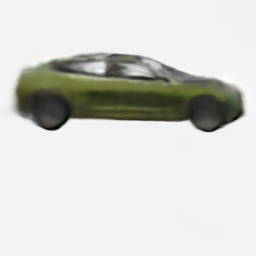}}\hfill
{\includegraphics[width=0.33\linewidth]{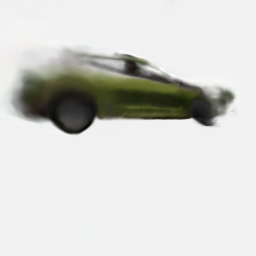}}\hfill &
{\includegraphics[width=0.33\linewidth]{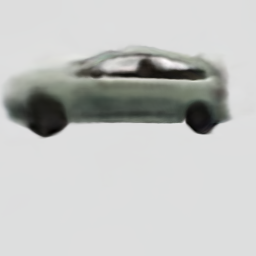}}\hfill
{\includegraphics[width=0.33\linewidth]{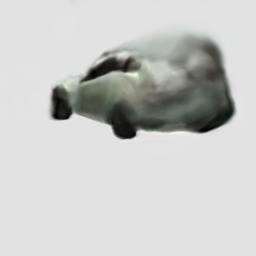}}\hfill
{\includegraphics[width=0.33\linewidth]{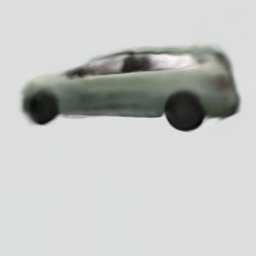}}\hfill \\
\rotatebox{90}{\makecell{Edit \\ NeRF}} &
{\includegraphics[width=0.33\linewidth]{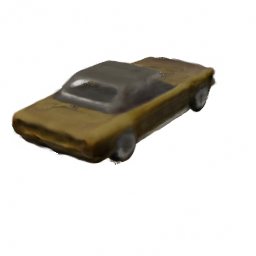}}\hfill
{\includegraphics[width=0.33\linewidth]{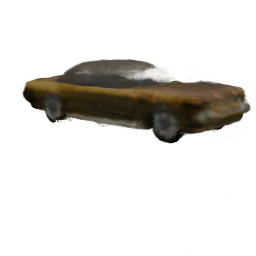}}\hfill
{\includegraphics[width=0.33\linewidth]{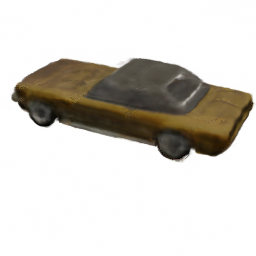}}\hfill &
{\includegraphics[width=0.33\linewidth]{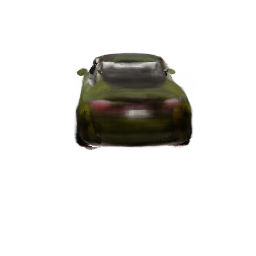}}\hfill
{\includegraphics[width=0.33\linewidth]{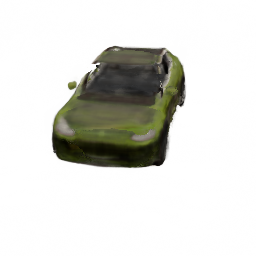}}\hfill
{\includegraphics[width=0.33\linewidth]{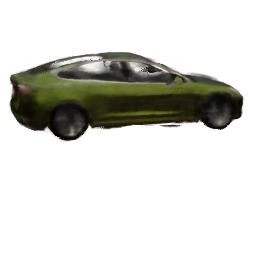}}\hfill &
{\includegraphics[width=0.33\linewidth]{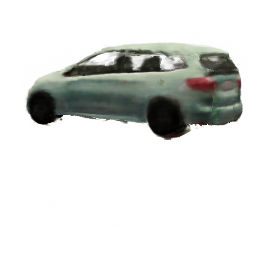}}\hfill
{\includegraphics[width=0.33\linewidth]{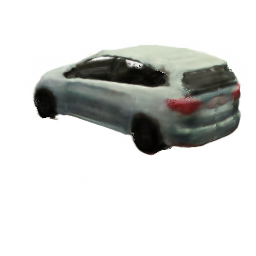}}\hfill
{\includegraphics[width=0.33\linewidth]{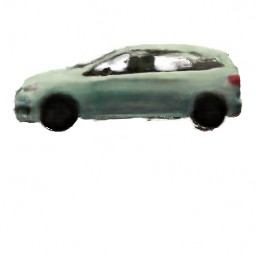}}\hfill \\

\rotatebox{90}{\makecell{AE- \\ NeRF}} &
{\includegraphics[width=0.33\linewidth]{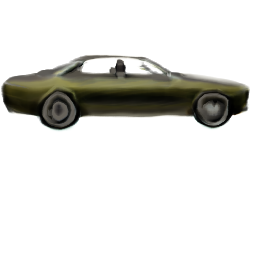}}\hfill
{\includegraphics[width=0.33\linewidth]{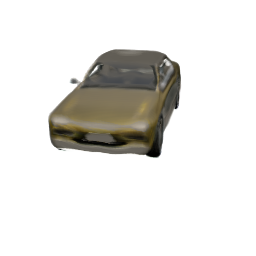}}\hfill
{\includegraphics[width=0.33\linewidth]{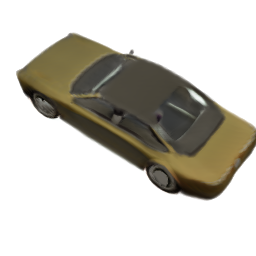}}\hfill &
{\includegraphics[width=0.33\linewidth]{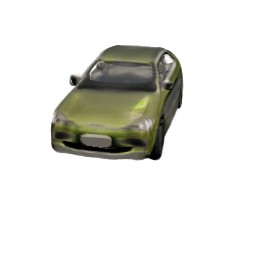}}\hfill
{\includegraphics[width=0.33\linewidth]{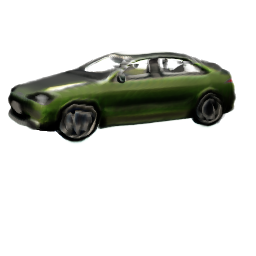}}\hfill
{\includegraphics[width=0.33\linewidth]{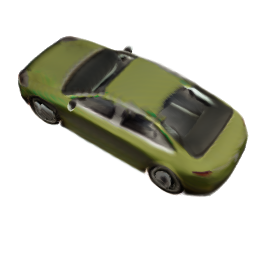}}\hfill &
{\includegraphics[width=0.33\linewidth]{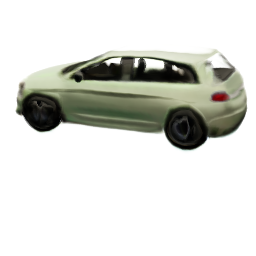}}\hfill
{\includegraphics[width=0.33\linewidth]{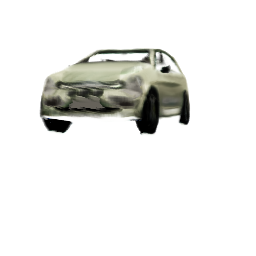}}\hfill
{\includegraphics[width=0.33\linewidth]{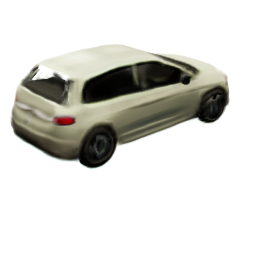}}\hfill \\
\end{tabular}
  
\caption{\textbf{Random perturbation comparison:} results of AE-NeRF with SA~\cite{park2020swapping}, EditNeRF~\cite{liu2021editing}, and CodeNeRF~\cite{jang2021codenerf} on CARLA~\cite{dosovitskiy2017carla} benchmark of applying random perturbations on shape (row 2-5), appearance codes (row 6-9) and camera (row 10-12) respectively while maintaining other attributes consistent.}\vspace{-5pt}
\label{random_perturbations}
\end{figure*}

\begin{figure*}[h]
\begin{center}
\begin{tabular}{M{0.04\linewidth} M{0.27\linewidth} M{0.27\linewidth} M{0.27\linewidth}}
&{\includegraphics[width=0.33\linewidth]{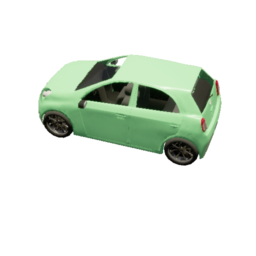}}\hfill
{\includegraphics[width=0.33\linewidth]{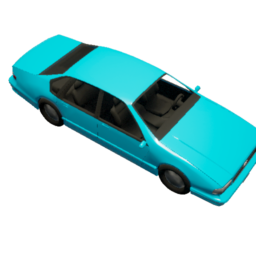}}\hfill &
{\includegraphics[width=0.33\linewidth]{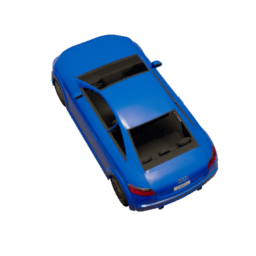}}\hfill
{\includegraphics[width=0.33\linewidth]{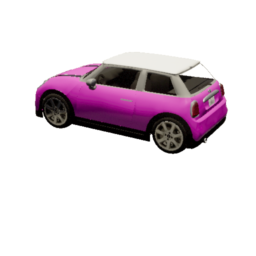}}\hfill &
{\includegraphics[width=0.33\linewidth]{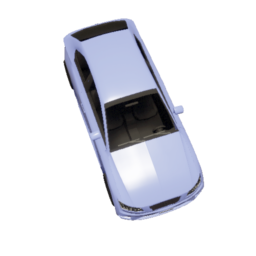}}\hfill
{\includegraphics[width=0.33\linewidth]{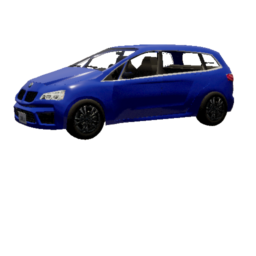}}\hfill \\
\rotatebox{90}{SA} &
{\includegraphics[width=0.33\linewidth]{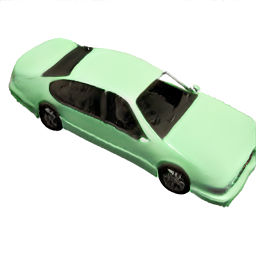}}\hfill
{\includegraphics[width=0.33\linewidth]{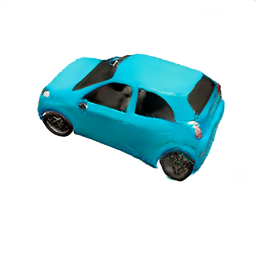}}\hfill
{\includegraphics[width=0.33\linewidth]{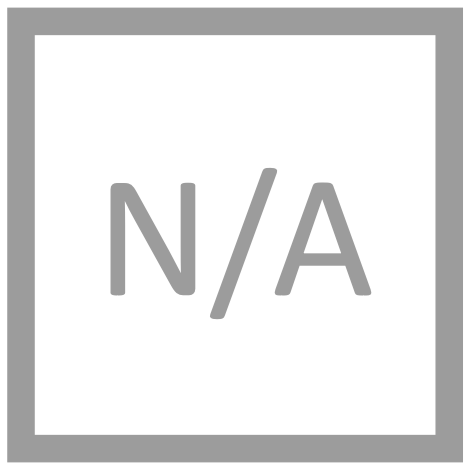}}\hfill&
{\includegraphics[width=0.33\linewidth]{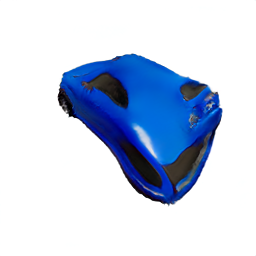}}\hfill
{\includegraphics[width=0.33\linewidth]{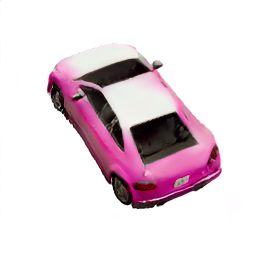}}\hfill
{\includegraphics[width=0.33\linewidth]{swap_comparison/NA.png}}\hfill&
{\includegraphics[width=0.33\linewidth]{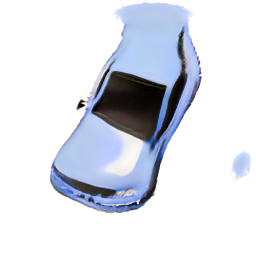}}\hfill
{\includegraphics[width=0.33\linewidth]{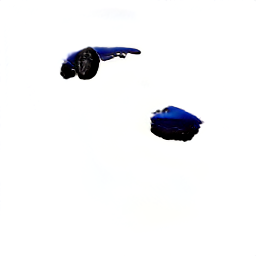}}\hfill
{\includegraphics[width=0.33\linewidth]{swap_comparison/NA.png}}\hfill\\
\rotatebox{90}{\makecell{Code \\ NeRF}} &
{\includegraphics[width=0.33\linewidth]{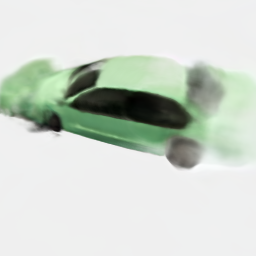}}\hfill
{\includegraphics[width=0.33\linewidth]{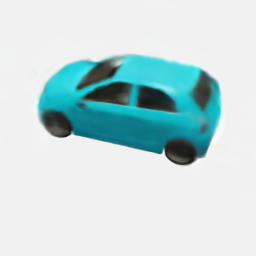}}\hfill
{\includegraphics[width=0.33\linewidth]{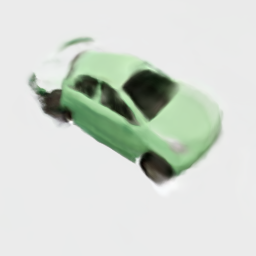}}\hfill&
{\includegraphics[width=0.33\linewidth]{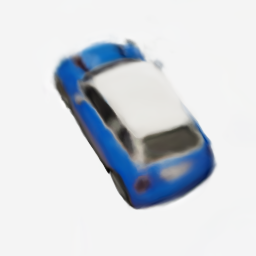}}\hfill
{\includegraphics[width=0.33\linewidth]{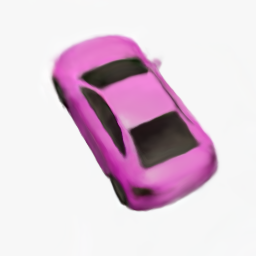}}\hfill
{\includegraphics[width=0.33\linewidth]{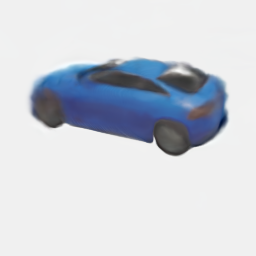}}\hfill&
{\includegraphics[width=0.33\linewidth]{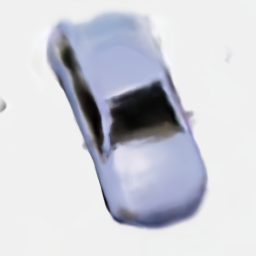}}\hfill
{\includegraphics[width=0.33\linewidth]{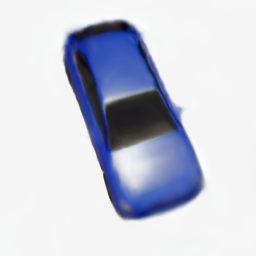}}\hfill
{\includegraphics[width=0.33\linewidth]{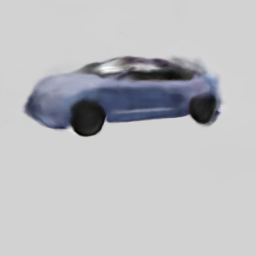}}\hfill\\
\rotatebox{90}{\makecell{Edit \\ NeRF}} &
{\includegraphics[width=0.33\linewidth]{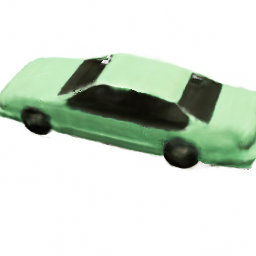}}\hfill
{\includegraphics[width=0.33\linewidth]{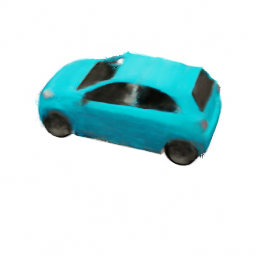}}\hfill
{\includegraphics[width=0.33\linewidth]{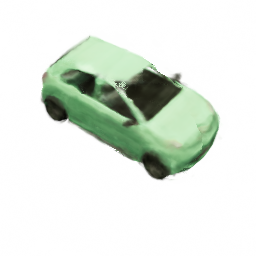}}\hfill&
{\includegraphics[width=0.33\linewidth]{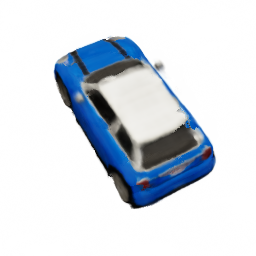}}\hfill
{\includegraphics[width=0.33\linewidth]{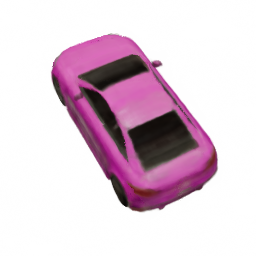}}\hfill
{\includegraphics[width=0.33\linewidth]{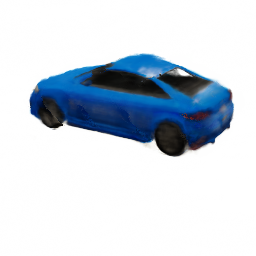}}\hfill&
{\includegraphics[width=0.33\linewidth]{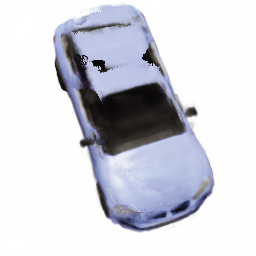}}\hfill
{\includegraphics[width=0.33\linewidth]{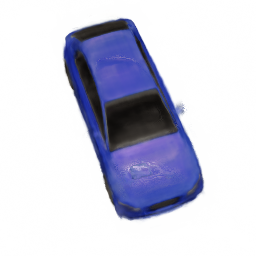}}\hfill
{\includegraphics[width=0.33\linewidth]{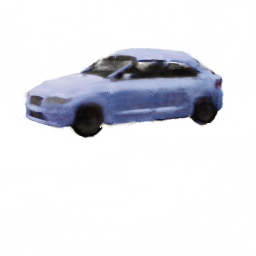}}\hfill\\
\rotatebox{90}{\makecell{AE- \\ NeRF}} &
{\includegraphics[width=0.33\linewidth]{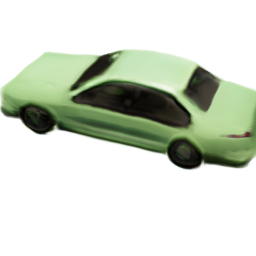}}\hfill
{\includegraphics[width=0.33\linewidth]{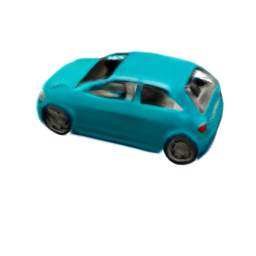}}\hfill
{\includegraphics[width=0.33\linewidth]{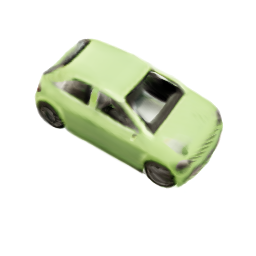}}\hfill&
{\includegraphics[width=0.33\linewidth]{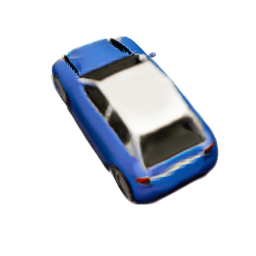}}\hfill
{\includegraphics[width=0.33\linewidth]{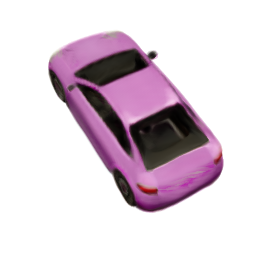}}\hfill
{\includegraphics[width=0.33\linewidth]{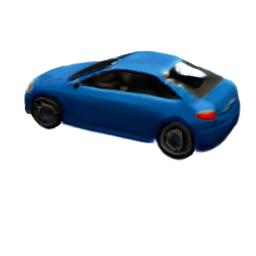}}\hfill&
{\includegraphics[width=0.33\linewidth]{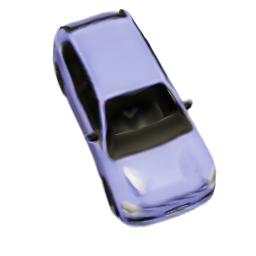}}\hfill
{\includegraphics[width=0.33\linewidth]{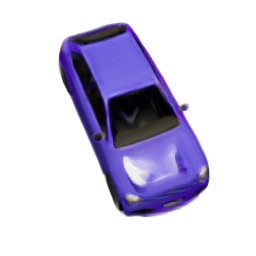}}\hfill
{\includegraphics[width=0.33\linewidth]{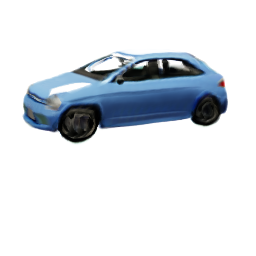}}\hfill\\

\end{tabular}
\end{center}
\vspace{-10pt}
\caption{\textbf{Qualitative comparison on swapping disentangled shape and appearance on CARLA~\cite{chang2015shapenet} benchmark.} Given images (row 1) as a pair of input images, we compare the attribute swapping results in the order of shape, appearance, and camera pose: SA~\cite{park2020swapping} (row 2), CodeNeRF~\cite{jang2021codenerf} (row 3), EditNeRF~\cite{liu2021editing} (row 4), and AE-NeRF (row 5).}
\label{swap_comparison}\vspace{-5pt}
\end{figure*}
\begin{figure*}[!t]
\centering
\begin{tabular}{M{0.06\linewidth} M{0.9\linewidth}}
\rotatebox{90}{\makecell{Source}} &
{\includegraphics[width=0.11\linewidth]{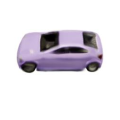}}\hfill
{\includegraphics[width=0.11\linewidth]{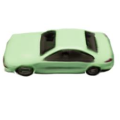}}\hfill 
{\includegraphics[width=0.11\linewidth]{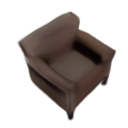}}\hfill
{\includegraphics[width=0.11\linewidth]{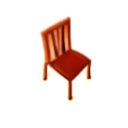}}\hfill
\hspace{10pt}
{\includegraphics[width=0.11\linewidth]{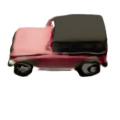}}\hfill 
{\includegraphics[width=0.11\linewidth]{clipnerf/gt_carla/src1.png}}\hfill
{\includegraphics[width=0.11\linewidth]{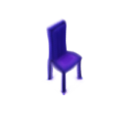}}\hfill
{\includegraphics[width=0.11\linewidth]{clipnerf/gt_chairs/src1.png}}\hfill
\\

\rotatebox{90}{\makecell{Target}} &
{\includegraphics[width=0.11\linewidth]{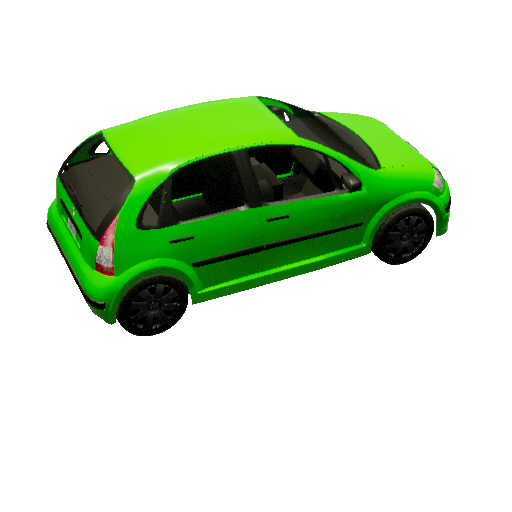}}\hfill
{\includegraphics[width=0.11\linewidth]{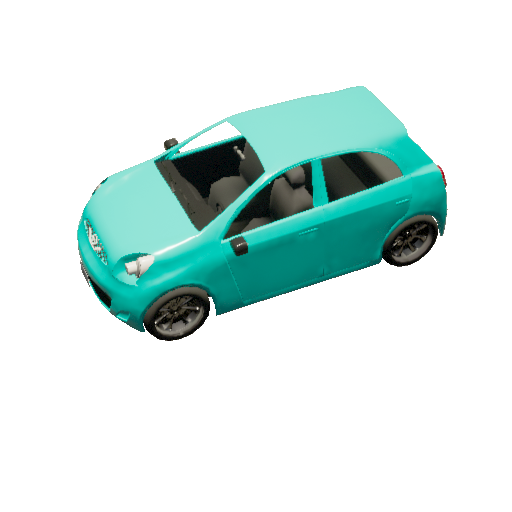}}\hfill 
{\includegraphics[width=0.11\linewidth]{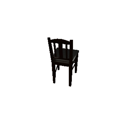}}\hfill
{\includegraphics[width=0.11\linewidth]{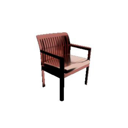}}\hfill
\hspace{10pt}
{\includegraphics[width=0.11\linewidth]{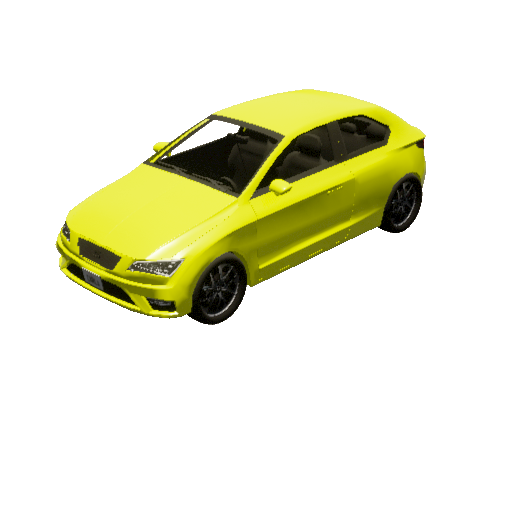}}\hfill 
{\includegraphics[width=0.11\linewidth]{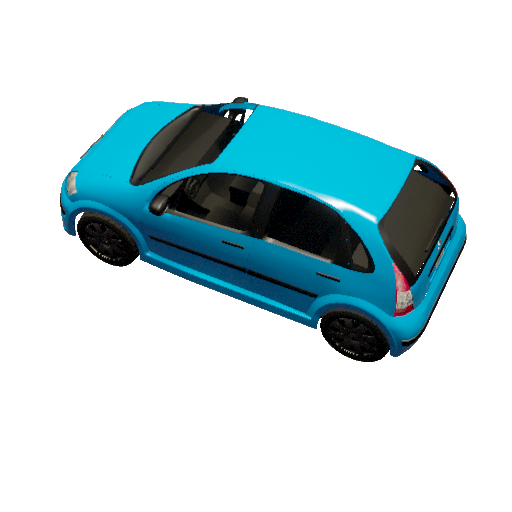}}\hfill
{\includegraphics[width=0.11\linewidth]{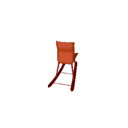}}\hfill
{\includegraphics[width=0.11\linewidth]{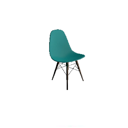}}\hfill
\\

\rotatebox{90}{\makecell{Clip \\ NeRF}} &
{\includegraphics[width=0.11\linewidth]{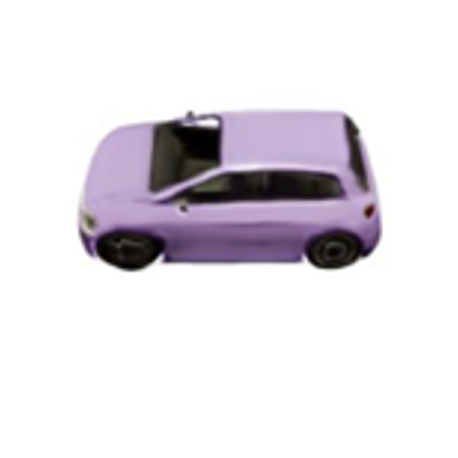}}\hfill
{\includegraphics[width=0.11\linewidth]{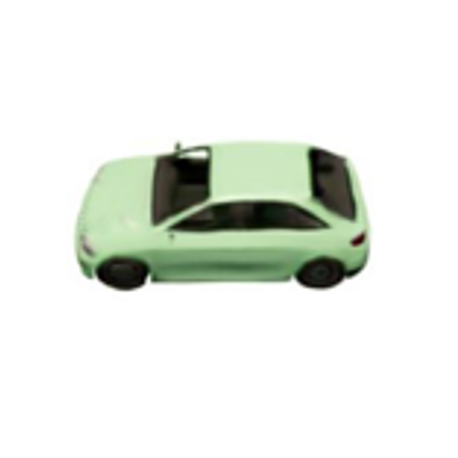}}\hfill
{\includegraphics[width=0.11\linewidth]{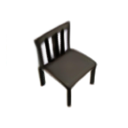}}\hfill 
{\includegraphics[width=0.11\linewidth]{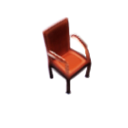}}\hfill
\hspace{10pt}
{\includegraphics[width=0.11\linewidth]{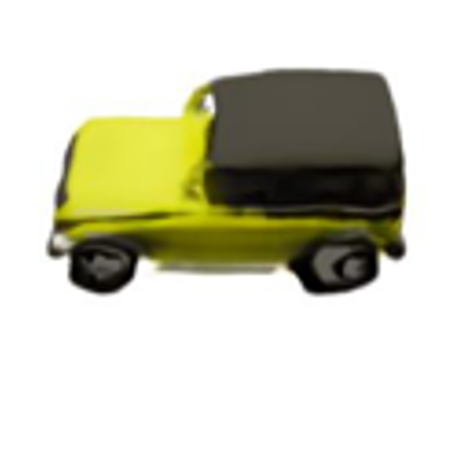}}\hfill 
{\includegraphics[width=0.11\linewidth]{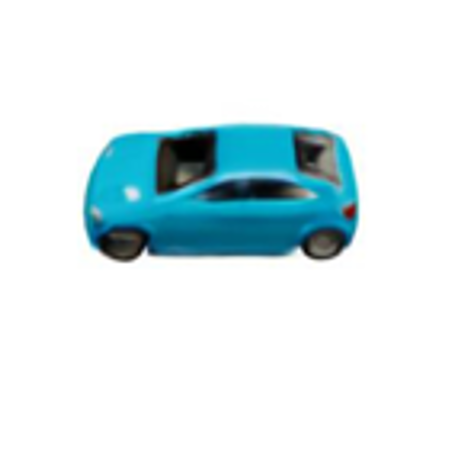}}\hfill
{\includegraphics[width=0.11\linewidth]{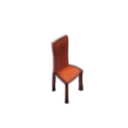}}\hfill
{\includegraphics[width=0.11\linewidth]{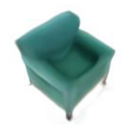}}\hfill \\

\rotatebox{90}{\makecell{AE- \\ NeRF}} &
{\includegraphics[width=0.11\linewidth]{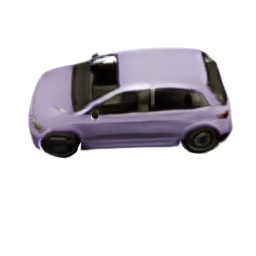}}\hfill
{\includegraphics[width=0.11\linewidth]{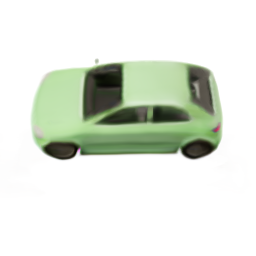}}\hfill
{\includegraphics[width=0.11\linewidth]{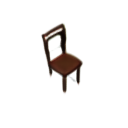}}\hfill 
{\includegraphics[width=0.11\linewidth]{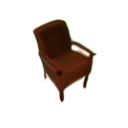}}\hfill
\hspace{10pt}
{\includegraphics[width=0.11\linewidth]{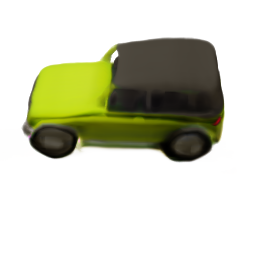}}\hfill 
{\includegraphics[width=0.11\linewidth]{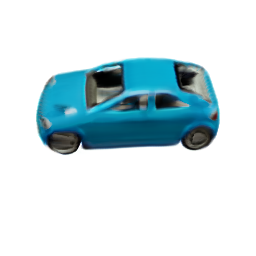}}\hfill
{\includegraphics[width=0.11\linewidth]{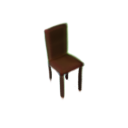}}\hfill
{\includegraphics[width=0.11\linewidth]{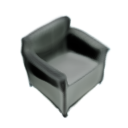}}\hfill \\

\end{tabular}

\caption{\textbf{Code swapping comparison with ClipNeRF on CARLA~\cite{dosovitskiy2017carla} and Photoshapes~\cite{park2018photoshapes} dataset.}}
\label{auto-encoding-chairs}
\end{figure*}

\begin{figure*}[!t]
\centering
\begin{center}
\begin{tabular}{M{0.10\linewidth} c M{0.64\linewidth} M{0.10\linewidth}}
\multirow{8}{*}{\begin{tabular}{c} 
{\includegraphics[width=0.8\linewidth]{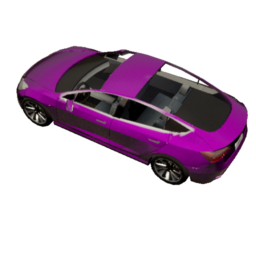}\hfill } \\  Image 1 \\ 
\end{tabular}  }
& 
sh. &
\centering
{\includegraphics[width=0.1\linewidth]{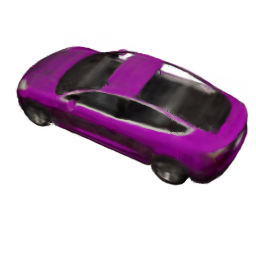}
\includegraphics[width=0.1\linewidth]{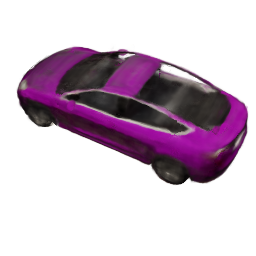}
\includegraphics[width=0.1\linewidth]{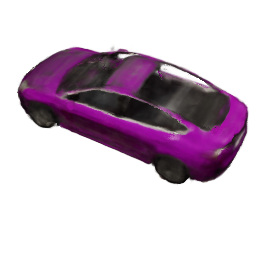}
\includegraphics[width=0.1\linewidth]{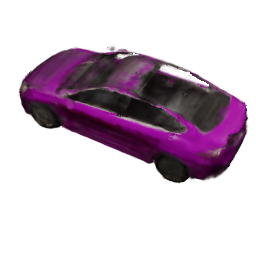}
\includegraphics[width=0.1\linewidth]{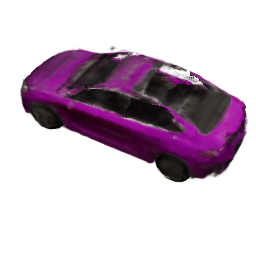}
\includegraphics[width=0.1\linewidth]{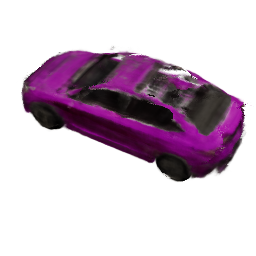}
\includegraphics[width=0.1\linewidth]{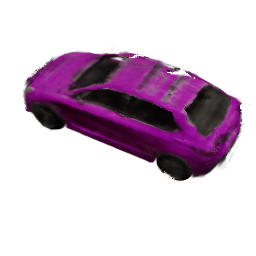}
\includegraphics[width=0.1\linewidth]{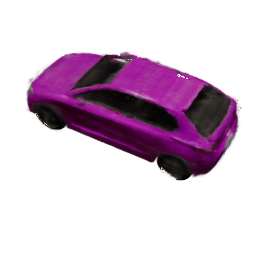}
\includegraphics[width=0.1\linewidth]{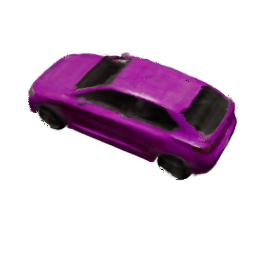}
} &
\multirow{8}{*}{\begin{tabular}{c} 
{\includegraphics[width=0.8\linewidth]{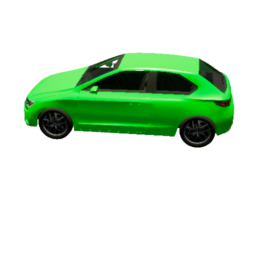}\hfill } \\  Image 2 
\end{tabular}  }
\\ 
& app. & 
\centering
{\includegraphics[width=0.1\linewidth]{interpolation/editnerf/0191_rgb.png}
\includegraphics[width=0.1\linewidth]{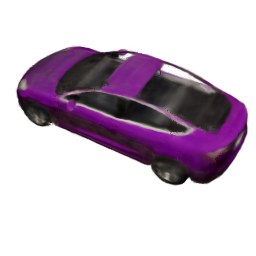}
\includegraphics[width=0.1\linewidth]{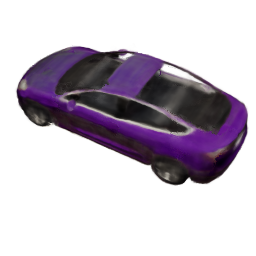}
\includegraphics[width=0.1\linewidth]{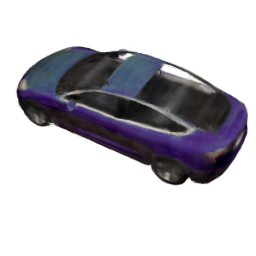}
\includegraphics[width=0.1\linewidth]{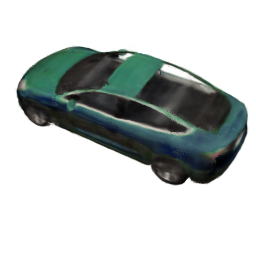}
\includegraphics[width=0.1\linewidth]{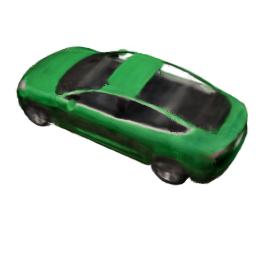}
\includegraphics[width=0.1\linewidth]{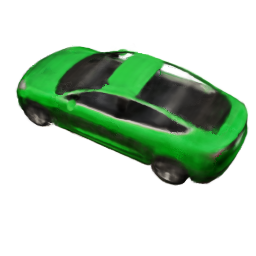}
\includegraphics[width=0.1\linewidth]{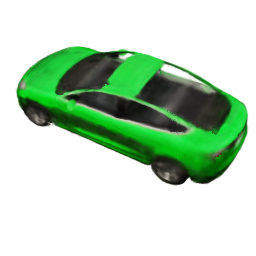}
\includegraphics[width=0.1\linewidth]{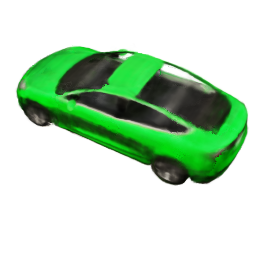}
} &
\\
& cam. & 
\centering
{\includegraphics[width=0.1\linewidth]{interpolation/editnerf/0191_rgb.png}
\includegraphics[width=0.1\linewidth]{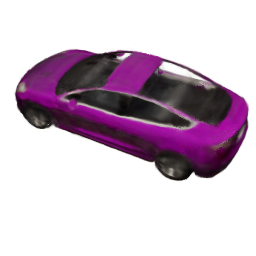}
\includegraphics[width=0.1\linewidth]{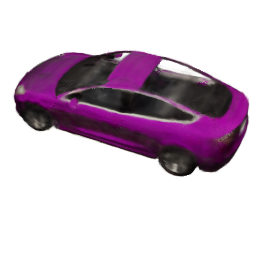}
\includegraphics[width=0.1\linewidth]{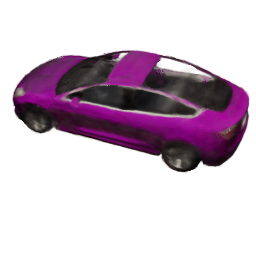}
\includegraphics[width=0.1\linewidth]{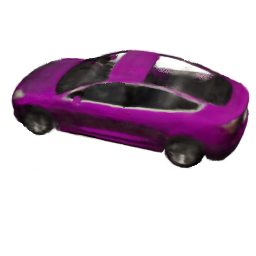}
\includegraphics[width=0.1\linewidth]{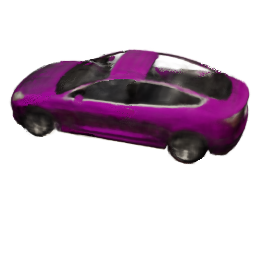}
\includegraphics[width=0.1\linewidth]{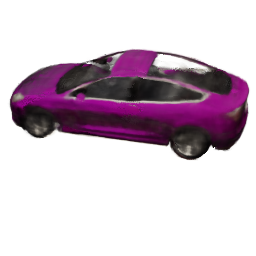}
\includegraphics[width=0.1\linewidth]{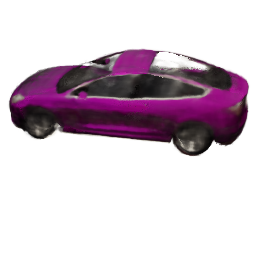}
\includegraphics[width=0.1\linewidth]{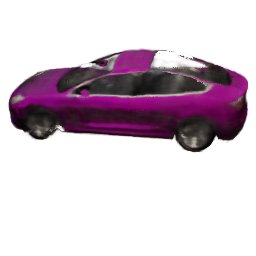}} &
\\
& all & 
\centering
{\includegraphics[width=0.1\linewidth]{interpolation/editnerf/0191_rgb.png}
\includegraphics[width=0.1\linewidth]{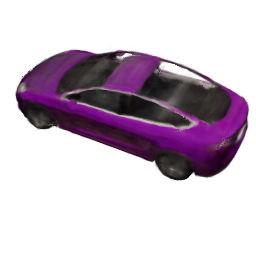}
\includegraphics[width=0.1\linewidth]{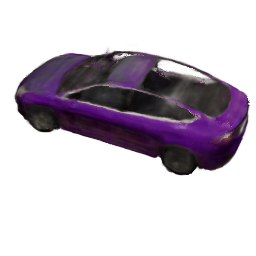}
\includegraphics[width=0.1\linewidth]{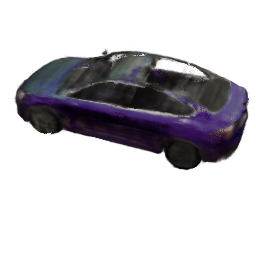}
\includegraphics[width=0.1\linewidth]{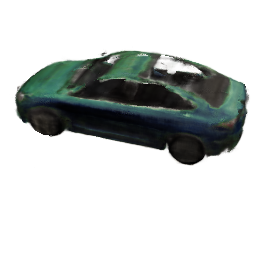}
\includegraphics[width=0.1\linewidth]{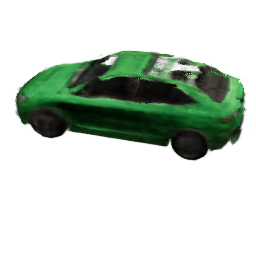}
\includegraphics[width=0.1\linewidth]{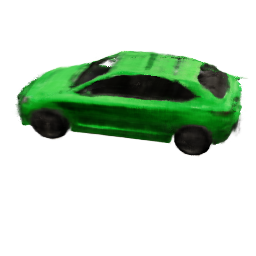}
\includegraphics[width=0.1\linewidth]{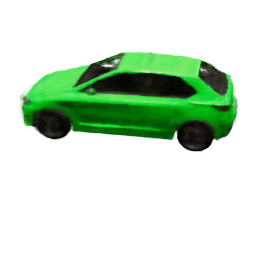}
\includegraphics[width=0.1\linewidth]{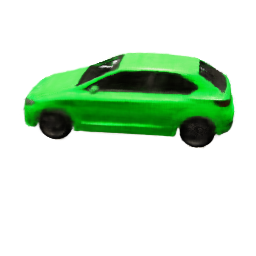}} &
\\ [-5pt]
& & \multicolumn{1}{c}{$\xleftarrow{\hspace*{3.9cm}}$ interpolation ratio $\xrightarrow{\hspace*{3.9cm}}$}& \\
\multirow{8}{*}{\begin{tabular}{c} 
{\includegraphics[width=0.8\linewidth]{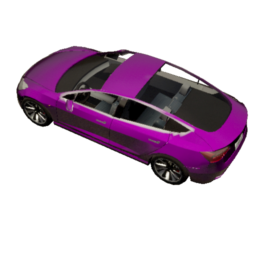}\hfill } \\  Image 1 \\ 
\end{tabular}  }
& 
sh. &
\centering
{\includegraphics[width=0.1\linewidth]{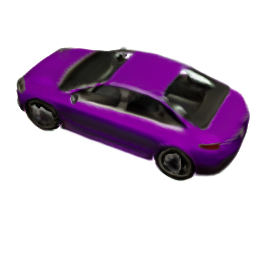}
\includegraphics[width=0.1\linewidth]{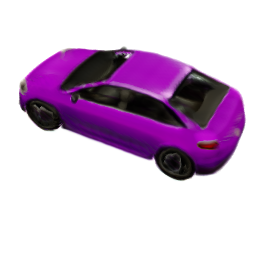}
\includegraphics[width=0.1\linewidth]{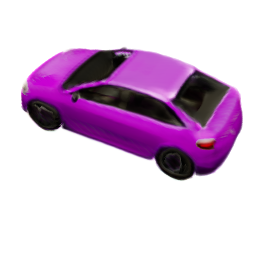}
\includegraphics[width=0.1\linewidth]{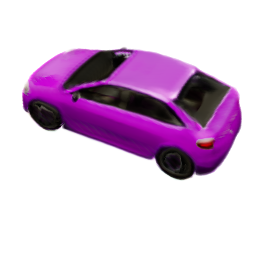}
\includegraphics[width=0.1\linewidth]{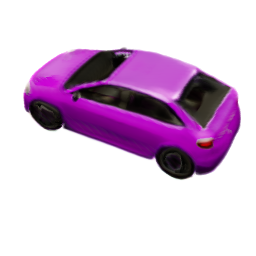}
\includegraphics[width=0.1\linewidth]{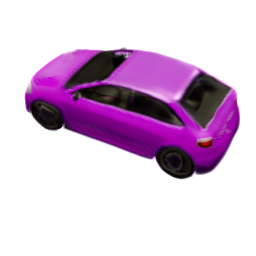}
\includegraphics[width=0.1\linewidth]{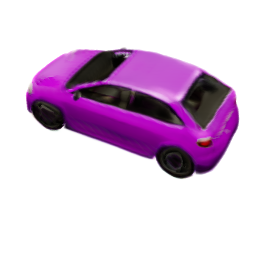}
\includegraphics[width=0.1\linewidth]{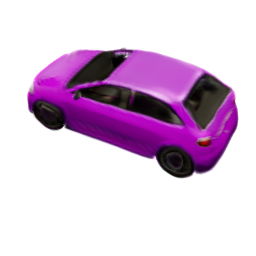}
\includegraphics[width=0.1\linewidth]{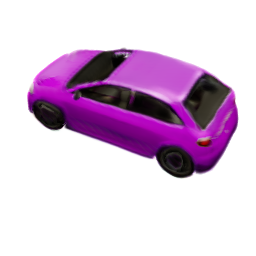}} &
\multirow{8}{*}{\begin{tabular}{c} 
{\includegraphics[width=0.8\linewidth]{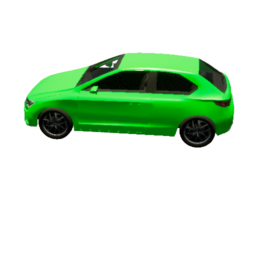}\hfill } \\  Image 2 
\end{tabular}  }
\\ 
& app. & 
\centering
{\includegraphics[width=0.1\linewidth]{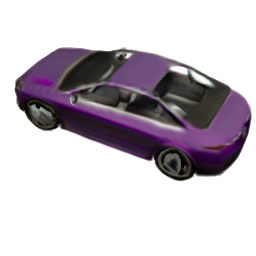}
\includegraphics[width=0.1\linewidth]{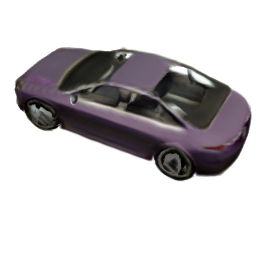}
\includegraphics[width=0.1\linewidth]{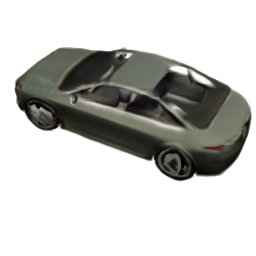}
\includegraphics[width=0.1\linewidth]{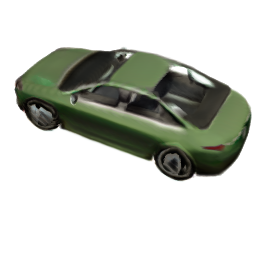}
\includegraphics[width=0.1\linewidth]{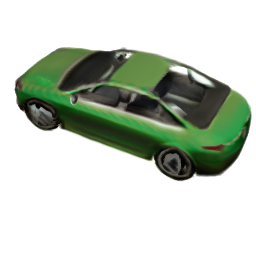}
\includegraphics[width=0.1\linewidth]{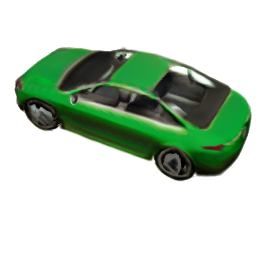}
\includegraphics[width=0.1\linewidth]{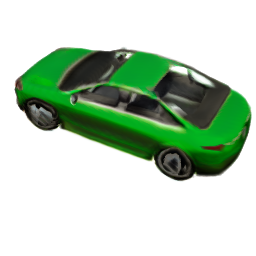}
\includegraphics[width=0.1\linewidth]{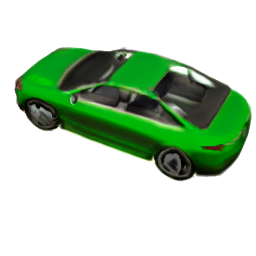}
\includegraphics[width=0.1\linewidth]{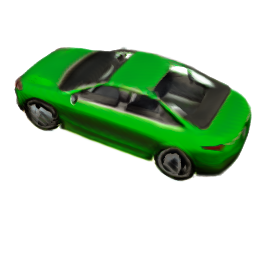}} &
\\
& cam. & 
\centering
{\includegraphics[width=0.1\linewidth]{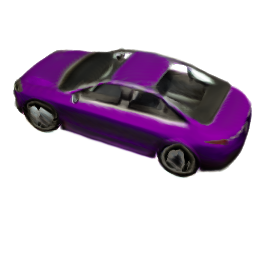}
\includegraphics[width=0.1\linewidth]{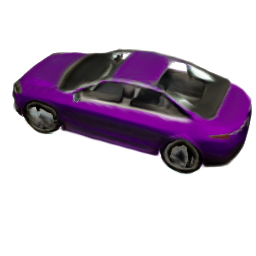}
\includegraphics[width=0.1\linewidth]{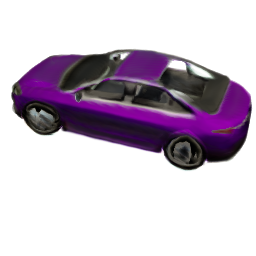}
\includegraphics[width=0.1\linewidth]{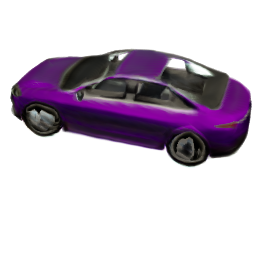}
\includegraphics[width=0.1\linewidth]{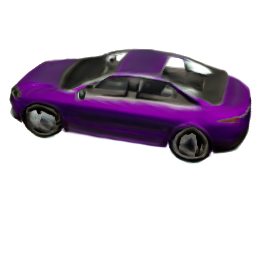}
\includegraphics[width=0.1\linewidth]{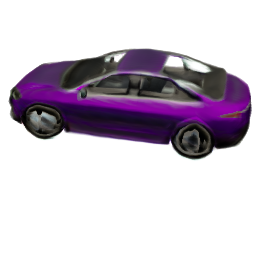}
\includegraphics[width=0.1\linewidth]{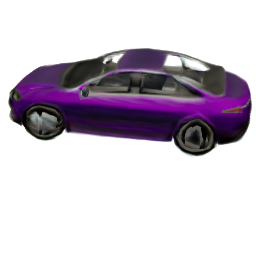}
\includegraphics[width=0.1\linewidth]{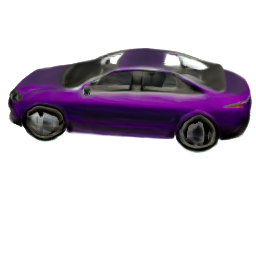}
\includegraphics[width=0.1\linewidth]{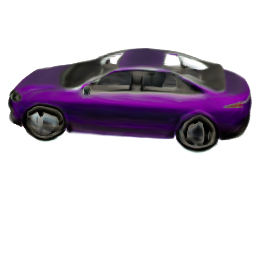}} &
\\
& all & 
\centering
{\includegraphics[width=0.1\linewidth]{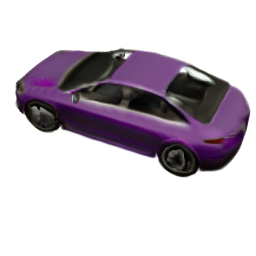}
\includegraphics[width=0.1\linewidth]{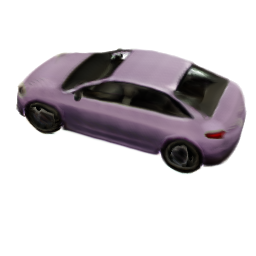}
\includegraphics[width=0.1\linewidth]{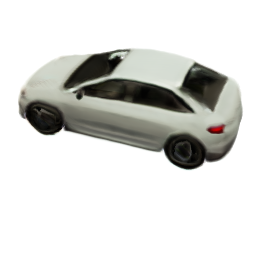}
\includegraphics[width=0.1\linewidth]{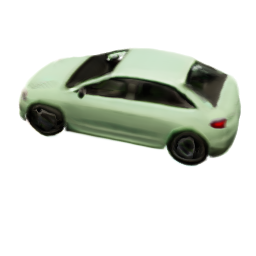}
\includegraphics[width=0.1\linewidth]{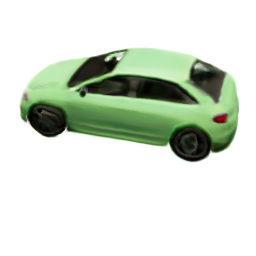}
\includegraphics[width=0.1\linewidth]{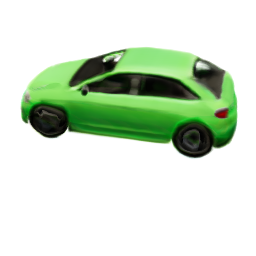}
\includegraphics[width=0.1\linewidth]{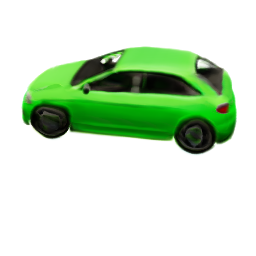}
\includegraphics[width=0.1\linewidth]{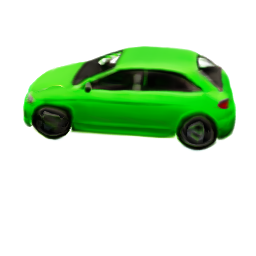}
\includegraphics[width=0.1\linewidth]{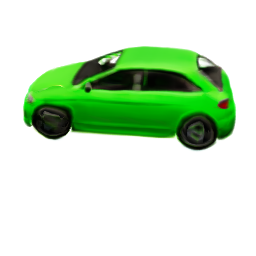}} &
\\ [-5pt]
& & \multicolumn{1}{c}{$\xleftarrow{\hspace*{3.9cm}}$ interpolation ratio $\xrightarrow{\hspace*{3.9cm}}$}&  \\
\vspace{10pt} 

\multirow{8}{*}{\begin{tabular}{c} 
{\includegraphics[width=0.8\linewidth]{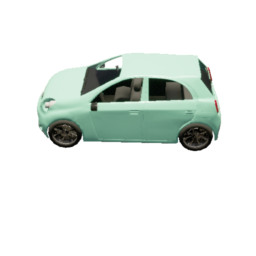}\hfill } \\  Image 1 \\ 
\end{tabular}  }
& 
sh. &
\centering
{\includegraphics[width=0.1\linewidth]{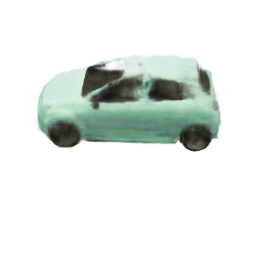}
\includegraphics[width=0.1\linewidth]{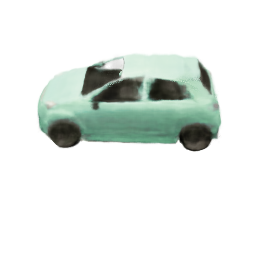}
\includegraphics[width=0.1\linewidth]{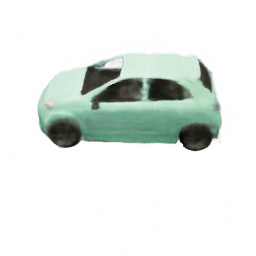}
\includegraphics[width=0.1\linewidth]{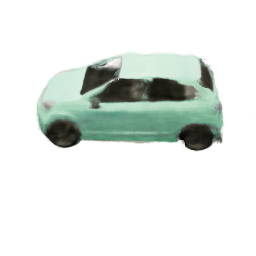}
\includegraphics[width=0.1\linewidth]{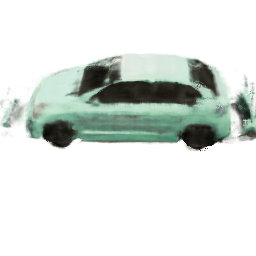}
\includegraphics[width=0.1\linewidth]{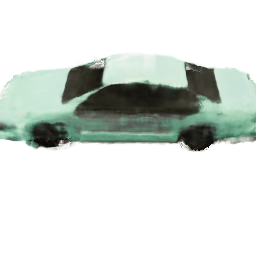}
\includegraphics[width=0.1\linewidth]{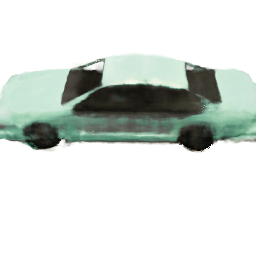}
\includegraphics[width=0.1\linewidth]{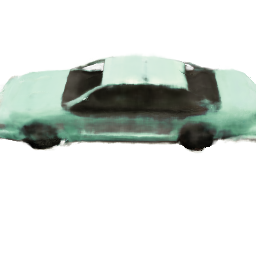}
\includegraphics[width=0.1\linewidth]{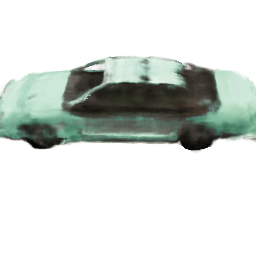}} &
\multirow{8}{*}{\begin{tabular}{c} 
{\includegraphics[width=0.8\linewidth]{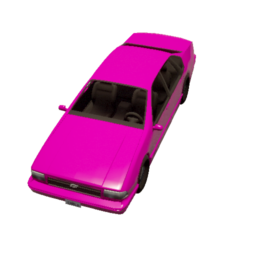}\hfill } \\  Image 2 
\end{tabular}  }
\\ 
& app. & 
\centering
{\includegraphics[width=0.1\linewidth]{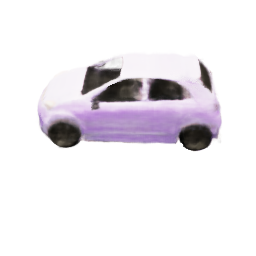}
\includegraphics[width=0.1\linewidth]{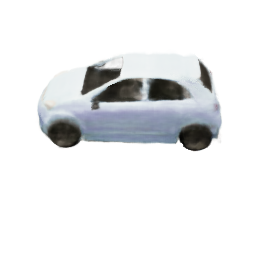}
\includegraphics[width=0.1\linewidth]{extrapolation/editnerf/edit_recon.png}
\includegraphics[width=0.1\linewidth]{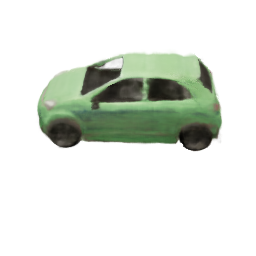}
\includegraphics[width=0.1\linewidth]{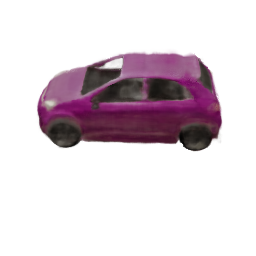}
\includegraphics[width=0.1\linewidth]{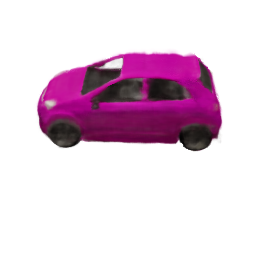}
\includegraphics[width=0.1\linewidth]{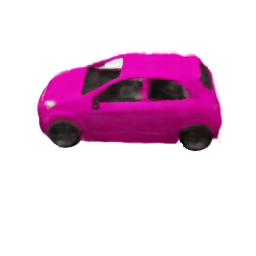}
\includegraphics[width=0.1\linewidth]{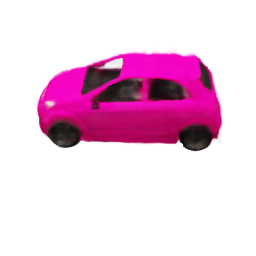}
\includegraphics[width=0.1\linewidth]{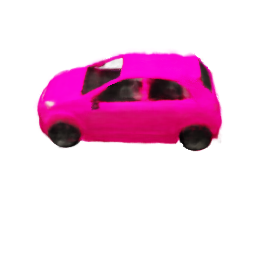}} &
\\
& cam. & 
\centering
{\includegraphics[width=0.1\linewidth]{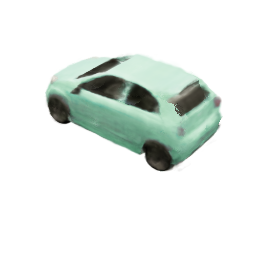}
\includegraphics[width=0.1\linewidth]{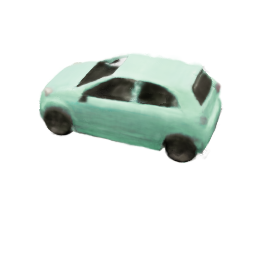}
\includegraphics[width=0.1\linewidth]{extrapolation/editnerf/edit_recon.png}
\includegraphics[width=0.1\linewidth]{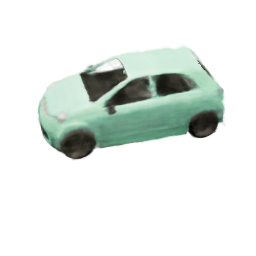}
\includegraphics[width=0.1\linewidth]{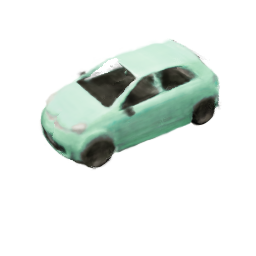}
\includegraphics[width=0.1\linewidth]{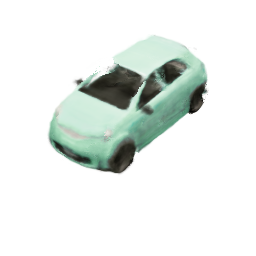}
\includegraphics[width=0.1\linewidth]{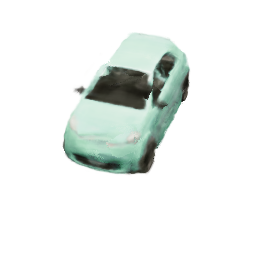}
\includegraphics[width=0.1\linewidth]{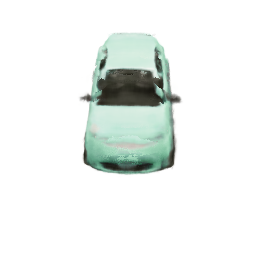}
\includegraphics[width=0.1\linewidth]{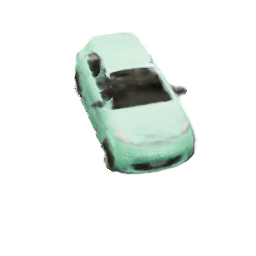}} &
\\
& all & 
\centering
{\includegraphics[width=0.1\linewidth]{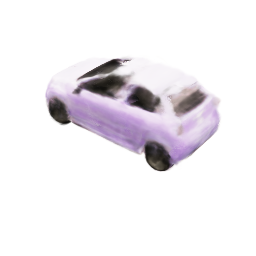}
\includegraphics[width=0.1\linewidth]{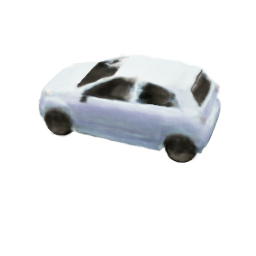}
\includegraphics[width=0.1\linewidth]{extrapolation/editnerf/edit_recon.png}
\includegraphics[width=0.1\linewidth]{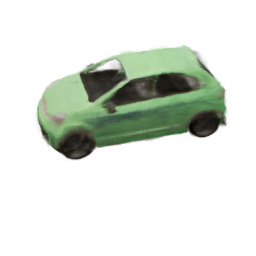}
\includegraphics[width=0.1\linewidth]{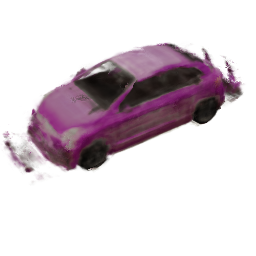}
\includegraphics[width=0.1\linewidth]{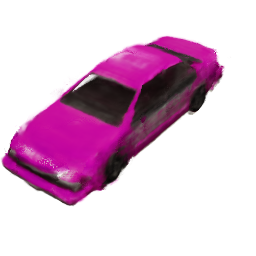}
\includegraphics[width=0.1\linewidth]{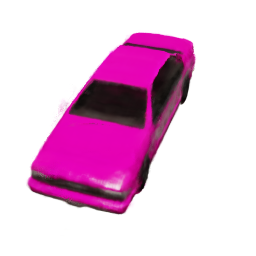}
\includegraphics[width=0.1\linewidth]{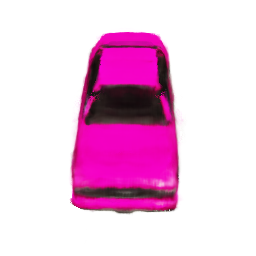}
\includegraphics[width=0.1\linewidth]{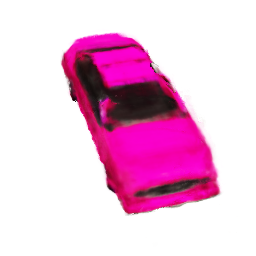}} &
\\ [-5pt]
& &\multicolumn{1}{c}{$\xleftarrow{\hspace*{1.05cm}}$ extra. $\xrightarrow{\hspace*{1.05cm}}$ $\xleftarrow{\hspace*{1.05cm}}$ inter. $\xrightarrow{\hspace*{1.05cm}}$ $\xleftarrow{\hspace*{1.05cm}}$ extra. $\xrightarrow{\hspace*{1.05cm}}$ }& \\[10pt] 

\multirow{8}{*}{\begin{tabular}{c} 
{\includegraphics[width=0.8\linewidth]{extrapolation/79_153/0_gt.png}\hfill } \\  Image 1 \\ 
\end{tabular}  }
& 
sh. &
\centering
{\includegraphics[width=0.1\linewidth]{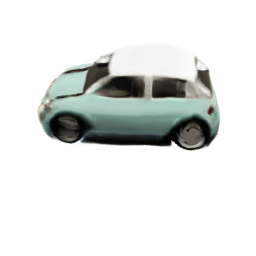}
\includegraphics[width=0.1\linewidth]{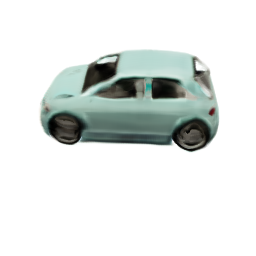}
\includegraphics[width=0.1\linewidth]{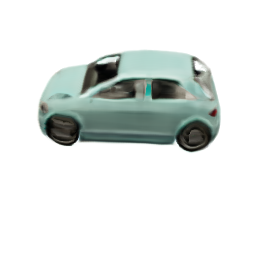}
\includegraphics[width=0.1\linewidth]{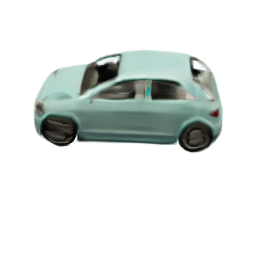}
\includegraphics[width=0.1\linewidth]{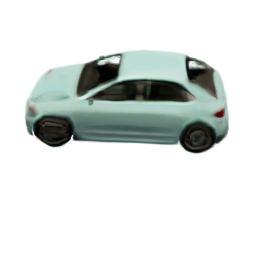}
\includegraphics[width=0.1\linewidth]{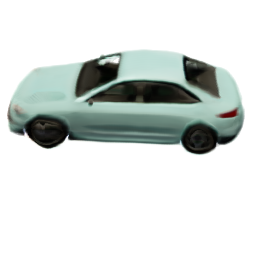}
\includegraphics[width=0.1\linewidth]{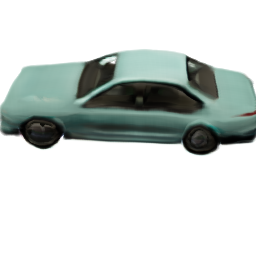}
\includegraphics[width=0.1\linewidth]{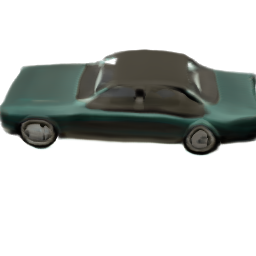}
\includegraphics[width=0.1\linewidth]{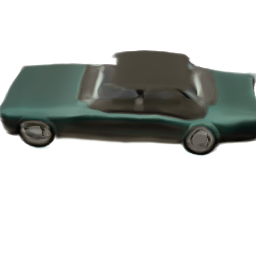}} &
\multirow{8}{*}{\begin{tabular}{c} 
{\includegraphics[width=0.8\linewidth]{extrapolation/79_153/1_gt.png}\hfill } \\  Image 2 
\end{tabular}  }
\\ 
& app. & 
\centering
{\includegraphics[width=0.1\linewidth]{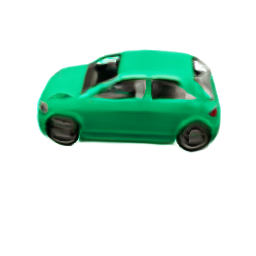}
\includegraphics[width=0.1\linewidth]{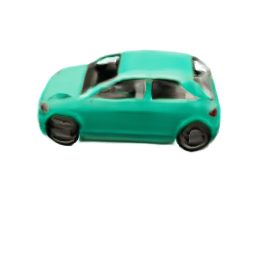}
\includegraphics[width=0.1\linewidth]{extrapolation/79_153/0_recon.png}
\includegraphics[width=0.1\linewidth]{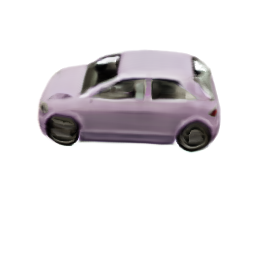}
\includegraphics[width=0.1\linewidth]{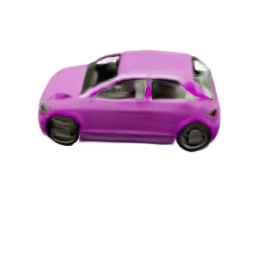}
\includegraphics[width=0.1\linewidth]{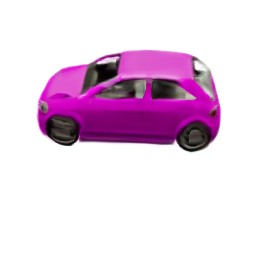}
\includegraphics[width=0.1\linewidth]{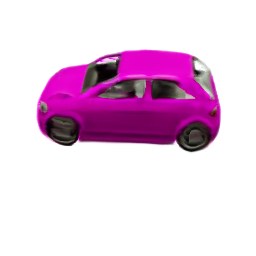}
\includegraphics[width=0.1\linewidth]{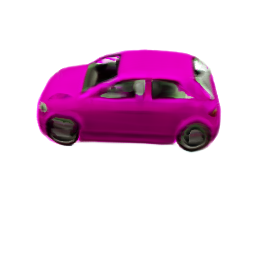}
\includegraphics[width=0.1\linewidth]{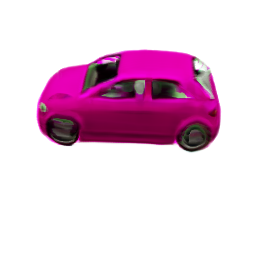}} &
\\
& cam. & 
\centering
{\includegraphics[width=0.1\linewidth]{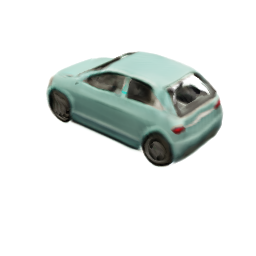}
\includegraphics[width=0.1\linewidth]{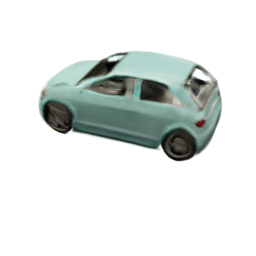}
\includegraphics[width=0.1\linewidth]{extrapolation/79_153/0_recon.png}
\includegraphics[width=0.1\linewidth]{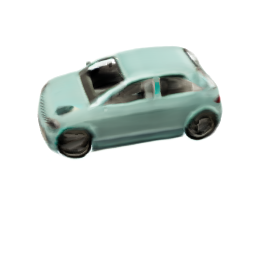}
\includegraphics[width=0.1\linewidth]{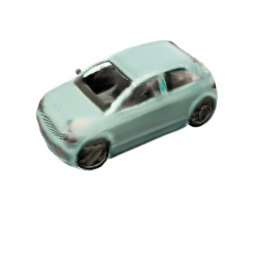}
\includegraphics[width=0.1\linewidth]{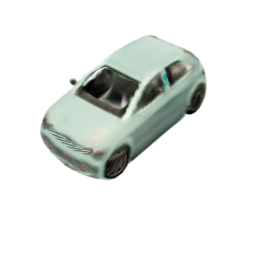}
\includegraphics[width=0.1\linewidth]{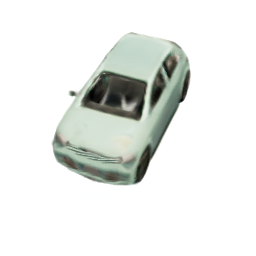}
\includegraphics[width=0.1\linewidth]{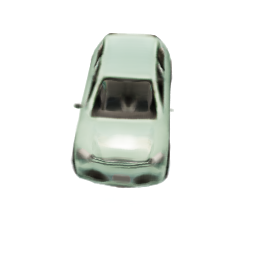}
\includegraphics[width=0.1\linewidth]{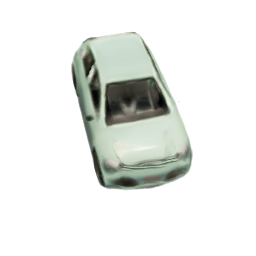}} &
\\
& all & 
\centering
{\includegraphics[width=0.1\linewidth]{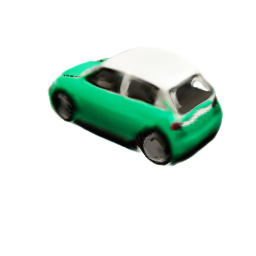}
\includegraphics[width=0.1\linewidth]{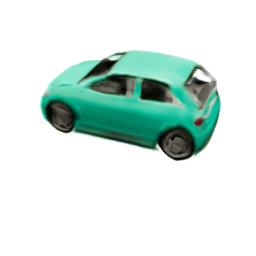}
\includegraphics[width=0.1\linewidth]{extrapolation/79_153/0_recon.png}
\includegraphics[width=0.1\linewidth]{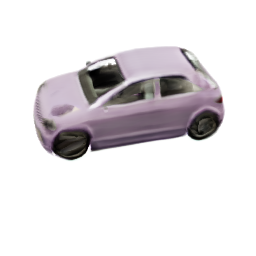}
\includegraphics[width=0.1\linewidth]{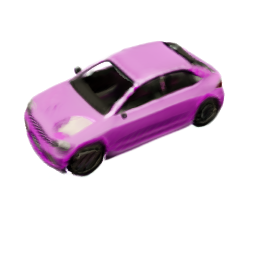}
\includegraphics[width=0.1\linewidth]{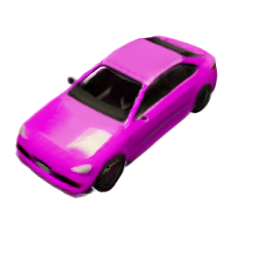}
\includegraphics[width=0.1\linewidth]{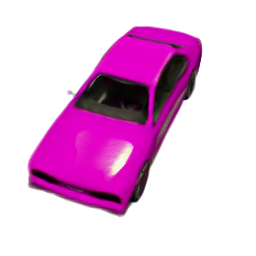}
\includegraphics[width=0.1\linewidth]{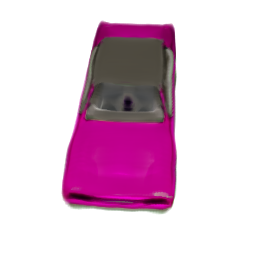}
\includegraphics[width=0.1\linewidth]{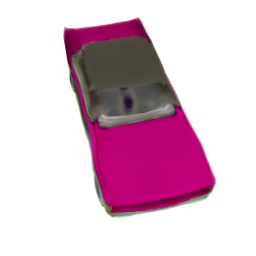}} &
\\ [-5pt]
& &\multicolumn{1}{c}{$\xleftarrow{\hspace*{1.05cm}}$ extra. $\xrightarrow{\hspace*{1.05cm}}$ $\xleftarrow{\hspace*{1.05cm}}$ inter. $\xrightarrow{\hspace*{1.05cm}}$ $\xleftarrow{\hspace*{1.05cm}}$ extra. $\xrightarrow{\hspace*{1.05cm}}$ }& \\

\end{tabular}

\end{center}
\vspace{-5pt}
\caption{\textbf{Comparison of attribute interpolation and extrapolation between (EditNeRF~\cite{liu2021editing} (1-4, 9-12 rows) and AE-NeRF (5-8, 13-16 rows)}. We manipulate shape, appearance, camera viewpoint attribute and the combination of the three between two images. Given two different images, our AE-NeRF smoothly interpolates the generated images within real data distribution as well as disentangles the shape, appearance, and camera poses, while EditNeRF~\cite{liu2021editing} often fails.}
\label{extrapolation example}
\end{figure*}

\begin{figure*}[!t]
\centering
\begin{tabular}{M{0.06\linewidth} M{0.9\linewidth}}
\rotatebox{90}{\makecell{GT}} &
{\includegraphics[width=0.11\linewidth]{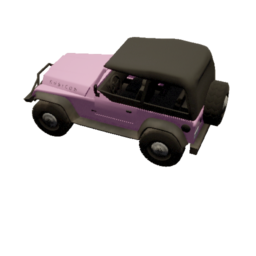}}\hfill
{\includegraphics[width=0.11\linewidth]{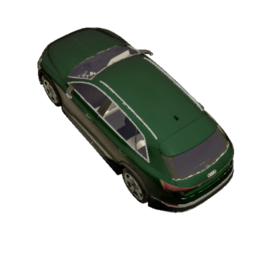}}\hfill 
{\includegraphics[width=0.11\linewidth]{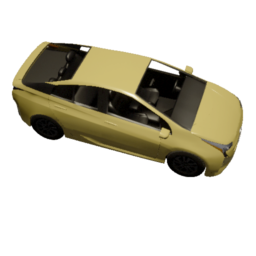}}\hfill
{\includegraphics[width=0.11\linewidth]{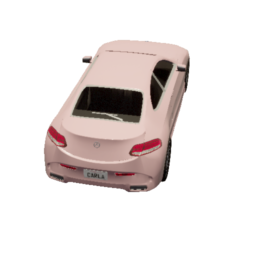}}\hfill
{\includegraphics[width=0.11\linewidth]{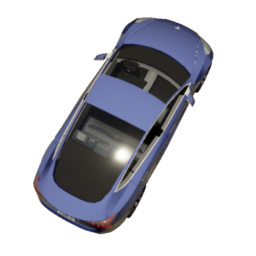}}\hfill 
{\includegraphics[width=0.11\linewidth]{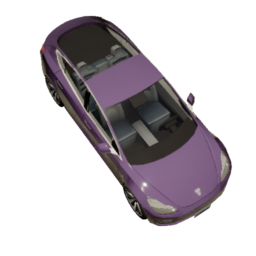}}\hfill
{\includegraphics[width=0.11\linewidth]{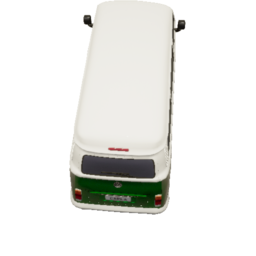}}\hfill
\\

\rotatebox{90}{\makecell{Scratch}} &
{\includegraphics[width=0.11\linewidth]{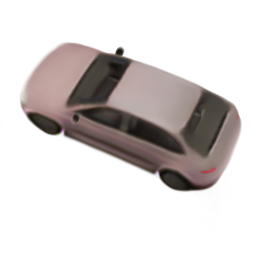}}\hfill
{\includegraphics[width=0.11\linewidth]{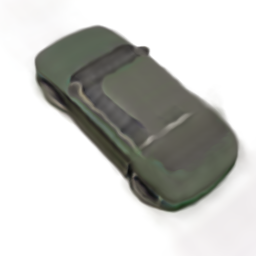}}\hfill 
{\includegraphics[width=0.11\linewidth]{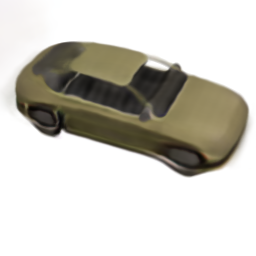}}\hfill
{\includegraphics[width=0.11\linewidth]{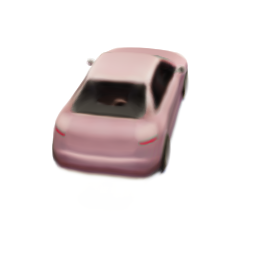}}\hfill
{\includegraphics[width=0.11\linewidth]{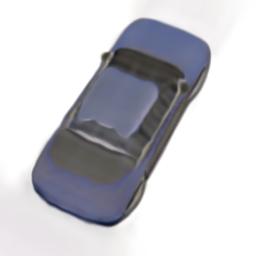}}\hfill 
{\includegraphics[width=0.11\linewidth]{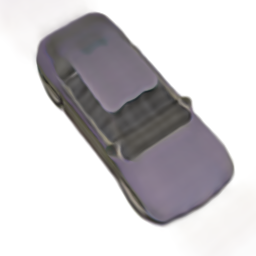}}\hfill
{\includegraphics[width=0.11\linewidth]{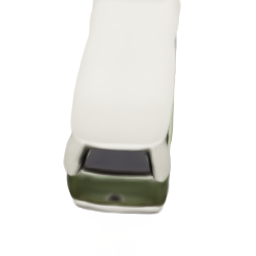}}\hfill
\\

\rotatebox{90}{\makecell{Stage 2}} &
{\includegraphics[width=0.11\linewidth]{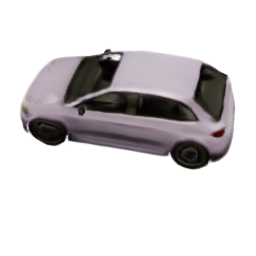}}\hfill
{\includegraphics[width=0.11\linewidth]{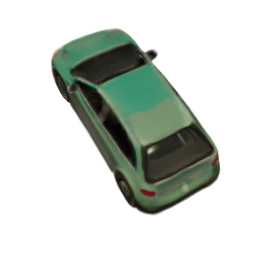}}\hfill 
{\includegraphics[width=0.11\linewidth]{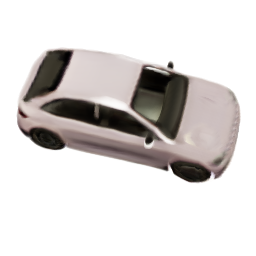}}\hfill
{\includegraphics[width=0.11\linewidth]{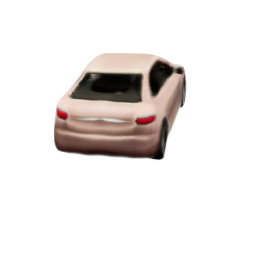}}\hfill
{\includegraphics[width=0.11\linewidth]{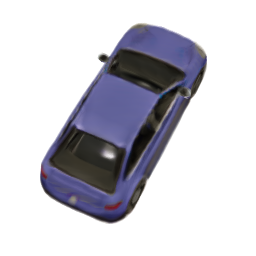}}\hfill 
{\includegraphics[width=0.11\linewidth]{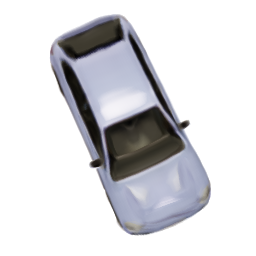}}\hfill
{\includegraphics[width=0.11\linewidth]{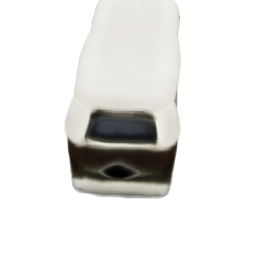}}\hfill
\\

\rotatebox{90}{Stage 3} &
{\includegraphics[width=0.11\linewidth]{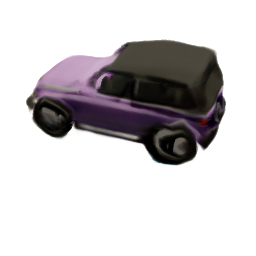}}\hfill
{\includegraphics[width=0.11\linewidth]{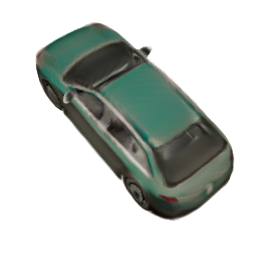}}\hfill 
{\includegraphics[width=0.11\linewidth]{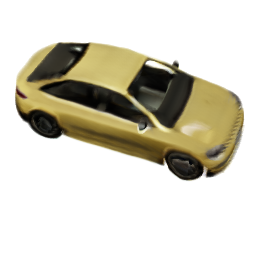}}\hfill
{\includegraphics[width=0.11\linewidth]{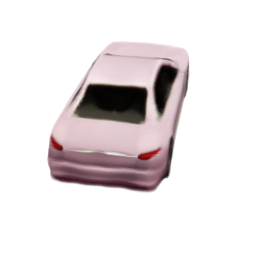}}\hfill
{\includegraphics[width=0.11\linewidth]{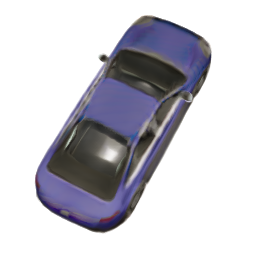}}\hfill 
{\includegraphics[width=0.11\linewidth]{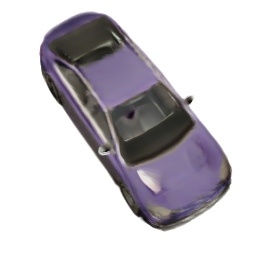}}\hfill
{\includegraphics[width=0.11\linewidth]{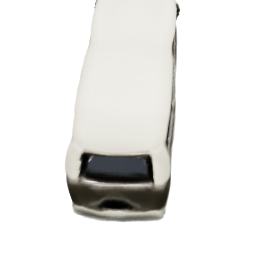}}\hfill
\\

\end{tabular}

\caption{\textbf{Results of stage-wise training:} Given a row of input image GT, following rows are results from different learning framework, in terms of stage-wise training. Direct training from scratch tends to generates blurry images, and frequently fails in reconstruction. Comparing stage2 and stage3, the latter represent better quality on predicting suitable shape and appearance code for reconstruction.}
\label{stagewise_ablation}
\end{figure*}

\subsection{Experimental Settings} 
\subsubsection{Datasets} We evaluate our method on two publicly available datasets of varying complexity: \texttt{cars} from GRAF CARLA dataset~\cite{dosovitskiy2017carla} and \texttt{chairs} are images from Photoshapes dataset~\cite{park2018photoshapes}, following the rendering protocol of~\cite{Oechsle_2020_3DV}. CARLA contains 10K synthetic car images which are 18 different kinds of car models in different rendering viewpoints and colors with resolution $256 \times 256$. Photo contains 150k images with resolution $256 \times 256$. 

\subsubsection{Test Dataset for Quantitative Evaluation}
In addition, there is no public dataset to quantitatively measure the model performance of disentangling attributes from given images and 3D-aware manipulation quality, we build a new test subset of CARLA~\cite{dosovitskiy2017carla}, where there are pairs of images that share one of such attributes, as exemplified in~\figref{proposed_dataset}.
\vspace{-10pt} 

\subsubsection{Evaluation Metric} 
To measure the model performance of disentangling 3D attributes from a given image, we use Fréchet Inception Distance (FID) score~\cite{heusel2017fid} to measure the degree of fidelity for generated images on 1k dataset of CARLA~\cite{dosovitskiy2017carla} whose optimized style codes are offered by EditNeRF~\cite{liu2021editing}.

\begin{table}[!t]
\caption{\textbf{Quantitative evaluation of 3D reconstruction using metric evaluation on CARLA dataset~\cite{dosovitskiy2017carla}.} The best result is shown in bold, and the second best is underlined.}\vspace{-5pt}
\label{recon_comparison_tab}
\begin{center}
\begin{tabular}{lc}
\toprule
 & FID \\ \midrule
SA ~\cite{jang2021codenerf}  & \underline{58.31}  \\ 
CodeNeRF ~\cite{jang2021codenerf}  & 128.82 \\ 
EditNeRF~\cite{liu2021editing}  & 76.63  \\ 
AE-NeRF & \textbf{42.92} \\
\bottomrule
\end{tabular}
\end{center}
\vspace{-10pt}
\end{table}

\subsection{Reconstruction}
We quantitatively evaluate the image quality and accuracy of auto-encoding results using the Frechet Inception Distances (FID)~\cite{heusel2017fid} by reconstructing randomly selected 1K images for CARLA Dataset~\cite{dosovitskiy2017carla} and 1K images for Photoshapes~\cite{park2018photoshapes} Dataset. Quantitative and qualitative results showed in \tabref{recon_comparison_tab} and \figref{recon_comparison_fig} prove that we have significantly higher results compared to previous 3D attribute controlling models such as CodeNeRF~\cite{jang2021codenerf} and EditNeRF~\cite{collins2020editing}. More importantly, as our encoder first learns to map the distribution of generative models and then optimizes the output in an instance-wise manner, the accurate reconstruction with high-fidelity is enabled. We also demonstrate the qualitative results of CLIP-NeRF~\cite{wang2021clip} in a GAN-inversion manner in \figref{auto-encoding-chairs}. Note that at test time the CLIP-NeRF additionally optimizes the model for a given input while AE-NeRF directly predicts the output.

\subsection{Image Manipulation for a Single Image}
In this section, we first evaluate 3D-aware attribute manipulation from a single image. We prove that the proposed method disentangles each attribute of the given image by injecting perturbations to each attribute separately.
The results in~\figref{random_perturbations} show that our model is aware of each attribute and thus separately manipulate the attribute very well while maintaining the others more consistently than compared methods~\cite{park2020swapping, liu2021editing, jang2021codenerf}.
In addition, since EditNeRF~\cite{liu2021editing} and CodeNeRF~\cite{jang2021codenerf} optimize the radiance field focused on given camera pose, 3D attributes are disentangled based on their viewpoint. This property discourages EditNeRF and CodeNeRF to generate realistic images using disentangled attributes which are insufficient for novel viewpoints. On the other hand, AE-NeRF can disentangle 3D attributes regardless of viewpoints by using stage 1 (adversarial learning) where the model is focused on synthesize realistic images from any latent attributes or viewpoints. This ensures the high fidelity of generated images and consecutively reconstructed images and even synthesized images from novel viewpoints.

\begin{table}[!t]
\caption{\textbf{Comparison of AE-NeRF to SA~\cite{park2020swapping}, CodeNeRF~\cite{jang2021codenerf}, EditNeRF~\cite{liu2021editing} on code swapping results with CARLA~\cite{dosovitskiy2017carla}.} We evaluate the swapping attribute results of a source object instance to match a target instance. The best result is shown in bold, and the second best is underlined.}\label{table:disentanglement}\vspace{-5pt} 
\begin{center} 
\renewcommand{\thesubtable}{}
\resizebox{\linewidth}{!}{
\begin{tabular}{l|cccccc}
\toprule
&  \multicolumn{3}{c}{Percep.} & \multicolumn{3}{c}{FID}    \\ \cmidrule(lr){2-4}\cmidrule(lr){5-7}
  &  Sh. & App. & Cam. &  Sh. & App. & Cam. \\
\midrule
SA~\cite{park2020swapping} & 0.2390 & 0.2831 & - & \underline{58.32} & \underline{57.26} & - \\
CodeNeRF~\cite{jang2021codenerf} & 0.3421 & 0.3311 & 0.3212 & 102.5 & 113.21 & 127.52 \\
EditNeRF~\cite{liu2021editing} & \underline{0.1912} & \textbf{0.1142} & \textbf{0.1641} & 97.27 & 77.84 & \underline{97.95}\\
\midrule
AE-NeRF & \textbf{0.1731} & \underline{0.1948} & \underline{0.1691} & \textbf{42.88} & \textbf{41.97} & \textbf{42.29}\\
\bottomrule
\end{tabular}
}
\end{center}
\end{table}

\subsection{Image Manipulation Across Multiple Images}
We then compare our 3D-aware image manipulation across multiple images, e.g., swapping appearance, shape, or camera pose across different instances on CARLA~\cite{liu2021editing} dataset as in EditNeRF~\cite{liu2021editing}. 

In addition, as shown in \tabref{table:disentanglement}, we evaluate our model by swapping the attribute codes, where given a pair of source and target images the source image is manipulated by replacing an extracted certain attribute code from target image, either shape, appearance or camera pose, so that the source image eventually obtains the attribute of target image. We measure the code swapping performance on proposed test set as a ground truth pair.
Comparing functional aspects, EditNeRF~\cite{liu2021editing} manipulates the radiance field given user's editing scribbles by only optimization procedures. This is rather time-consuming for manipulating overall texture or shape of given instance. Furthermore, EditNeRF optimize their radiance field only in the editor's viewpoint, giving inconsistent manipulation result from novel viewpoint. AE-NeRF on the other hand, since the latent features are provided by trained encoder, manipulation of the objects' overall shape and texture is relatively faster than the optimization-based methods.

\begin{figure}[!t]
\centering
{\includegraphics[width=\linewidth]{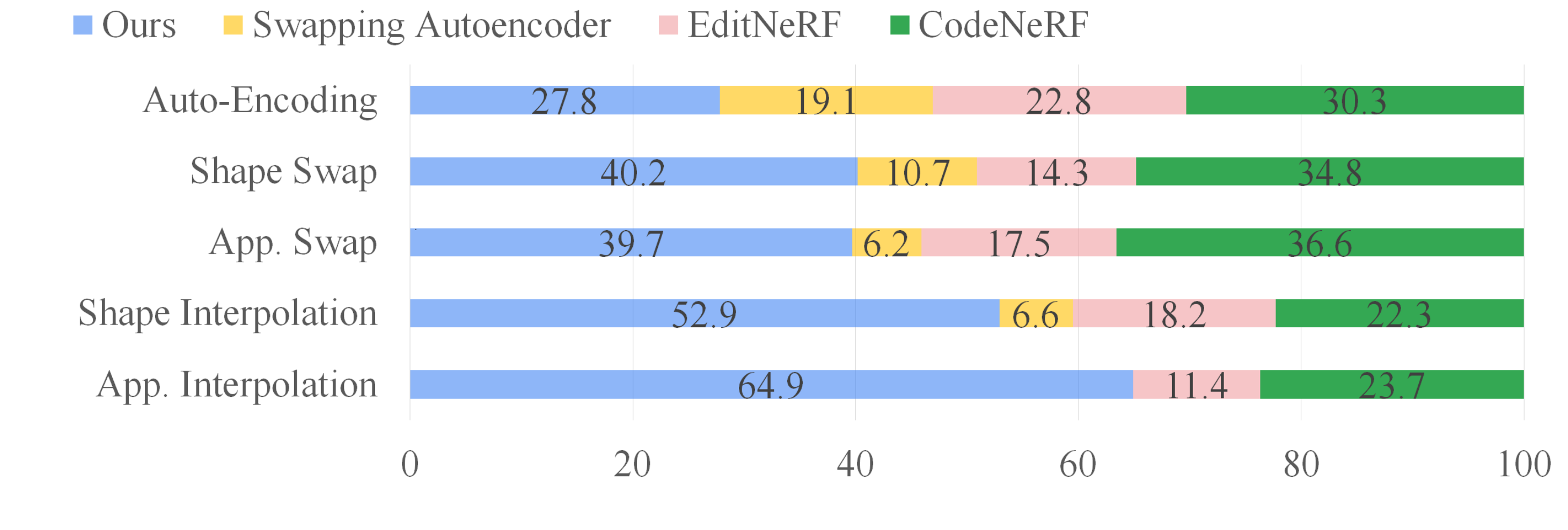}}\hfill
\vspace{-20pt}
\caption{\textbf{User study results}, which is conducted in respect of auto-encoding quality, disentanglement quality, and interpolation ability. Compared to conventional methods~\cite{liu2021editing,park2020swapping, jang2021codenerf}, our AE-NeRF ranks the first in every cases.}
\label{fig_userstudy}
\end{figure}

\subsection{User Study}
We conducted user study on 300 participants to compare the synthesized results obtained from different models. First, we obtained 10 reconstructed images from SA~\cite{park2020swapping}, EditNeRF~\cite{liu2021editing}, CodeNeRF~\cite{jang2021codenerf}, and ours. We showed them to respondents and asked them to indicate 10 images they prefer in aspect of the image reconstruction quality with high-fidelity. Second, to compare disentanglement ability of 3 models and ours, we synthesized 10 images with each attributes swapped. Then we asked subjects to choose 10 images that show preferable disentanglement quality. Finally, we generated 10 images with attribute interpolation from 3 models and ours. We asked participants to pick 10 images that show the most well interpolated images. According to the user study displayed in Fig.~\ref{fig_userstudy}, we could identify that our model is preferred more frequently than any other models, and outperforms SA~\cite{park2020swapping}, CodeNeRF~\cite{jang2021codenerf} and EditNeRF~\cite{liu2021editing} in terms of controllability.

\subsection{Ablation Study}
\subsubsection{Effectiveness of Stage-wise Training.} \tabref{ablation:stagewise-training} and~\figref{stagewise_ablation} show the effectiveness of stage-wise training. As described in \secref{training_strategy}, training the proposed auto-encoder model from scratch (\tabref{ablation:stagewise-training} (\textbf{I}), ~\figref{stagewise_ablation} (\textbf{Scratch})) is extremely challenging and easy to get stuck in local minimum. Thus, we propose a stage-wise training, where we pre-train the decoder in stage 1 to generate pseudo labels used for encoder training in stage 2, followed by stage 3 with losses for disentanglement. We figured out that the performance of our model gets boosted when the model is supervised with \textit{pseudo data} as this strongly enforce the model to map the given image and the generated attributes. The comparison of \tabref{ablation:stagewise-training} (\textbf{III}), (\textbf{IV}) proves that mapping the output distribution of encoder with pseudo labels highly contributes in performance improvement. \\ In addition, \tabref{ablation:stagewise-training} (\textbf{III}) depends on its disentanglement ability on the structural architecture of decoder, we propose additional losses that raise the disentanglement performance of our model as detailed in~\secref{loss functions}. This can be confirmed by comparing~\tabref{ablation:stagewise-training} (\textbf{II}), (\textbf{IV}).

\subsubsection{Effectiveness of Each Loss.} Here, we evaluated effectiveness of each loss function compared to the basic loss for the auto-encoder, i.e., $\mathcal{L}_\mathrm{render}$ and $\mathcal{L}_\mathrm{stage-2}$, as shown in \tabref{ablation:loss_function}. Note that, with $\mathcal{L}_\mathrm{cam}$ that belongs to the $\mathcal{L}_\mathrm{p}$ on \textit{pseudo data}, our model is capable of extracting the camera pose given a single image. $\mathcal{L}_\mathrm{GAN}$ helps the model to synthesize images with fine details such as object's surface and edge. The additional $\mathcal{L}_\mathrm{GLAC}$ raises the disentanglement performance with understanding of consistent object attribute in generated local patch from swapped attributes. With additional convolutional binary classifier, $\mathcal{L}_\mathrm{SwAC}$ improves the disentanglement performance of our model by supervising our encoder with more accurate paired attribute-swapped image. 
\begin{table}[t]
\caption{\textbf{Ablation study on stage-wise training.} The best result is shown in bold.}\vspace{-5pt}\label{ablation:stagewise-training}
\begin{center}
\begin{small}
\begin{tabular}{ll|cccccccccccc}
\toprule
\multicolumn{2}{c|}{\multirow{2}{*}{Stage}}  & \multicolumn{3}{c}{FID}  \\ \cmidrule(lr){3-5}
& &  Sh. & App. & Cam.  \\
\midrule
(\textbf{I}) & Baseline (Stage 3)        
  & {136.49} & {139.32} & {137.13}    \\
(\textbf{II}) & Stage 1, 2        & {47.14} & {49.43} & {46.77}\\
(\textbf{III}) & Stage 1, 3    & {46.52} & {47.79} & {47.37} \\
(\textbf{IV}) & Ours (Stage 1, 2, 3)  & \textbf{42.88} & \textbf{41.97} & \textbf{42.29}\\
\bottomrule
\end{tabular}
\end{small}
\end{center}
\end{table}

\begin{table}[t!]
\caption{\textbf{Ablation study on effectiveness of loss functions.} The best result is shown in bold.}\vspace{-5pt}\label{ablation:loss_function}
\begin{center}
\begin{tabular}{cl|cccccc}
\toprule
&\multirow{2}{*}{Loss combination} & \multicolumn{3}{c}{FID}    \\ \cmidrule(lr){3-5}
& &  Sh. & App. & Cam. \\
\midrule
(\textbf{I}) & $\mathcal{L}_\mathrm{render} + \mathcal{L}_\mathrm{stage-2}$  &  {50.18}    & {50.87}    & {49.43} \\
(\textbf{II}) &(\textbf{I}) + $\mathcal{L}_\mathrm{GAN}$        & {47.14}  & {49.43}    & {46.77}   \\
(\textbf{III}) &(\textbf{II}) + $\mathcal{L}_\mathrm{GLAC}$     & {45.64}  & {45.24}    & {44.57}  \\
(\textbf{IV})& (\textbf{II}) + $\mathcal{L}_\mathrm{SwAC}$      &  {47.46}    & {45.49}    & {46.44} \\
(\textbf{V}) &(\textbf{II}) + $\mathcal{L}_\mathrm{GLAC}$ + $\mathcal{L}_\mathrm{SwAC}$   &  {\textbf{42.88}}    & {\textbf{41.97}}    & {\textbf{42.29}}   \\
\bottomrule
\end{tabular}
\end{center}
\end{table}

\begin{table}[t]
\caption{\textbf{Ablation study on the number of pseudo data.} (a) compares the quality of feature swapped images while (b) shows the relationship between the numbers of pseudo data and the corresponding data diversity. Data diversity does not correlate with the number of pseudo data, and 15K samples is the optimal number of pseudo data used to train the network. }\label{ablation:number of pseudo}\vspace{-10pt}
\begin{center} 
\begin{small}
\subfloat[][]{
\begin{tabular}{l|cccccc}
\toprule
\multirow{2}{*}{Num.}   & \multicolumn{3}{c}{FID}   \\ \cmidrule(lr){2-4}
&  Sh. & App. & Cam. \\
\midrule
5k       &43.91    &42.83 &43.06  \\
10k       &43.56    &42.17 &42.44  \\
15k       &\textbf{42.88}    &\textbf{41.97} &\textbf{42.29}  \\
20k      &43.02    &42.54 &42.43 \\
\bottomrule
\end{tabular}}
\subfloat[][]{
\begin{tikzpicture}

    \begin{axis}[
        xlabel=Num.(1000),
        ylabel=MS-SSIM,
        xmin=-2, xmax=25,
        ymin=0.31, ymax=0.35,
        xtick={5, 10, 15, 20},
        ytick={0.32, 0.34},
        yticklabel style={rotate=90}
        ]
    \addplot[smooth,color=red,mark=*,mark options={fill=cyan,scale=0.7}] plot coordinates {
        (5, 0.3358)
        (10, 0.3325)
        (15, 0.3284)
        (20, 0.3333)
    };
\end{axis}
\end{tikzpicture} }

\end{small}
\end{center}
\end{table}

\subsubsection{Effectiveness of The Number of Pseudo Data.} 
We evaluated how the number of generated pseudo data affects the final model performance. Diverse set of images of high fidelity, with a high variance in the pseudo shape and appearance codes, is essential for supervising the encoder in stage 2. Based on this idea, we first sampled 5k to 20k images and compared the diversity by the mean of MS-SSIM~\cite{pmlr-v70-odena17a} which measures the similarity across all pseudo images generated in stage 1. In this study, We regard the high MS-SSIM as low diversity across images. Note that increasing the number of pseudo data does not always guarantee the data diversity proportionally, described in~\tabref{ablation:number of pseudo} (right). Then we measure Fréchet Inception Distance (FID) score~\cite{heusel2017fid} of the final model performance depending on different numbers of pseudo data. As illustrated in~\tabref{ablation:number of pseudo}, the model achieves better performance with more pseudo data.

\section{Conclusion}
\label{sec:conc}
In this paper, we have presented, for the first time, an auto-encoder architecture based on conditional NeRF for 3D-aware object manipulation, dubbed Auto-Encoding Neural Radiance Fields (AE-NeRF). 
We designed our model consisting of the encoder to extract disentangled 3D attributes from an image, and the decoder to render an output image through disentangled generative NeRF, which is well tailored for 3D-aware object manipulation. To improve the disentanglement, we present two losses, global-local attribute consistency loss and swapped-attribute classification loss. In addition, since training the networks from a scratch is extremely challenging, we presented a stage-wise training scheme, which dramatically helps to boost the performance. We have shown that our method surpasses other state-of-the-art in several benchmarks and manipulation tasks. Moreover, we have conducted extensive ablation studies to validate our design choices.  
\bibliography{egbib}
\bibliographystyle{IEEEtran}

\end{document}